\documentclass[pdflatex,sn-mathphys-num]{sn-jnl}

\usepackage[T1]{fontenc}
\usepackage[utf8]{inputenc}
\usepackage[english]{babel} 

\usepackage{hyperref}
\usepackage{graphicx}%
\usepackage{multirow}%
\usepackage{amsmath,amssymb,amsfonts}%
\usepackage{amsthm}%
\usepackage{mathrsfs}%
\usepackage[title]{appendix}%
\usepackage{xcolor}%
\usepackage{textcomp}%
\usepackage{manyfoot}%
\usepackage{booktabs}%
\usepackage{algorithm}%
\usepackage{algorithmicx}%
\usepackage{algpseudocode}%
\usepackage{listings}%
\usepackage{verbatim}%
\usepackage{afterpage}%
\usepackage{attachfile}%
\usepackage{orcidlink}
\theoremstyle{thmstyleone}%
\theoremstyle{thmstyletwo}%

\theoremstyle{thmstylethree}%

\begin{document}

\title[Article Title]{Unraveling Media Perspectives: A Comprehensive Methodology Combining Large Language Models, Topic Modeling, Sentiment Analysis, and Ontology Learning to Analyse Media Bias}


\author[1,2]{\fnm{Orlando} \sur{Jähde} \orcidlink{0009-0000-6789-7279}}\email{orlando.jaehde@fom-net.de}

\author*[1,3]{\fnm{Thorsten} \sur{Weber} \orcidlink{0000-0002-5090-0651}}\email{thorsten.weber@fom-net.de}

\author[1,4]{\fnm{Rüdiger} \sur{Buchkremer} \orcidlink{0000-0002-4130-9253} }\email{ruediger.buchkremer@fom-net.de}

\affil[1]{\orgdiv{Institute of IT Management and Digitization Research (IFID)},\\ \orgname{FOM University of Applied Sciences in Economics and Management},\\ \orgaddress{\street{Toulouser Allee 53}, \city{Düsseldorf}, \postcode{40476}, \state{North Rhine-Westphalia}, \country{Germany}}}

\affil[2]{\href{https://scholar.google.com/citations?user=EQ9adRAAAAAJ}{Google Scholar}}
\affil[3]{\href{https://scholar.google.com/citations?user=7JohnPMAAAAJ}{Google Scholar}}
\affil[4]{\href{https://scholar.google.com/citations?user=zVGyk2QAAAAJ}{Google Scholar}}

This is a preprint of an article published in the Journal of Computational Social Science. The final authenticated version is available online at: https://doi.org/10.1007/s42001-025-00372-0


\abstract{Biased news reporting poses a significant threat to informed decision-making and the functioning of democracies. This study introduces a novel methodology for scalable, minimally biased analysis of media bias in political news. The proposed approach examines event selection, labeling, word choice, and commission and omission biases across news sources by leveraging natural language processing techniques, including hierarchical topic modeling, sentiment analysis, and ontology learning with large language models. Through three case studies related to current political events, we demonstrate the methodology's effectiveness in identifying biases across news sources at various levels of granularity. This work represents a significant step towards scalable, minimally biased media bias analysis, laying the groundwork for tools to help news consumers navigate an increasingly complex media landscape.}


\maketitle

\section{Introduction}\label{sec1}

News is essential for keeping people and citizens informed. Reporting on world events shapes how we view our world and forms societies \cite{Hamborg2019, Kroon2023}. Sometimes, to such an extent that the public perception of events no longer has an appropriate connection with reality \cite{Opperhuizen2019}. News can unintentionally influence public opinion but can also be deliberately exploited \cite{Galgoczy2022}. Especially with political news, the scale can be enormous. Democracies rely on voters making informed decisions based on reality-aligned information \cite{Bernhardt2008, Wilson2020}. Biased reporting can potentially sway election outcomes  \cite{Bernhardt2008}. As far back as 2004, 78\% of the U.S. population believed news to be biased \cite{Baron2006}, a problem that continues to persist \cite {Wilson2020}. Therefore, there is a need for action: Efforts should be directed toward reducing biased news propagation while empowering individuals to navigate media bias better.

The phenomenon of inclination or prejudice in the news is called media bias in literature \cite{Hamborg2019, Bernhardt2008, Baron2006, DAlonzo2022, Park2009}; sometimes, it is referred to as partisan or hyperpartisan news \cite{Naredla2022, Shultziner2021}. Scientific methodologies have been developed repeatedly to detect and analyze media bias in news. Initially, these methods were predominantly qualitative and required substantial human involvement \cite{Hamborg2019}. While effective, they were limited in scope and could not be scaled or generalized \cite{Hamborg2019}. Today, we can process vast amounts of unstructured data, which was previously unmanageable. Particularly in the case of media bias analysis, advances in the field of natural language processing (NLP) and machine learning are opening new possibilities. Scalable approaches to media bias analysis are emerging \cite{Hamborg2019, Nemeth2022}. 

The concept of media bias, and bias in general, has a central issue: "bias relative to what?" \cite{Puglisi2015}. In this paper, we begin with the assumption that humans are inevitably biased. Hence, news and large language models (LLMs) trained on human data are biased. Therefore, news has no neutral or ’zero-bias’ reference point \cite{Shultziner2021, Brandtzaeg2017, Kang2022}. Following this logic, media bias is natural and does not have to be willful, though it can be\footnotemark. Addressing this conundrum, this work follows a similar approach as, e.g., Shultziner and Stukalin in \cite{Shultziner2021}. Instead of pinpointing a ’zero-bias’ baseline, we compare the presentation of the same political topic across different newspapers over the same timeframe. This comparative analysis helps us understand and highlight the variations in bias.
\footnotetext{We should also note that there are differing viewpoints in the literature. Some argue that something considered media bias must be intentional \cite{DAlessio2000, Mullainathan2002}. However, this paper operates on the assumption that bias, to some extent, is inherent and unavoidable in news.}

Many recent strategies for analyzing media bias rely on supervised machine-learning techniques, which require pre-labeled data \cite{Rodrigo-Gines2024, Nemeth2022}. However, this labeling process is typically performed by human beings, introducing an inherent bias into the model that mirrors the perspectives of the individuals who annotated the data. It becomes particularly significant when labeling explicit biased information, like political leanings (left or right).

Our research aims to address this issue by developing a methodology to investigate media bias in political news that is simultaneously scalable, incorporating machine learning methods such as NLP and LLMs, and as bias-free as possible, minimizing the need for human intervention. Bias-free because a bias-induced method defies the purpose of comparing and measuring bias in news. Scalable because this work should contribute in a way that allows real-world applications to implement the methodology and benefit society. Therefore, it must apply to extensive data sets independent of the news topic.

This paper is organized as follows: The first section overviews relevant forms of media bias and describes related work on media bias analysis. The second section describes our proposed methodology by giving an overview and then describing each step in detail. The proposed methodology is tested in three case studies in the third section. Two case studies deal with a specific news topic, and the third case study is a cross-topic analysis. In the next section, the methodology is discussed based on the results of the case studies and compared with different studies. We then describe the limitations of our method and highlight future research fields. The last section concludes our work and highlights the main contributions.

\section{Media bias and its detection}\label{sec2}

This section outlines the various forms of media bias and clarifies the focus of this study. It describes the forms of media bias that are in focus and puts them into context. It concludes with a firm literature review of current detection methods for the relevant forms of media bias.

\subsection{Media bias and its forms}\label{subsec21}

Media bias has varied, sometimes even partly opposite definitions \cite{Rodrigo-Gines2024}. As stated, we define it as inclination or prejudice in the news, which can be intentional but does not have to be, as it is part of the human condition \cite{Kang2022}. To put more context, some studies separate opinion from bias \cite{Rodrigo-Gines2024}. We do not, but view opinion as biased. 

However, intentional or not, the forms of media bias stay the same. To elaborate further, we build upon the framework introduced in \cite{Park2009} and enhanced in \cite{Hamborg2019}. In this framework, the news production process is divided into six main steps: The process starts with actual news events, which can be anything. The second step is gathering, in which journalists select the facts they want to report. This includes news event selection, source selection, commission, and omission of specific perspectives, facts, or parts of the news event. Third is writing, in which bias can be added through labeling and word choice. The fourth step is the presentation style, including placement and size allocation of stories and picture selection and explanation. The fifth step is the result, which is the produced news article. A sixth step was added in \cite{Hamborg2019} and includes the perception of the news consumer. Hamborg, Donnay, and Gip also add the concept of spin bias, which represents the overall bias of a news article and spans gathering, writing, and editing. While there can be various motives for media bias that influence the spin of an article, the framework is ultimately independent since whatever the motive, even with none, media bias manifests in these described forms.

\begin{figure}[h]%
\centering
\includegraphics[width=0.8\textwidth]{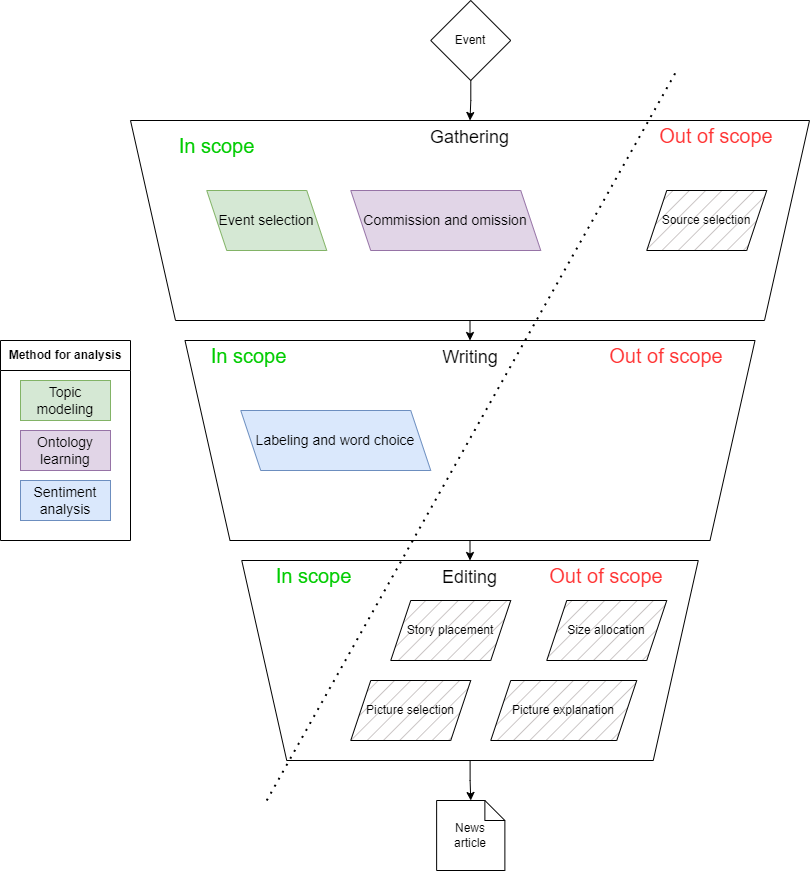}
\caption{Forms of media bias in and out of scope of this work. Based on \cite{Park2009} and \cite{Hamborg2019}}\label{fig1}
\end{figure}

Steps one to five are illustrated in Fig. \ref{fig1}. It maps out the three media bias types we focus on: Event selection, commission and omission, and labeling and word choice. Event Selection pertains to whether a news source chooses to cover a particular event or a specific category of events, and if so, to what extent. Commission and Omission refer to including or excluding specific perspectives, facts, or aspects in the news. Labeling and word choice revolves around how events or their components are presented, be it in a positive, neutral, or negative light, and the terminology used to describe them (for instance, ”military conflict” vs. ”war”). Detecting and analyzing the other forms of media bias described in the framework is not in the scope of this study but can be part of future research.

\subsection{Automatic detection of media bias}\label{subsec22}

Automatic media bias detection often involves NLP and machine learning \cite{Rodrigo-Gines2024} - Non-deep learning methods were standard in the past. In recent years, however, deep learning models have proven more capable of detecting media bias \cite{Rodrigo-Gines2024}. Both recurrent neural networks (RNNs) and language transformers have been used for this task. Transformers were found to outperform RNNs: They are better at modeling the sequential structure of a sentence \cite{Vaswani2017}, which leads to better results than models based on linguistic features or RNNs \cite{Baly2020a}.

Several studies rely on a training model that classifies articles as biased or non-biased \cite{Quijote2019, Agrawal2022, Sinha2021, Chen2018, Fan2019, Spinde2022}. While the definition of bias in those studies is similar, the analyzed form and research goal differ, preventing comparisons from being straightforward. To ease the comparison of our methodology with other studies, we provide examples that analyzed the same or at least similar forms of media bias.

According to \cite{Hamborg2019} and \cite{Rodrigo-Gines2024}, only a few approaches exist to analyze event selection bias. In \cite{Bourgeois2018}, the \textit{Global Database of Events, Language, and Tone} (GDELT) was utilized to span a matrix between sources and events. It helped identify patterns of differences and similarities across various news sources. In \cite{Saez-Trumper2013}, articles on the same topic or covering the same event are clustered using cosine similarity. The fundamental idea in both studies is the same \cite{Rodrigo-Gines2024, Bourgeois2018}: they organize articles from varying news sources into topics and then compare their respective coverage. This methodology is adopted in this study as well by implementing topic modeling.

For commission and omission bias, Park et al. and Erhardt et al. leveraged entity or aspect analysis to evaluate their occurrence in various articles \cite{Ehrhardt2021, Park2009}. For instance, in \cite{Ehrhardt2021}, those entities are used to estimate the political slant of different news sources; besides that, no comparison of news sources is made. \cite{Park2009} offers \textit{"aspect-level browsing"}, which provides readers with articles on the same topic mentioning different aspects. However, it does not show which aspects are omitted or committed in newspapers across multiple articles. The current literature lacks a focused exploration of the commission and omission bias in news and the methods to identify or address them correctly.

The issue of labeling and word choice bias is frequently addressed using NLP techniques, mainly through the sentiment analysis of articles \cite{Hamborg2019, Spinde2022}. With the rise of transformer models \cite{Vaswani2017}, the quality of sentiment classification models was significantly enhanced \cite{Gandhi2023}. Besides sentiment analysis, D'Alonzo and Tegmark present an unsupervised machine-learning method to create phrase statistics \cite{DAlonzo2022}. These phrase statistics allow us to locate newspapers along two dimensions, left-right bias and establishment bias, and reflect previous bias classifications based on human judgment.

\section{Structured analysis of media bias}\label{sec3}

\begin{figure}[h]%
\centering
\includegraphics[width=0.7\textwidth]{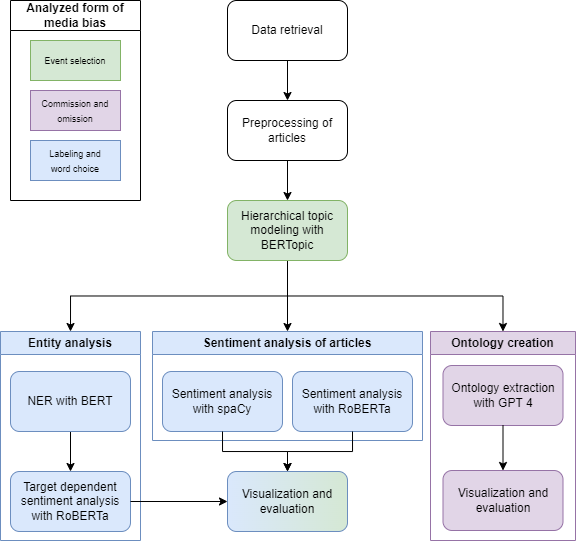}
\caption{Overview of our proposed methodology and mapping to analyzed forms of media bias}\label{fig2}
\end{figure}

This section outlines a systematic approach to examining media bias in news articles. We extract additional attributes from the data by leveraging various NLP techniques and large language models. These attributes are subsequently subjected to statistical analysis. We apply Python as the programming language of choice for all these steps. Fig. \ref{fig2} provides an overview of the steps involved and the respective forms of media bias analyzed at each stage. We then delve into each step, elaborating on the resulting outcomes. Although the final outputs are unique to a specific form of media bias, the methods employed are somewhat interdependent.

The dataset should encompass each article's textual content, title, and publishing newspaper. Supplementary details such as the author’s name, publication date, accompanying images, and URL (if applicable) could potentially enhance topic representation and streamline evaluation processes. However, these elements are not currently integrated into our statistical analysis. In forthcoming studies, we plan to explore whether these attributes contribute value to our methodology. It is particularly relevant considering the growing prominence of multimodal approaches in artificial intelligence (AI) \cite{Molenaar2023, Gandhi2023, Xu2023}.

\subsection{Preprocessing and hierarchical topic modeling}\label{subsec31}

In general, the particulars of data preprocessing depend on the characteristics of the present data \cite{Alexandropoulos2019} as well as on the further steps of data processing and models used \cite{Siino2024a}. According to \cite{Siino2024a}, minimal or no preprocessing yielded the best results for transformer models, depending on the specific dataset and transformer model used. In this context, minimal preprocessing refers to removing stopwords or converting text to lowercase. Given these findings, and considering our methodology primarily employs transformer models, no standard preprocessing techniques like those described in \cite{Siino2024a, Symeonidis2018, Hickman2022} are necessary. 

Nonetheless, two optional preprocessing steps to increase the quality of results are recommended. Firstly, we suggest excluding non-English texts to increase the quality of results and ensure human understanding. Transformer models tend to perform best within the English language \cite{Goyal2021}. The library \textit{spacy\_language\_detection} is used for that \cite{Beauchemin2021}. Secondly, we suggest removing noise, meaning all text unrelated to the article's topic is removed. Depending on the data source, this can include different elements, e.g., advertisements, sections with related articles, and metadata about the article, like author and date in the text body. It reduces the chance of topics being formed based on information that is not related to the actual topic of the article. Similarly, it also increases the reliability of sentiment analysis and named entity recognition (NER). Besides that, duplicates based on article text and title are not to be removed. Two different newspapers publishing the same article or one newspaper publishing the same article multiple times can contribute to forms of media bias \cite{Park2009}.

Topic modeling helps to map a document corpus to topics \cite{Buchkremer2019}, which is why we use it to detect news topics in a news article corpus. As an algorithm, BERTopic with its default steps is used \cite{Grootendorst2020}. As of January 2024, \cite{Grootendorst2024a}, those steps are still as initially proposed by Grotendorst \cite{Grootendorst2020}: The embedding is done with SBERT \cite{Reimers2019}, dimensionality reduction with UMAP \cite{Mcinnes2020} and clustering with HDBSCAN \cite{Campello2013}. A CountVectorizer \cite{Scikitlearn} is chosen as a vectorizer, and c-TF-IDF \cite{Grootendorst2020} is used for weighting.

While the methods for each step are primarily interchangeable, HDBSCAN remains a critical component in this methodology. Its appeal lies in its ability to provide hierarchical topics, which, in turn, facilitates the reduction of topics or, put differently, increases the amount of data within a topic. This feature is beneficial when altering the scope of inter-topic comparisons. Moreover, HDBSCAN is efficient with large amounts of data and identifies noise within it instead of adding it to the nearest topic. To assess the quality of the generated topics, we can refer to the ”Red Flags for Degenerate Clustering Results” as a standard quality check \cite{Schubert2017}. While they were initially proposed for evaluating DBSCAN results \cite{Schubert2017}, they stay helpful for the lowest hierarchy level of HDBSCAN results. Nevertheless, the human judgment of the resulting topics remains compulsory to ensure reasonableness.

Post-topic modeling, topics that solely encompass articles from a single newspaper are integrated with the noise cluster. This process eliminates topics that do not allow for inter-newspaper topic comparisons. Subsequently, the quality of these topics is reassessed. To simplify the process of working with hierarchical topics, we enhance the structure of the BERTopic native pandas \cite{Mckinney2011} data frames: the topic data frame resulting from \textit{topic\_model.get\_topic\_info()} is concatenated with the topics in higher hierarchy levels which are obtained by calling \textit{topic\_model.hierarchical\_topics()} with \textit{topic\_model} being a fitted instance of \textit{BERTopic()} \cite{Grootendorst2024}. The data in columns containing information about a topic is also concatenated and inherited to the higher-level topic. Also, a column containing the ID of the parent topic is added to ease browsing through topics when integrating the data in a user interface.

\subsection{Sentiment analysis of articles}\label{subsec32}

The sentiment analysis is done for each article's title and text body. RoBERTa \cite{Liu2019} is used for sentiment analysis of titles. Since RoBERTa takes at most 512 tokens \cite{Liu2019} and longer inputs will be truncated, spaCy is used for the sentiment analysis of text bodies. Article bodies are often longer than 512 tokens. All measured sentiment scores are then compared. The sentiment analysis could also be done before topic modeling; however, it saves resources only to analyze the articles not in the noise cluster. For clarity, any text that has undergone analysis, whether a title or the body of an article, is referred to as a ’document’ in the subsequent sections.

For spaCy, the \textit{spacytextblob} library is used along with \textit{en\_core\_web\_sm} pre-trained pipeline \cite{Edwardes2024, Explosion2023}. The training data of this pipeline consists of different kinds of news media, among others \cite{Weischedel2013, SpaCy2023}, and it is widely used for NLP tasks with news data \cite{Jadhav2020, Vychegzhanin2019, Kamel2019, Mandalapu2019}. The results of \textit{spacytextblob} provide an overarching sentiment score for the entire document and individual sentiment values for contributing words. These features enable us to conduct an in-depth analysis and verify the plausibility of the results.

We use the RoBERTa model of the \textit{NewsSentiment} library since it is specifically trained for sentiment analysis of news data \cite{Hamborg2021}. For titles, the \textit{infer\_from\_text()} function is called with empty strings for the "left" and "right" parameters, and the document is passed to the "target" parameter. As a result, it is coded as \textit{infer\_from\_text("", document, "")}. The output of this function is the probability for each of three classes: positive, neutral, and negative. Those probabilities add up to one. To compare these results with those from spaCy and ease working with them, we calculate a simplified sentiment score by subtracting the negative probability (which is also positive in value) from the positive probability \cite{Harth2023}. The neutral probability is preserved separately, as this data is valuable for later evaluations.

\subsection{Entity analysis}\label{subsec33}

We first identify specific entities mentioned within the articles for the entity analysis. This process, known as Named Entity Recognition (NER), is accomplished using BERT \cite{Devlin2019}. Second, we examine the sentiment or tone expressed towards these entities using RoBERTa.

We use the \textit{transformers} library \cite{Huggingface2024} to load the \textit{bert-base-NER} tokenizer and model as trained in \cite{Devlin2019}. The model was trained on the standard CoNLL-2003 Named Entity Recognition dataset comprising Reuters news articles \cite{sang2003}. Even though the English data of this dataset is from between August 1996 and August 1997, \cite{sang2003}, and newer datasets are available \cite{Tedeschi2022, Ringland2020}, Liu and Ritter concluded that models trained on the CoNLL-2003 dataset still work well due to good generalization \cite{Liu2023}. Other studies confirm that by comparing the performance of \textit{bert-base-NER} or the \textit{bert-base} model with other NER tools \cite{Hu2024, Polignano2021}. \textit{bert-base-NER} categorizes the entities into one of four entity groups: person, organization, location, and miscellaneous entity. The parameter \textit{aggregation\_strategy} of the \textit{transformers.pipeline()} is used to prevent the entities from being split into multiple parts after running the model. This setting ensures the results are on an entity rather than a token level \cite{Devlin2019}. ]. The result for each article is an array of Python dictionaries, one for each entity. The result is transformed by creating a new data frame where each row represents an entity, making it easier to work with and re-evaluate the data. The former dictionary keys are now the column names, and one more column containing an article ID has been added to keep each entity linked with its corresponding article. Table \ref{tab1} shows the data frame structure with sample data.

\begin{table}[h]
\caption{Entity data frame structure with sample data}\label{tab1}
\begin{tabular}{@{}lllllll@{}}
\toprule
 & entity group & score & word & start & end & article id\\
\midrule
0 & LOC & 0.1 & USA & 0 & 3 & 0 \\
1 & ORG & 0.5 & United Nations & 14 & 28 & 0 \\
... & PER & 0.3 & Biden & 151 & 156 & 0 \\
n & MISC & 0.9 & Supernova & 60 & 69 & m \\
\botrule
\end{tabular}
\end{table}

The "start" and "end" columns specify the position of the identified entity in the document. This information is utilized for analyzing the target-dependent sentiment by using the pre-trained RoBERTa model from the \textit{NewsSentiment} library. In literature, target-dependent sentiment analysis takes the sentence as context to determine the target sentiment \cite{Hamborg2021, Zhang2023, Wu2022}. That means this time, the "left" and "right" parameters of \textit{infer\_from\_text("", document, "")} are filled depending on the position of the target in the sentence. The following example illustrates this: To analyze the target-dependent sentiment for the entity "fool" in the sentence, "And here I am, for all my lore, The wretched fool I was before.", the sentence needs to be split into "left", "target" and "right":

\bigskip

"And here I am, for all my lore, The wretched " 

"fool",

" I was before.".

\bigskip

To split the article into sentences, we use the \textit{sent\_tokenize} function from the \textit{nltk} library \cite{Bird2006}. To prevent inefficient looping through the resulting array, a pandas data frame with four columns is created: article ID, start, end, and sentence. This way, the sentence of an entity can be selected with a single line of code. The output for each entity is simplified like before by calculating a simplified sentiment score and keeping the neutral score as an additional column. Each entity's newspaper is joined using the article ID for evaluation purposes.

\subsection{Ontology learning}\label{subsec34}

The concept of "Ontology" has roots in philosophy \cite{Biemann2005, Antunes2022}, but has gained significant traction in the field of information science, thanks to the rise of the Semantic Web \cite{Karabulut2024, Watrobski2020, Maedche2001}. While it is used ambiguously in information science \cite{Watrobski2020, Antunes2022, Karabulut2024, Reyes-Pena2019, Biemann2005}, fundamentally, it can be understood as "an explicit specification of a conceptualization’’ \cite{Gruber1993}. However, the level of observation, or "specialization," can differ \cite{Antunes2022}. Five key components are typically used to construct an ontology \cite{Reyes-Pena2019}: 

\begin{itemize}
\item \textbf{Classes}, the main formalized elements of the conceptualization
\item \textbf{Objects}, being instances of classes
\item \textbf{Relationships}, which are links between classes or objects
\item \textbf{Functions}, which calculate information from the other ontology elements
\item \textbf{Axioms}, that compose restrictions, rules, and logic correspondences definition
\end{itemize}

\begin{figure}[h]%
\centering
\includegraphics[width=0.9\textwidth]{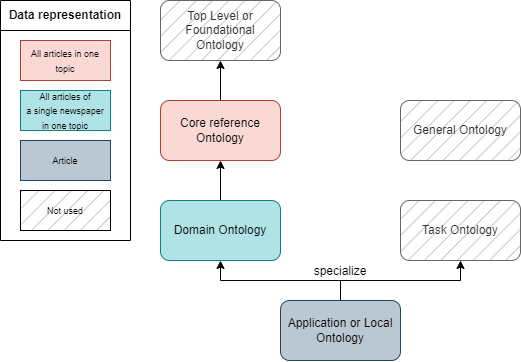}
\caption{Classification of ontologies. Based on \cite{Roussey2011}.}\label{fig3}
\end{figure}

The concept of varying specialization levels is effectively captured in the two classifications described in \cite{Antunes2022}. While both classifications are similar, Roussey, Pniet, Kang, and Corcho offer a more comprehensive classification in their work \cite{Roussey2011}, especially adding one more hierarchy level in the classification consisting of core reference ontologies and general ontologies \cite{Roussey2011, Stephan}. Fig. \ref{fig3} shows their classification and the corresponding data representation in this methodology. 

Three different types of ontologies will be created. On the lowest level, an ontology created based on a single article is classified as a local ontology. It represents the knowledge contained within that particular article, or in other words, it conceptualizes the article topic based on it. An ontology created based on all articles of a single newspaper on one topic is classified as a domain ontology. It conceptualizes the topic based on all articles from a single newspaper. On the highest level used in this methodology, an ontology created based on all articles on one topic is classified as a core reference ontology, conceptualizing a topic based on all articles about it. The concept of core reference ontologies, introduced in \cite{Roussey2011}, is essential for this work since it allows for different domain ontologies of the same topic and integrates them into a standard core ontology, thus providing a reference when analyzing domain or local ontologies \cite{Antunes2022}.

The term "ontology learning" (OL) was coined by M\"adche and Staab \cite{Maedche2001} and can be understood as "extracting conceptual knowledge from input and building ontolog[ies] from [it]" \cite{Watrobski2020}. While this task was not accomplished in the 2000s, as Biemann wrote in \cite{Biemann2005}, he auspiciously concluded with the open question "whether the manual approach of hand-coding semantics will be outperformed by inconsistent, statistical black-box methods again." About 20 years later, statistical black-box methods are on the rise \cite{Xi2023}, and large language models (LLMs) like GPT-4 \cite{OpenAI2023} and Gemini \cite{GeminiTeam2023} are prominent not only in the scientific but also the public discourse. However, they are still scarcely used for OL \cite{Watrobski2020, Giglou2023, Konys2018}. 

At first, local ontologies are created by extracting the knowledge of each article. For this, a prompt template for GPT-4 is used. The extracted ontologies consist of classes, objects, and the relationships of objects. To process the results, the answers need to be in computer-readable format. This is also specified in the prompt template in Fig. \ref{fig4}. For each article, a new conversation is started. Otherwise, the context of the conversation, speaking another article, impacts the ontology creation.

\begin{figure}[h]%
\textit{Your task is to create an ontology for a given news article. The ontology consists of the following elements:\\
\hspace*{0.4in}- Classes: are the main formalized element of the article\\
\hspace*{0.4in}- Objects: are the representation of the main objects of the article and represent \hspace*{0.4in}instances of a class\\
\hspace*{0.4in}- Relationships: are links or relationships between the objects for representing \hspace*{0.4in}the ontology structure\\
Please answer in the following JSON format:\\
\{\\
\hspace*{0.2in}"Class:" ["ClassName"],\\
\hspace*{0.2in}"Object": [\{\\
\hspace*{0.4in}"Name": "ObjectName",\\
\hspace*{0.4in}"InstanceOf": "Class"\\
\hspace*{0.2in}\}],\\
\hspace*{0.2in}"Relationship": \{\\
\hspace*{0.4in}"RelationshipName": \{\\
\hspace*{0.6in}"RelationshipFrom": "ObjectName",\\
\hspace*{0.6in}"RelationshipTo": "ObjectName"\\
\hspace*{0.4in}\}\\
\hspace*{0.2in}\}\\
\}\\
Please consider that "RelationshipName" should be an active verb and optionally contain a preposition.\\
Please create an ontology of the following article:}
\caption{Prompt template used for extracting ontology elements}\label{fig4}
\end{figure}

The results are stored in a data frame containing the article ID, the corrected JSON reply, the newspaper, and the topic of each article obtained by again joining on the article ID. The JSON replies allow for automatic quality checks, which is crucial for two reasons: Firstly, tackling the hallucination problem of LLMs \cite{Ji2023, Martino2023, Rawte2023, Zhang2023a, Huo2023} and, hence, secondly preventing the necessity of manual verification for each article. A consistence-based estimation \cite{Zhang2023a} of the uncertainty of the results is achieved by developing evaluation metrics for logical consistency \cite{Rawte2023}. The following quality checks measure those metrics:

\begin{itemize}
\item \textbf{Object-Class consistence:} Check whether the "InstanceOf" value of objects is an existing class
\item \textbf{Object-Object consistence:} Check whether each object only occurs once per article
\item \textbf{Object-Relation consistence:} Check whether the "RelationshipFrom" and "RelationshipTo" values of relationships refer to previously defined objects
\end{itemize}

Each metric is expressed as a percentage, signifying the accuracy rate for each assessment. These percentages are calculated by subtracting the result of dividing the count of errors by the total number of corresponding elements by one. Any elements that do not pass the quality assessment are eliminated from the outcome. Human judgment is necessary to determine whether a metric result is acceptable, keeping the human-in-the-loop (HITL) \cite{Mosqueira-Rey2023}. In future research, we may establish threshold values to decide whether the whole reply should be discarded and re-prompted, resulting in a trust level that allows for human-on-the-loop (HOTL) \cite{Li2020, Nahavandi2017}. 

\subsection{Visualization and evaluation}\label{subsec35}

We use the topic modeling result to evaluate the event selection media bias. To determine the relevant topic level, we refer to a dendrogram that outlines the hierarchy of topics. When a topic above the base level is selected, we combine the articles from the subordinate topics. Per newspaper \textit{N}, the percentage \textit{p} of articles \textit{a()} about the topic \textit{t} of interest is calculated by dividing their number by the total number of articles of the correlating newspaper (including articles of the noise cluster), as shown in Eq. \ref{eq1}. This method compensates for the varying publication rates among different newspapers. The number of articles, or their cardinality, is $|a()|$.

\begin{equation}
p(N_t) = \frac{|a(N_t)|}{|a(N)|}\label{eq1}
\end{equation}

Next, the mean percentage $\bar{p}$ of all newspapers for topic \textit{t} is calculated. This value shows how many articles are published, on average, about topic \textit{t} about the total number of articles. This value is used for cross-topic analysis to compare the significance of topics in a given dataset. Contrasting topic size accounts for relation and eliminates a possible bias from single or few newspapers publishing a significant part of articles about topic \textit{t}. For inter-topic analysis, the deviation \textit{d} of \textit{p} from $\bar{p}$ is calculated, enabling a comparison of different newspapers by showing whether they publish more or less articles about topic \textit{t} than average, as shown in Eq. \ref{eq2}.

\begin{equation}
d(N_t) = p(N_t) - \bar{p_t}\label{eq2}
\end{equation}

\begin{figure}[h]%
\centering
\includegraphics[width=0.4\textwidth]{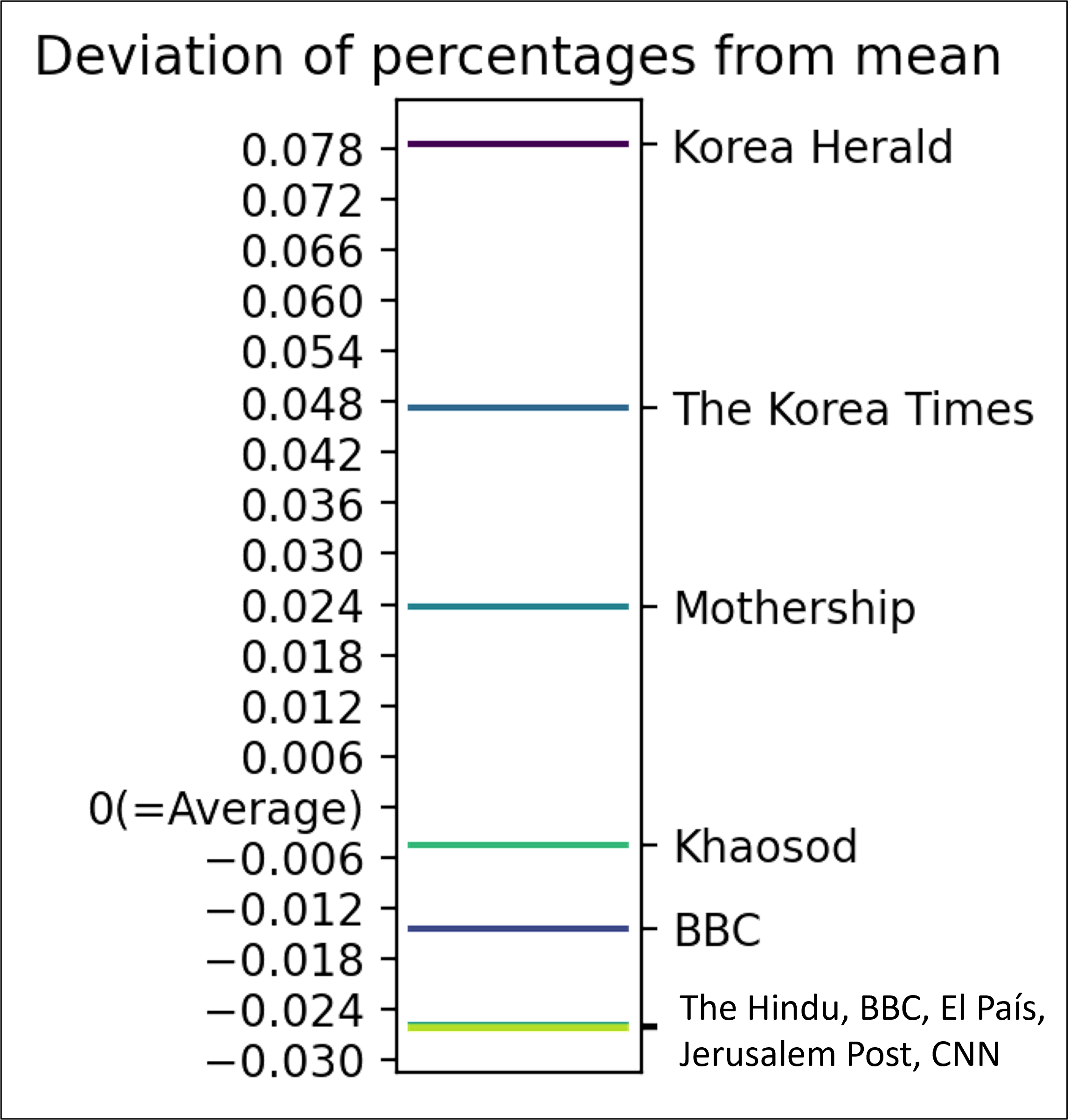}
\caption{Example for visualization of deviation of percentages from mean}\label{fig5}
\end{figure}

Several visual representations can illustrate this concept; one example is presented in Fig. \ref{fig5}. The figure represents a topic about KPop from the dataset described in Section \ref{subsec41}. However, it is essential to note that the figure does not include newspapers that did not cover the topic. Despite this omission, such information can still be retrieved from the topic modeling results by adding the missing newspapers and assigning a zero as a value.

The analysis of media bias in terms of labeling and word choice is conducted at various levels: the title, the main body of the text, and the entity level. The evaluation method mirrors the approach used to assess media bias in event selection. At first, the mean sentiment score $\bar{s}$ is calculated for each newspaper as shown in Eq. \ref{eq3}, with \textit{s(x)} being the sentiment score of \textit{x}: 

\begin{equation}
\overline{s(N_t)} = \frac{\sum\limits_{i=0}^{| a(N_t) |} s(a_i(N_t))}{|a(N_t)|}\label{eq3}
\end{equation}

Please note that $s(a_i(N_t))$ can either be the sentiment score of the title of an article or the sentiment score of the main body of an article, as it works for both. However, they are not mixed. While the entity evaluation method keeps the same principle, the formula slightly differs. Here, the cardinality and sentiment scores of entities \textit{e} are relevant, as shown in Eq. \ref{eq4}. Also, $\bar{s}$ is now only calculated for an entity in t; hence, it is represented as $\bar{s_e}$.

\begin{equation}
\overline{s_e(N_t)} = \frac{\sum\limits_{i=0}^{| e(N_t) |} s(e_i(N_t))}{|e(N_t)|}\label{eq4}
\end{equation}

Then, the mean sentiment score of the newspaper means is calculated. The newspapers' means are not weighted by article count to again prevent a possible bias from a single or few newspapers publishing a significant part of articles about topic \textit{t}. This method applies to all levels, title, text body, and entity level. We then calculate the sentiment deviation \textit{sd} in the same manner as we do for the publishing rate, as shown in Eq. \ref{eq5}.

\begin{equation}
sd(N_t) = s(N_t) - \bar{s_t}\label{eq5}
\end{equation}

\begin{figure}[h]%
\centering
\includegraphics[width=0.8\textwidth]{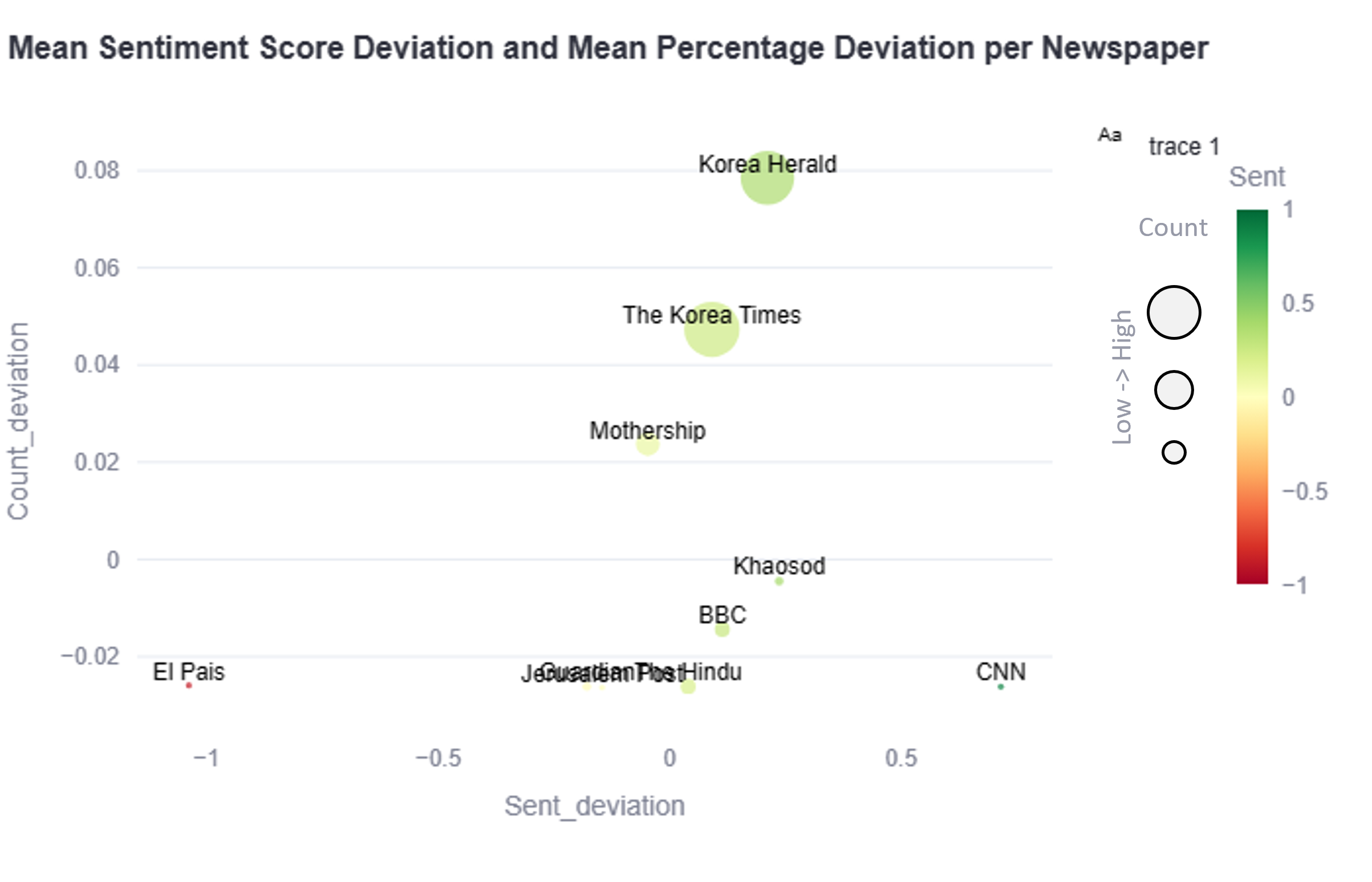}
\caption{Media bias spectrum}\label{fig6}
\end{figure}

We can visualize sentiment data in the same way as depicted in Fig. \ref{fig5}. However, we are introducing a new concept for a more effective comparison: a media bias spectrum, as shown in Fig. \ref{fig6}. Again, as an example, this figure presents a topic related to KPop, sourced from the dataset we discuss in Section \ref{subsec41}. 

The x-axis represents the deviation of the sentiment score to the mean sentiment score, be it title sentiment, text body sentiment, or target-dependent sentiment of an entity. The mean across all newspapers is always zero in this visualization, as the deviation to it is plotted. The y-axis represents a different thing when evaluating articles or entities. For articles, it represents the deviation of the mean described above, shown in Fig. \ref{fig5}. For entities, the number of each entity per newspaper is calculated, and then the mean of those values is calculated. Again, the deviation to this mean is then used to compare the different newspapers and hence is taken as the y-axis in the media bias spectrum for entities. In addition to the mean deviation, the media bias spectrum also shows whether a newspaper’s mean sentiment score is positive or negative through color. The corresponding color scale is right in Fig. \ref{fig6}. The size of each point is determined by either the number of articles in a newspaper or the number of times an entity has been mentioned. This way, the absolute values of each axis are also represented in the media bias spectrum. The bigger the point, the more significant it is regarding deviation on the x- and y-axis, as it has a considerable amount of underlying data. Nevertheless, the size of the point itself already holds information about the event selection media bias for media bias spectrums on the article level or information about labeling and word choice media bias on the entity level. In essence, the media bias spectrum is a precise tool to measure and compare newspapers regarding two media bias forms: event selection, labeling, and word choice.

\begin{figure}[h]%
\centering
\includegraphics[width=0.8\textwidth]{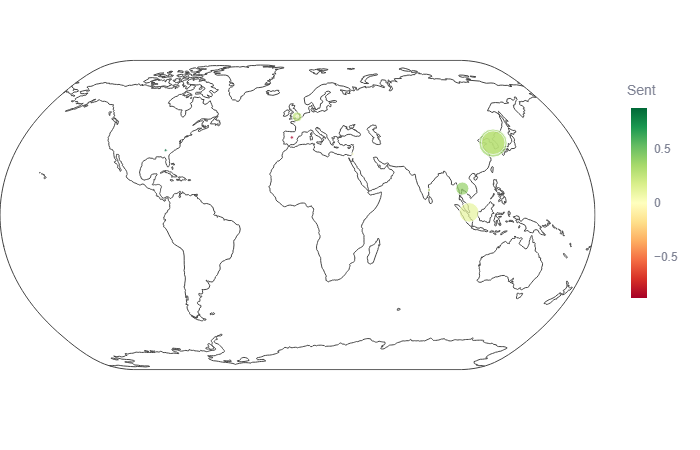}
\caption{Publishing rate and average sentiment score of newspapers plotted on a world map}\label{fig7}
\end{figure}

Adding the newspaper's headquarters locations to the data makes it possible to plot the data on a map. It may offer insight into regional correlations or outliers. To illustrate this, Fig. \ref{fig7} plots the KPop topic on a map. The size of the bubbles represents the deviation of the newspaper-specific publishing rate from the mean publishing rate, and the color indicates the sentiment score. However, information about the sentiment deviation and the number of articles is lost.

\begin{figure}[h]%
\centering
\includegraphics[width=0.9\textwidth]{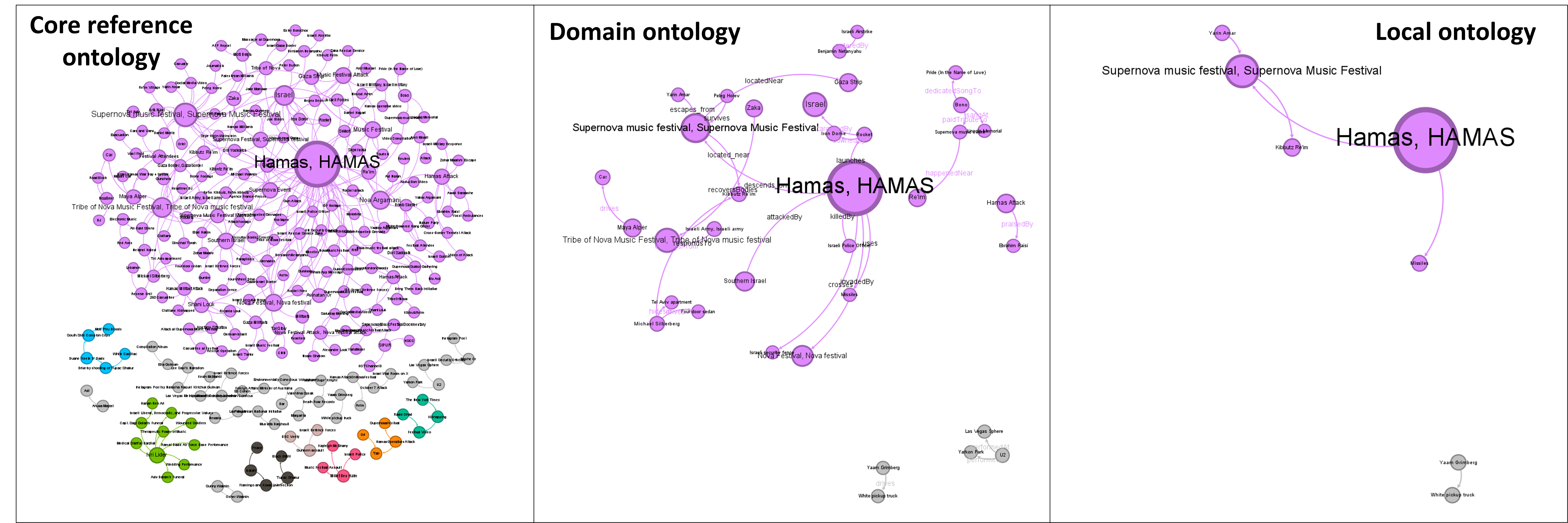}
\caption{Filtering the same data for different level ontologies}\label{fig8}
\end{figure}

The ontologies are visualized in graph visualization software to analyze the commission and omission of media bias. We suggest \textit{Gephi}, as it natively includes graph calculations and is free to use \cite{Bastian2009}. After removing relations that failed the object-relation consistency check, the remaining relations are enriched with their corresponding newspaper and article ID. This data frame then creates a graph containing all the necessary information. The resulting ontology is the core reference ontology. The nodes are sized based on their degree, meaning the count of in- and outgoing relations determined their size. Nodes and edges are colored by the inferred classes of the nodes, which resulted from calculating the statistical interference proposed by Zhang and Peixoto in \cite{Zhang2020}. This method allows the detection of assortative communities in networks. Placing the nodes should be handled individually, as different data may require different algorithms for the layout. The number of times a node has been mentioned might differ from the degree of the corresponding node. The degree is better for visualization as it highlights which nodes are the most prominent ”actors” instead of showcasing plain mentions. The network, domain, and local ontologies can be retrieved by filtering. It is illustrated in Fig. \ref{fig8}.

\section{Case studies}\label{sec4}

In this section, the methodology is tested on three case studies. Each result is manually revised and tested for coherence. Unlike other studies \cite{Spinde2021, Quijote2019, Hamborg2019a, Sinha2021}, we do not test whether a bias in one of the analyzed forms correlates with an ideological bias, like left-right or establishment bias \cite{Bennett2021, DAlonzo2022}. This rather sensitive task requires a comprehensive dataset of all articles from different newspapers through a representative time frame. Following this logic, we want to stress that the goal is to test the proposed methodology rather than do a complete and reliable news analysis on the chosen topics, as this would require a more comprehensive dataset. Section \ref{subsec54} describes our limitations in more detail.

At first, the underlying dataset and its processing are described. The processing covers methodology steps run on the whole dataset, i.e., preprocessing, topic modeling, entity analysis, and sentiment analysis of articles (overview given in Fig. \ref{fig2}). Then, three different case studies are conducted. The first one analyses articles about a single topic: the Hamas attack on Israel, specifically the attack of October 7, 2023, on the music festival (”Supernova Sukkot Gathering”). The second case study deals with a topic on level two in the topic hierarchy. It consists of two topics about Putin’s visit to Beijing, China, during the Third Belt and Road Forum for International Cooperation (BRI), held between October 17 and 18, 2023, and one about the BRI itself. The third case study analyzes the news across all topics in the dataset. While focusing only on labeling and word choice bias, it evaluates the potential for cross-topic analysis of media bias through the proposed methodology.

This choice of case studies allows for testing the methodology on different topics, a different number of articles, and a different scope, either a single topic, a hierarchical topic, or all topics. This ensures a comprehensive test of the methodology.

The case studies were conducted on a Windows 11 Home x64 PC with an AMD Ryzen 7 7700, a NVIDIA GeForce RTX 3080, an American Megatrends Inc. BIOS, version 1811, and 32 GB RAM. Two different virtual environments were used, one using Python 3.10 and one using Python 3.8. It was necessary due to compatibility restrictions of the \textit{NewsSentiment} library.

\subsection{Dataset overview}\label{subsec41}

\begin{figure}[h]%
\centering
\includegraphics[width=0.6\textwidth]{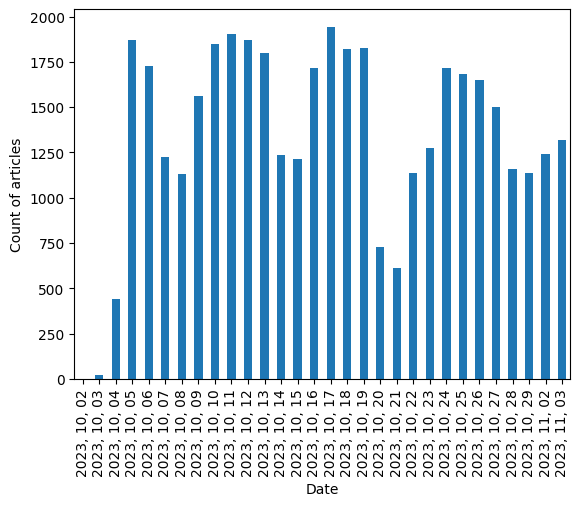}
\caption{Count of articles per date}\label{fig9}
\end{figure}

For data retrieval, a custom scraping script was built. The data retrieval aimed to create a comprehensive dataset regarding the origin country while focusing on newspapers to be read in the respective country. The Digital News Report 2023 of the Reuters Institute \cite{Newman2023} was used to achieve this. It reports, among other things, the most consumed online news brands. However, for the dataset, only those newspapers qualified that had their headquarters in the respective country. There were limitations to this approach: First, some countries need to be added to the report. Second, not all newspapers that are ranked offer English articles. Third, for efficiency reasons, only newspapers with an RSS feed are qualified for scraping. Lastly, some newspapers had scraping protection ruling them out. The resulting newspaper selection is shown in Table \ref{tab2}. In total, 40.385 articles have been scraped from RSS feeds of 37 different newspapers, dating between October 2 and November 3, 2023. Fig. \ref{fig9} details the counts of articles per date. Generally, fewer articles are published on weekends. Since adjusting the scraper to run correctly took a few days, the article counts before October 5 are not representative. However, since the date is not a relevant dimension in the proposed methodology for now, this issue can be ignored.

\begin{sidewaystable}
\caption{Overview of selected newspapers}\label{tab2}%
\begin{tabular*}{\textwidth}{@{\extracolsep\fill}llllllllcccccccccccc}
\toprule
Newspaper & Country & State/City & Rank\footnotemark[1] & Homepage\\
\midrule
9News & Australia & Sydney & 3 & https://www.9news.com.au/\\
The Business Standard (TBS) & Bangladesh & Dhaka & country not in report & https://www.tbsnews.net/\\
VRT nws & Belgium & Brussel & 2 & https://www.vrt.be/vrtnws/en/\\
CBC & Canada & Toronto & 1 & https://www.cbc.ca/\\
South China Morning Post & China & Hong Kong & not in ranking & https://www.scmp.com/\\
Deutsche Welle & Germany & Bonn & not in ranking & https://www.dw.com/en/\\
France24 & France	 & Paris & not in ranking & https://www.france24.com/en/\\
India Today & India & New Delhi & 7 & https://www.indiatoday.in/\\
NDTV (New Delhi Television) & India & New Delhi & 1 & https://www.ndtv.com/\\
The Hindu & India & Chennai & 11 & https://www.thehindu.com/\\
Antara & Indonesia & Jakarta & not in ranking & https://en.antaranews.com/\\
The Journal & Ireland & Dublin & 2 & https://www.thejournal.ie/\\
Irish Mirror & Ireland & Dublin & 10 & https://www.irishmirror.ie/\\
Times of Israel & Israel & Jerusalem & country not in report & https://www.timesofisrael.com/\\
The Jerusalem Post & Israel & Jerusalem & country not in report & https://www.jpost.com/\\
ANSA & Italy & Rome & 3 & https://www.ansa.it/english/\\
the japan times & Japan & Tokyo & not in ranking & https://www.japantimes.co.jp/\\
Mexico News Daily & Mexico & n.a.\footnotemark[2] & not in ranking & https://mexiconewsdaily.com/\\
NL TIMES & Netherlands & Amsterdam & not in ranking & https://nltimes.nl/\\
thefirstnews & Poland & Warsaw & not in ranking & https://www.thefirstnews.com/\\
Al Jazeera & Qatar & Doha & country not in report & https://www.aljazeera.com/\\
PravdaReport & Russia & Moscow & country not in report & https://english.pravda.ru/\\
Moscow Times & Russia/Netherlands\footnotemark[3] & Moscow/Amsterdam\footnotemark[3] & country not in report & https://www.themoscowtimes.com/\\
asiaone & Singapore & Singapore & 6 & https://www.asiaone.com/\\
Mothership & Singapore & Singapore & 1 & https://mothership.sg/\\
The Straits Times & Singapore & Singapore & 3 & https://www.straitstimes.com/global\\
The Korea Herald & South Korea & Seoul & not in ranking & https://www.koreaherald.com/\\
The Korea Times & South Korea & Seoul & not in ranking & https://www.koreatimes.co.kr/www2/index.asp\\
El País & Spain & Madrid & 2 & https://english.elpais.com/\\
Khaosod English & Thailand & Bangkok & 4 & https://www.khaosodenglish.com/\\
Bangkok Post & Thailand & Bangkok & 14 & https://www.bangkokpost.co.th/\\
BBC & UK & London & 1 & https://www.bbc.com/\\
The Guardian & UK & London & 2 & https://www.theguardian.com/europe\\
CommonWealth Beacon & USA & Boston, Massachusetts & not in ranking & https://commonwealthbeacon.org/\\
FOX News & USA & New York, New York & 3 & https://www.foxnews.com/\\
ABC News & USA & New York, New York & 12 & https://abcnews.go.com/\\
CNN & USA & Atlanta, Georgia & 2 & https://edition.cnn.com/\\
\botrule
\end{tabular*}
\footnotetext[1]{Represents the rank of online newspaper brands in the Digital News Report 2023 of the Reuters Institute \cite{Newman2023}}
\footnotetext[2]{Information not found}
\footnotetext[3]{Moved its headquarters in 2022}
\end{sidewaystable}

The distribution of articles varies across the newspapers. It ranges from 91 articles from CommonWealth Beacon to 10,230 articles from The Hindu. Since the methodology leverages relative values to compare event selection bias by calculating a percentage, it accounts for uneven distributions, as described in Section \ref{subsec35}. Therefore, the data is appropriate even though ”The Hindu” is an outlier regarding article count. However, the fewer newspaper articles in the dataset, the less significant the results of its analysis become because the influence of outliers is relatively more substantial. The methodology accounts for this by keeping the information about the absolute number of articles or entities mentioned in the media bias spectrum. In general, it needs to be considered that the scraped articles do not represent all of the articles in each newspaper, limiting the significance of newspaper-related findings since including them in the analysis could lead to different results.

\subsection{Dataset processing}\label{subsec42}

At first, all 75 non-English articles were removed (even though only English RSS feeds were scraped, we found non-English articles). 40,308 articles remained. For noise reduction, ten randomly chosen articles from each newspaper were manually checked for noise. Noise was found in articles in all newspapers, which was then removed using regex. The newspaper-specific regex commands were applied to all the corresponding newspaper articles. This method balanced the effort and benefit of noise removal. Working with the resulting data in the following steps showed that while some articles remained noisy, most contained little or no noise.

The topic modeling resulted in 717 topics, including the noise topic. The noise topic contained 13,639 articles representing $\approx$33.9\% of all articles. While it is 3.9\% above the recommendation in \cite{Schubert2017}, it was considered acceptable since the biggest cluster contained only 699 articles, showcasing a good distribution of articles across topics. 67 topics, each with articles from only one newspaper, were identified and merged with the noise topic. After merging, 650 topics, including the noise topic, remained. The noise topic contained 16,077 articles, representing $\approx$39.9\% of all articles. Still, since the most significant topic showed a good distribution of articles, with only 567 articles, the result was accepted. 

\begin{figure}[h]%
\centering
\includegraphics[width=0.8\textwidth]{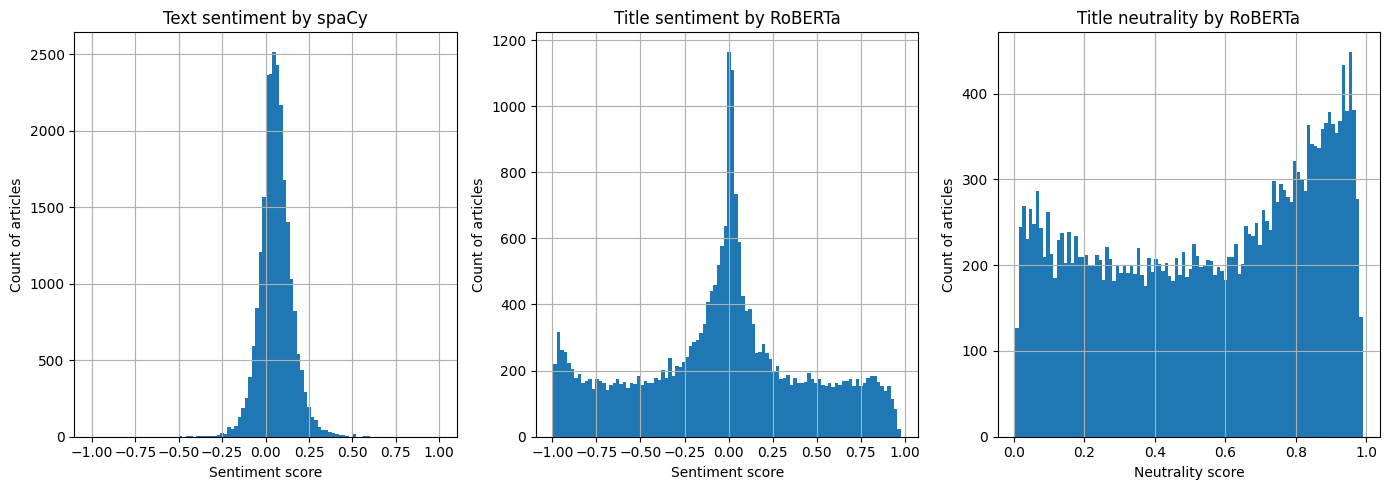}
\caption[help]{Count of articles per sentiment or neutrality score\footnotemark[1]}\label{fig10}
\end{figure}

Next, the sentiment analysis was done as described in Section \ref{subsec32}. Fig. \ref{fig10} shows the distribution of articles per sentiment score or, in the case of the right subplot, per neutrality score. Overall, the documents tend to be neutral, as seen in each subplot. When comparing the left to the middle subplot, it can be seen that spaCy fails to detect polarity in news texts. In contrast, the RoBERTa model trained by Hamborg and Donnay \cite{Hamborg2021} successfully identifies subliminal valuations in news titles. 

\footnotetext[1]{Note: When viewing, please pay attention to different documents analyzed in the left subplot (text bodies instead of titles) and a different score measured in the lower right subplot (neutrality instead of simple sentiment).}

Three examples for this are given in Table \ref{tab3}: while the sentiment score by spaCy measures no or almost no positive sentiment, RoBERTa measures a positive sentiment and does a better job of indicating the relatively positive nature of the article titles (for this comparison, we analyzed the titles with spaCy as well).

\begin{table}[h]
\caption{Title sentiment scores compared}\label{tab3}%
\begin{tabular}{@{}lllllll@{}}
\toprule
Title & RoBERTa sentiment score & spaCy sentiment score\\
\midrule
For one Jewish family, education & 0.902083 & 0.0\\
\quad about family cancer history pays dividends & & \\
OTSI develops unified portal for & 0.560795 & 0.133333\\
\quad agricultural statistics for Ministry of Agriculture & & \\
Shivpal and Dhaval pull off surprise wins & 0.742123 & 0.064965\\
\botrule
\end{tabular}
\end{table}

The NER was performed following sentiment analysis as described in Section \ref{subsec33}. They used sentences as context was not eligible on the given dataset as many sentences were too long. Hence, instead, 150 characters left from the target, or the start of the document, and 150 characters suitable of the target, or the end of the text, were used as context. A total of 1,238,281 entities were extracted. Due to this high number, resource limitations were encountered. It was not possible to extract the target-dependent sentiment of all entities. Instead, the top ten entities per topic were determined by count. Then, only their target-dependent sentiment was analyzed by looping through the topics, meaning that a top entity of topic \textit{x} will not be analyzed in topic \textit{y} if it is not also a top entity of \textit{y}. In this process, 275,066 entities were analyzed. A \textit{streamlit} application has been developed to evaluate and visualize event selection, labeling, and word choice media bias. Due to a lack of funding, ontology learning, as described in Section \ref{subsec34}, was done only for the topics part of this case study. Prompting GPT-4 for all 24,231 non-noise articles would cause high costs (approximately € 3,000). It led to no cross-topic analysis of ontologies.

\subsection{Hamas attack on Israel ("Supernova Sukkot Gathering")}\label{subsec43}

This case study is structured as follows: First, the media bias spectrum is viewed on three different levels: article titles, text bodies, and entities. Second, the ontologies for each case study are viewed and evaluated. After each step, it is assessed whether the machine learning method (topic modeling, sentiment analysis, ontology learning) successfully completed its task. 

The topic \textit{"138\_festival\_music\_argamani\_car"} (t138) dealt with the Hamas attack on Israel, specifically the attack on the ”Supernova Sukkot Gathering” on October 7, 2023. The topic consisted of 47 articles from 19 newspapers published between October 8 and 27.

At first, the topic modeling result was evaluated by manually reading the article titles. One article was not about the attack and was wrongly added to t138. It is the only article of El País on this topic. All other articles dealt with the attack but partly focused on different aspects. While not perfect, studies confirm that BERTopic has the best results compared to other topic modeling techniques \cite{Grootendorst2020, Gan2024}. 

Next, the RoBERTa title sentiment score was examined. The corresponding media bias spectrum is visualized in Fig. \ref{fig11} and the underlying data is in Table \ref{tabA1} in Appendix \ref{secA2}.

\begin{figure}[h]%
\centering
\includegraphics[width=0.8\textwidth]{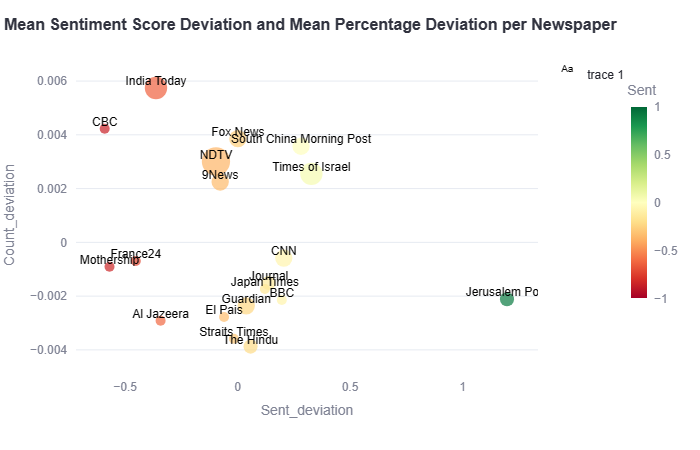}
\caption{Media bias spectrum of t138 based on RoBERTa title sentiment score}\label{fig11}
\end{figure}

To review this result, the titles of the newspapers were checked. The media bias spectrum was found to properly represent the sentiment of the titles, meaning the title sentiment was assessed well with the chosen method. Newspapers like India Today and CBC explicitly named acts of terror in their titles. Newspapers like Fox News or South China Morning Post had more moderate or unemotional titles or a mix of titles differing in sentiment, hence averaging in the mean. Jerusalem Post, on the other hand, refrained from describing the acts of violence during the attack and instead focused on somewhat positive or hopeful topics when mentioning the attack. To illustrate this, some titles are listed below.\\

\textbf{India Today} - Relatively more negative sentiment in titles than average:
\begin{itemize}
\item Blood stains ‘Free Love’: How the massacre at Israel's trance Music festival unfolded
\item Videos: How Hamas fighters swooped in, terrorised, abducted festivalgoers in Israel
\item Video: Hamas terrorists shot festivalgoers in their cars, blocked escape routes
\end{itemize}

\textbf{Fox News} - Close to average sentiment in titles:
\begin{itemize}
\item From the deadly desert rave to the front lines, Israeli reserve soldier recounts concert massacre
\item Israeli music festival survivor describes horror of Hamas-led attack that left 260 dead
\item Who is Noa Argamani? Woman kidnapped by Hamas terrorists at trance music festival in Israel
\end{itemize}

\textbf{Jerusalem Post} - Relatively more positive sentiment in titles than average:
\begin{itemize}
\item Israel at war - Artists bring hope to the wounded and the soldiers
\item Stars of the Israeli electronic scene aim to raise funds for war victims
\end{itemize}

When instead the spaCy text sentiment score was examined, the media bias spectrum changed, as shown in Fig. \ref{fig12}. The underlying data can be found in Table \ref{tabA2} in Appendix \ref{secA2}. Considering the distribution of the overall text sentiment score by spaCy in Fig. \ref{fig10}, it is expected that there are fewer differences in sentiment in absolute terms than when using the RoBERTa title sentiment score. To come up with this, the result could have been normalized. However, the information about the actual difference in sentiment score is lost when normalizing. Since this information was considered to be valuable, no normalization was done.

\begin{figure}[h]%
\centering
\includegraphics[width=0.8\textwidth]{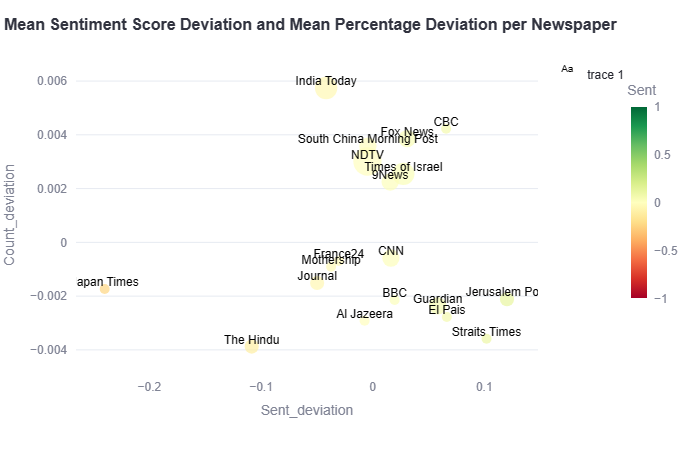}
\caption{Media bias spectrum of t138 based on spaCy text sentiment score}\label{fig12}
\end{figure}

For reviewing the spaCy result, the word attributions were checked as they explain the result deterministically. In contrast to RoBERTa, spaCy shows how its sentiment score was calculated by listing all words that impacted the overall score and its correlating score. SpaCy did not determine the positive or negative nature of the text topic, and it focused more on the positive or negative words used. It was not enough for the articles in t138 to grasp the overall sentiment in the text.

Following titles and text, the entities were analyzed. The top ten entities of t138, along with their average target-dependent sentiment score and neutrality score, are in Table \ref{tab4}. The entity ”im” refers to the location of the attack, Kibbutz Re’im. Here, the NER correctly determined the entity as a location but failed to extract the whole location. On average, there is a negative sentiment toward Hamas. For the following entities, the average target neutrality score is higher than the absolute value of the target score; therefore, they are treated as neutral on average. The slightly negative average target-dependent sentiment score of, for example,”im” and ”Gaza” likely (the used RoBERTa model is a black box) comes about due to the negative context in which they are mentioned. However, the model was well suited for its task since the average target neutrality score is higher.

\begin{table}[h]
\caption{Top ten entities of t138 sorted by "Average target-dependent sentiment score"}\label{tab4}%
\begin{tabular}{@{}lllllll@{}}
\toprule
Entity & Average target-dependent sentiment score & Average target neutrality score\\
\midrule
Hamas & -0.9199 & 0.0545\\
im & -0.2121 & 0.539\\
Gaza & -0.1805 & 0.6619\\
Gaza Strip & -0.0905 & 0.7544\\
Israeli & -0.076 & 0.695\\
CNN & -0.0095 & 0.9239\\
Israel & 0.0019 & 0.7708\\
Sukkot & 0.0908 & 0.7279\\
Jewish & 0.0945 & 0.6537\\
Supernova & 0.1262 & 0.6173\\
\botrule
\end{tabular}
\end{table}

\begin{figure}[h]%
\centering
\includegraphics[width=0.8\textwidth]{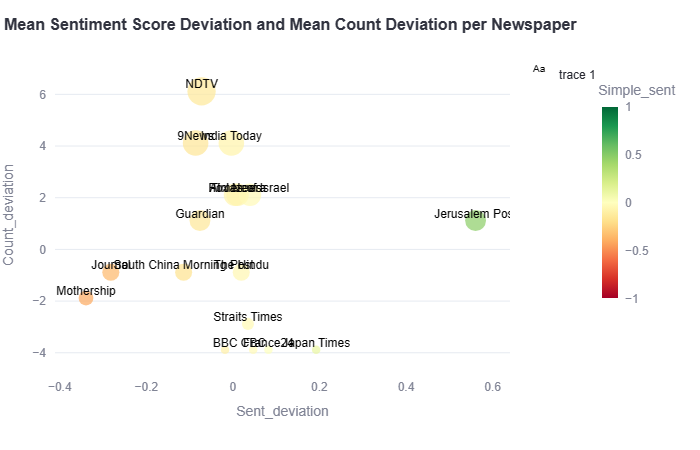}
\caption{Media bias spectrum of the entity "Israeli" in t138}\label{fig13}
\end{figure}

Since Table \ref{tab4} only offers average scores, the entities were also viewed individually by plotting their media bias spectrum. The results for each entity varied. For instance, all newspapers that mentioned ”Hamas” in their articles in t138 have a negative sentiment towards ”Hamas” in those articles. The underlying data can be found in Table \ref{tabA3} in Appendix \ref{secA2}. In contrast, the media bias spectrum of the entity ”Israeli” exhibits more differences in newspapers, as shown in Fig. \ref{fig13}. "Israeli" itself is more of a specification of the following noun, e.g., "Israeli history" or "Israeli army." 

The result was reviewed by manually reading the parts of the articles in which "Israeli" was mentioned. Among other things, The Journal wrote about actions from Israel before and after the attack, which led to a more negative target-dependent sentiment. Newspapers close to average sentiment, like NDTV or The Guardian, either had a high neutrality score or had a mix of positive and negative mentions. Jerusalem Post wrote more positively about "Israeli" entities. To illustrate this, some example sentences are listed below.\\

\textbf{The Journal} - Relatively more negative sentiment than average:
\begin{itemize}
\item "In the ensuing Israeli air strikes, at least 560 people have been killed in the Gaza Strip, [...]"
\item "We are just used to rockets launched from the enclave, which has been under Israeli blockade since Hamas took control in 2007."
\end{itemize}

\textbf{NDTV} - Close to average sentiment:
\begin{itemize}
\item "The Israeli army said tens of thousands of soldiers were deployed to fight terrorists [...]"
\item "[...] the suffering of ordinary Gazans in Israeli bombings in the densely packed strip of land."
\end{itemize}

\textbf{The Jerusalem Post} - Relatively more positive sentiment than average:
\begin{itemize}
\item "Israeli artists have played a crucial role in this heartwarming display of commitment, [...]"
\item "[...] he believes that Israeli music has a significant role in society [...]"
\end{itemize}

For articles from "Mothership," the result did not represent the target-dependent sentiment well. In general, the result for each newspaper got more accurate as more mentions of the entity existed. Overall, the implemented method did well in identifying the most relevant entities of a topic and measuring the sentiment towards them. 

\begin{figure}[h]%
\centering
\includegraphics[width=0.9\textwidth]{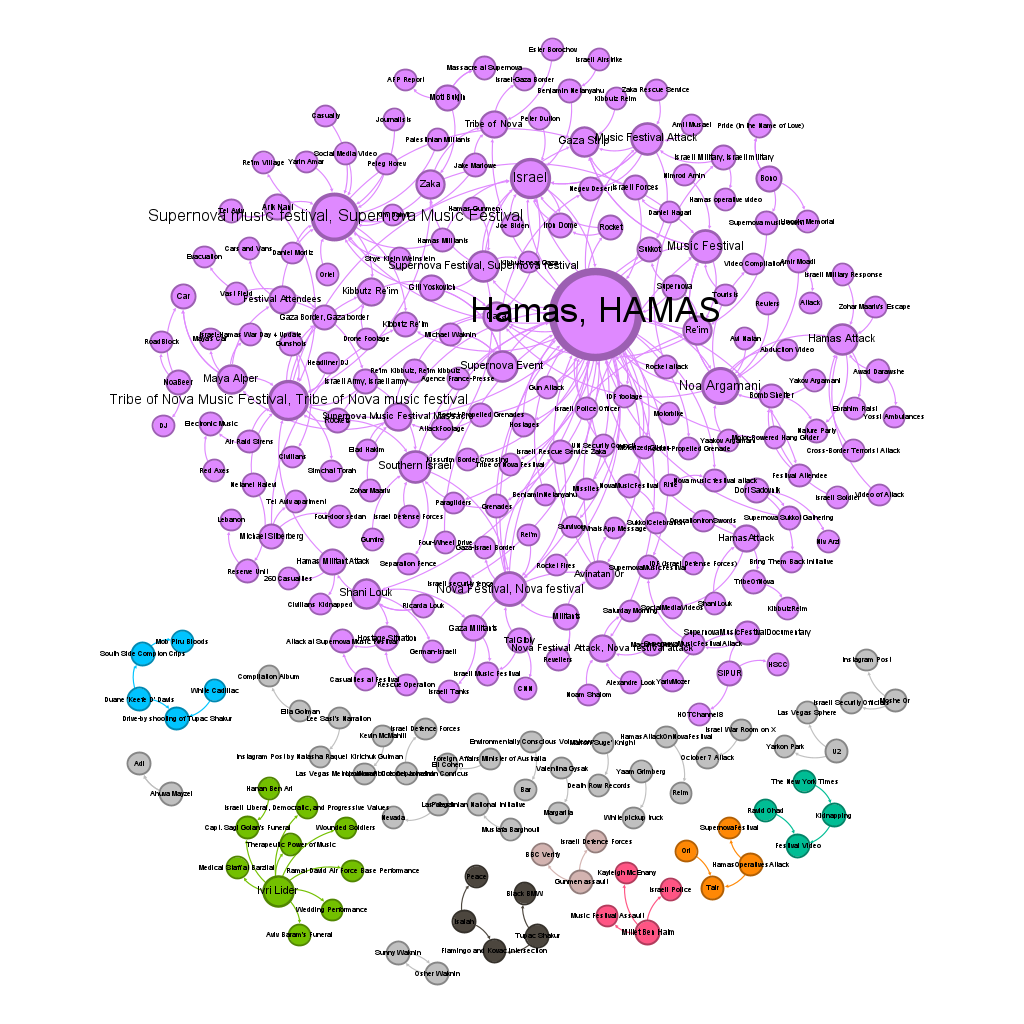}
\caption{Core reference ontology of t138}\label{fig14}
\end{figure}

To understand omission and commission bias, the created ontologies were analyzed. t138 had an object-class and object-object consistency of 100\%, meaning no errors were made in this regard by GPT-4. However, 78 out of 405 relationships failed the object-relation consistency check, resulting in a rate of $\approx$80.7\%. Relationships that failed the check were removed. The resulting core reference ontology comprising all articles of t138 is shown in \ref{fig14}. The placement of the nodes was automized by running the \textit{Force Atlas} algorithm and afterward \textit{Noverlap} in Gephi \cite{Bastian2009}. The nodes were sized based on their degree, meaning the count of in- and outgoing relations determined their size. Nodes and edges were colored by the inferred classes of the nodes, which resulted from the statistical interference calculation. Nodes with the same name but different capitalization were merged by concatenating their names with the delimiter ","

However, this did not account for nodes with the same semantics. For example, multiple nodes represent the Supernova Sukkot Gathering, as can be seen in the assortative community surrounding the "Hamas" node: "Nova Festival," "Tribe of Nova Music Festival," and ”Supernova Music Festival" are three different nodes representing the Supernova Sukkot Gathering. Except for "Ivri Lider" (green node), each node in assortative communities smaller than one with "Hamas" has a degree of only one, meaning they were only mentioned once in GPT-4 replies.

\begin{figure}[h]%
\centering
\includegraphics[width=0.85\textwidth]{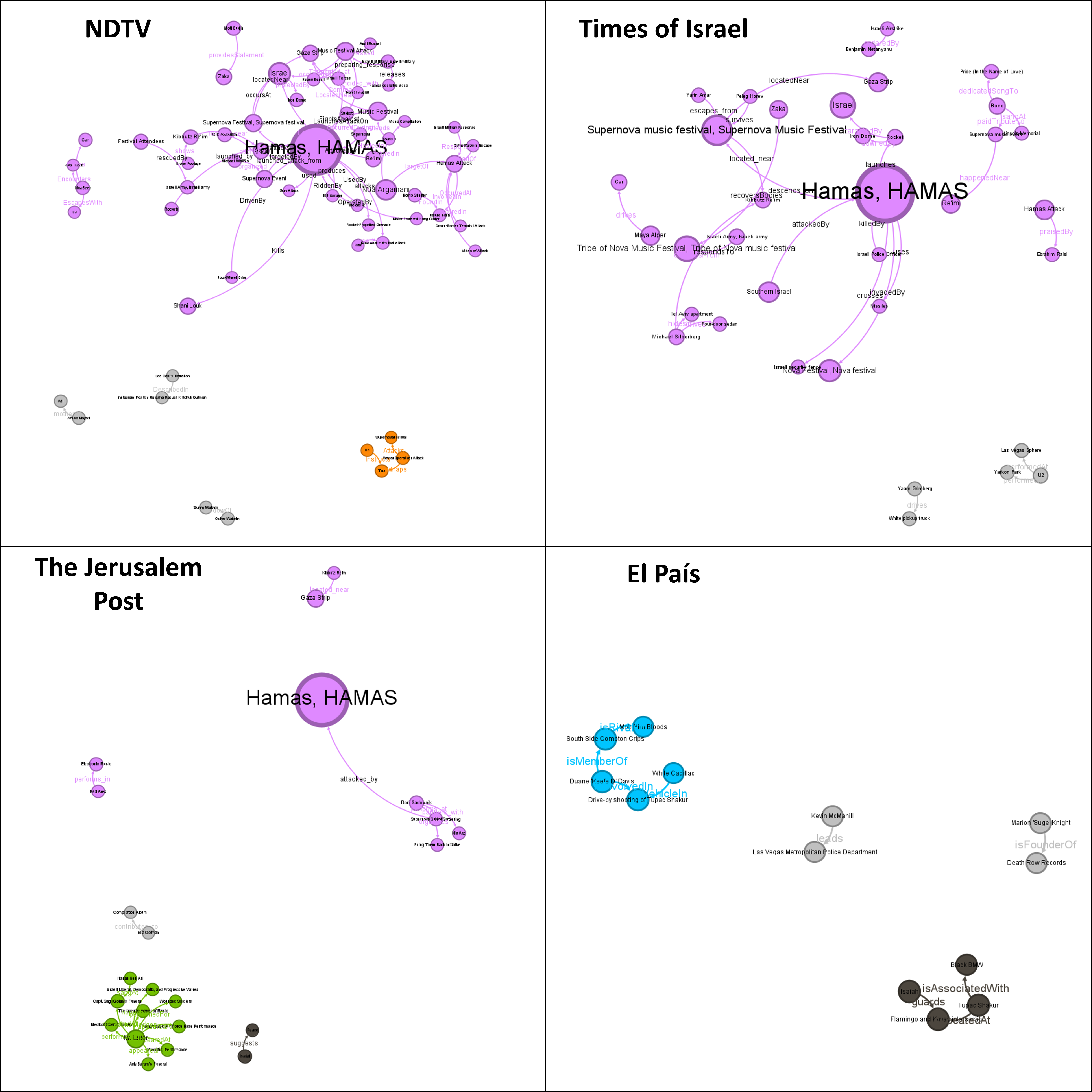}
\caption{Domain ontologies of t138}\label{fig15}
\end{figure}

For comparing domain ontologies, the degree of the nodes was not recalculated to ease comparison and orientation—still, different nodes of the same semantic proved to be a challenge. Multiple nodes of the same semantic made it more complex to analyze differences and commonalities of the domain ontologies. Nevertheless, the ontologies offered insight into relevant aspects or facts of t138. The domain ontologies of NDTV and Times of Israel illustrate this in Fig. \ref{fig15} (upper row).

It can be seen that both mention the same locations, e.g. "Israel", "Gaza Strip", "Re'im", both mention the Iron Dome of Israel and how it protects Israel, and both describe the Hamas attack. Also, both newspapers wrote about the individual fates of attendees of the Supernova Sukkot Gathering. However, they wrote about different people. While NDTV mentioned more people who were killed or abducted, Times of Israel mentioned more people who survived the attack. Both newspapers also wrote about Israeli actions following the attack, but here, nodes and relations are disjoint. Times of Israel wrote about an Israeli airstrike ordered by Benjamin Netanyahu, and NDTV wrote about the Israeli military preparing a response to the Music Festival attack. Times of Israel mentioned two aspects that NDTV did not mention: music performances dedicated to victims of the attack and the praise of the attack by an Iranian politician.

The domain ontology of The Jerusalem Post represented the differences in the article content compared to other newspapers. While a few nodes were explicitly about the attack, most were about more positive aspects. Here, the ontology succeeded in explaining the differences of The Jerusalem Post in the media bias spectrum of t138. On the other hand, the domain ontology of El País showed that El País did not write about the Hamas attack, meaning the article was misplaced in t138. It also was a valuable result as it further explained the media bias spectrum of t138 and highlighted a mistake in the topic. Both domain ontologies are shown in the lower row in Fig. \ref{fig15}. 

Overall, for the articles in t138, ontology learning was successful. The GPT-4 results had a quality that enabled a comparison of omitted or committed facts between different newspapers. However, the hierarchical nature of objects remains unused. The list of objects per class could have been used to remove semantic duplicates in the nodes, which would have been beneficial for analyzing the domain ontologies. The name most used for a node by the number of times mentioned would provide a statistical method to determine the name. However, multiple mentions in the same newspaper would have to count as one to prevent bias from spreading from newspapers with more articles.

\subsection{Third Belt and Road Forum for International Cooperation and Putin's visit}\label{subsec44}

The topic \textit{"751\_xi\_putin\_china\_bri\_beijing"} (t751) is on level two of the topic hierarchy. It consists of three topics: two topics are about Putin's visit to Beijing, China during the Third Belt and Road Forum for International Cooperation (BRI), which was held between October 17 and 18, 2023. The other topic is about the BRI itself. 26 different newspapers published in total 139 articles on this topic, dating between October 4 and 29. 

\begin{figure}[h]%
\centering
\includegraphics[width=0.8\textwidth]{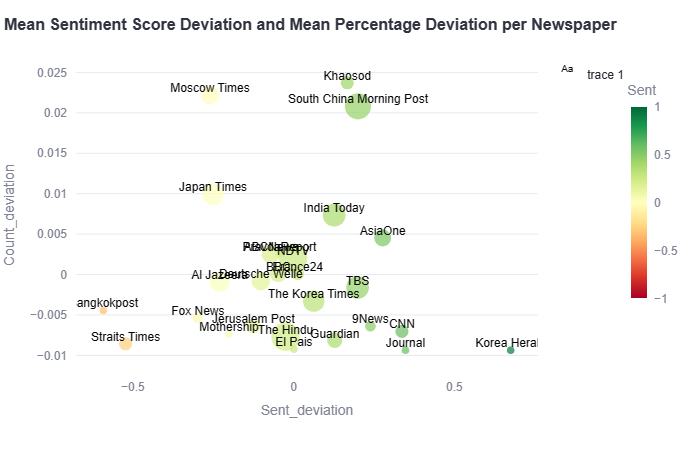}
\caption{Media bias spectrum of t751 based on RoBERTa title sentiment score}\label{fig16}
\end{figure}

Firstly, the RoBERTa title sentiment score was examined and visualized. Fig. \ref{fig16} shows the corresponding media bias spectrum, the underlying data can be found in Table \ref{tabA5} in Appendix \ref{secA3}. Three newspapers (South China Morning Post, Moscow Times, and Khaosod) published around 2\% more articles relative to their total number of articles than average, indicating higher importance of this topic for those newspaper. The sentiment scores were again checked by manually reading the titles. While most of the scores were found to represent the sentiment of titles well, Moscow Times had a title that was mispresented: "Putin Arrives in Beijing Seeking Diplomatic, Economic Support from ‘Dear Friend’ Xi", RoBERTa seemed not to understand the negative semantic induced by the quotation marks. 

In the text bodies, no significant differences were found by spaCy. The corresponding data can be found in Table \ref{tabA6} in Appendix \ref{secA3}. Like in the first case study, spaCy failed to grasp the overall sentiment of the texts and is unsuited to determining labeling and word choice bias.

\begin{figure}[h]%
\centering
\includegraphics[width=0.7\textwidth]{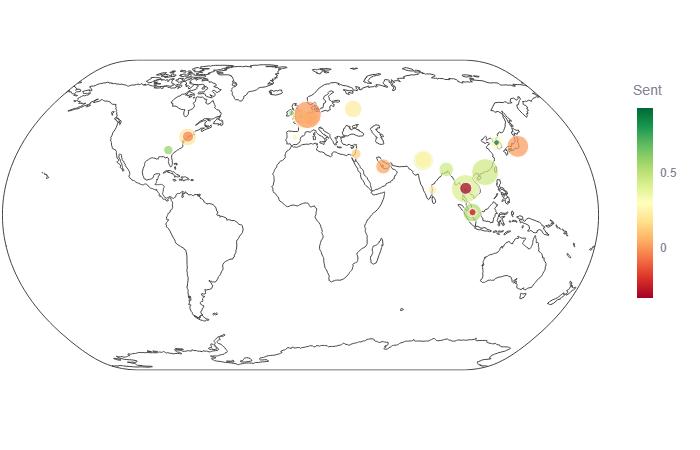}
\caption{RoBERTA title sentiment score and publishing rate per newspaper on world map}\label{fig17}
\end{figure}

By plotting the data on a map, as described in Section \ref{subsec35}, it became visible that the newspapers with the highest publishing rates were located in southeast Asia, as seen in Fig. \ref{fig17}. 

The one exception is the Moscow Times in Europe. However, the Moscow Times used to have its headquarters in Moscow until 2022. While for this topic, event selection bias is regionally dependent, labeling and word choice bias is not, as the average sentiment score in Southeast Asia varies, highlighted through the colors in Fig. \ref{fig17}.

Since t751 is a hierarchical topic, more than ten entities were analyzed. The top ten entities of all base-level topics that are part of t751 are shown in Table \ref{tab5}. However, not all entities have been analyzed in each topic; they were only analyzed in the topics in which they were the top entities. Here, there are different entities with the same meaning: "Xi" and "Xi Jinping" as well as "Belt and Road Initiative" and "BRI". For a more accurate evaluation of those entities, they must be merged.

\begin{table}[h]
\caption{Top entities of t751 sorted by "Average target-dependent sentiment score"}\label{tab5}%
\begin{tabular}{@{}lllllll@{}}
\toprule
Entity & Average target-dependent & Average target & Count\\
& \quad sentiment score & \quad neutrality score & \\
\midrule
Israel & -0.4987 & 0.3609 & 17\\
Ukraine & -0.3067 & 0.5865 & 24\\
Russia & -0.2916 & 0.3657 & 26\\
ICC & -0.1941 & 0.7177 & 8\\
Moscow & -0.1724 & 0.4581 & 21\\
Russian & -0.0986 & 0.5114 & 26\\
Vietnam & 0.0031 & 0.8077 & 5\\
Beijing & 0.0086 & 0.6982 & 26\\
Putin & 0.0227 & 0.4064 & 24\\
Thailand & 0.0801 & 0.7927 & 6\\
China & 0.0936 & 0.5256 & 25\\
Pakistan & 0.1072 & 0.6882 & 6\\
EU & 0.2194 & 0.6792 & 8\\
Africa & 0.2485 & 0.6464 & 14\\
Chinese & 0.251 & 0.46 & 24\\
Xi & 0.2768 & 0.4768 & 24\\
BRI & 0.2773 & 0.5397 & 13\\
Belt and Road Initiative & 0.4484 & 0.4117 & 10\\
Xi Jinping & 0.4545 & 0.3198 & 12\\
\botrule
\end{tabular}
\end{table}

\begin{figure}[h]%
\centering
\includegraphics[width=0.8\textwidth]{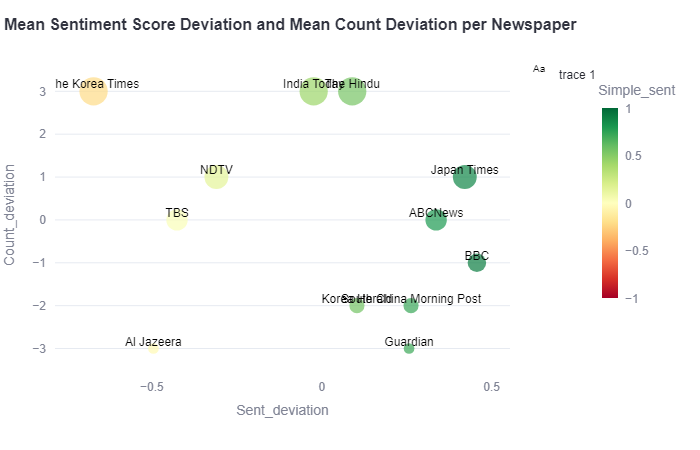}
\caption{Media bias spectrum of the entity "Xi Jinping" in t751}\label{fig18}
\end{figure}

Even though t751 is about the BRI, the entities "Ukraine" and "Israel" are among the top entities, showcasing the impact of the related events on political news. Again, in most cases, the average target neutrality score is higher than each entity's average target-dependent neutrality score in absolute terms. One of the exceptions was the entity "Xi Jinping". The corresponding media bias spectrum is shown in Fig. \ref{fig18}. The plot shows three newspapers writing more about him and highlights a difference in sentiments, especially between The Korea Times and India Today or The Hindu. Those three are good to compare since they have the same number of mentions of "Xi Jinping". The underlying data of Fig. \ref{fig18} can be found in Table \ref{tabA7} in Appendix \ref{secA3}.

As most articles of t751 date after the Hamas attack, the target-dependent sentiment tended to be negative for the entity "Israel", as shown in Fig. \ref{fig19}. The underlying data can be found in Table \ref{tabA8} in Appendix \ref{secA3}. 

\begin{figure}[h]%
\centering
\includegraphics[width=0.8\textwidth]{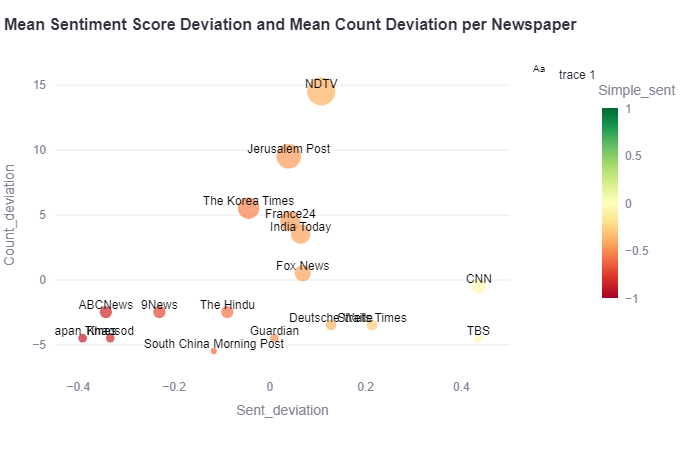}
\caption{Media bias spectrum of the entity "Israel" in t751}\label{fig19}
\end{figure}

This result was found to be valid, since the mention of Israel in articles of t751 was usually in the context of a war in which they are said to participate in some form. To illustrate this, here are some examples:

\begin{itemize}
\item \textbf{NDTV:} "And global headlines will be dominated by Israel's war with Palestinian militant organisation Hamas."
\item \textbf{FOX News:} "Russia and China have forged an informal alliance against the United States and other democratic nations that is now complicated by the Israel-Hamas war."
\item \textbf{The Jerusalem Post:} "Russian statements have slammed Israel for the explosion at a hospital in Gaza."
\end{itemize}

While RoBERTa correctly captures the negative nature in the sentence context of the entity, it becomes evident that this is not enough. For example, the following sentence in the cited article of The Jerusalem Post is: "Without waiting for evidence, Moscow sought to exploit the tragedy, much as Iran sought to do.", a sentence that defends Israel and hence represents a more positive sentiment towards Israel. It illustrates that sometimes more context than one sentence is needed to determine the sentiment towards an entity correctly.

\begin{figure}[h]%
\centering
\includegraphics[width=0.9\textwidth]{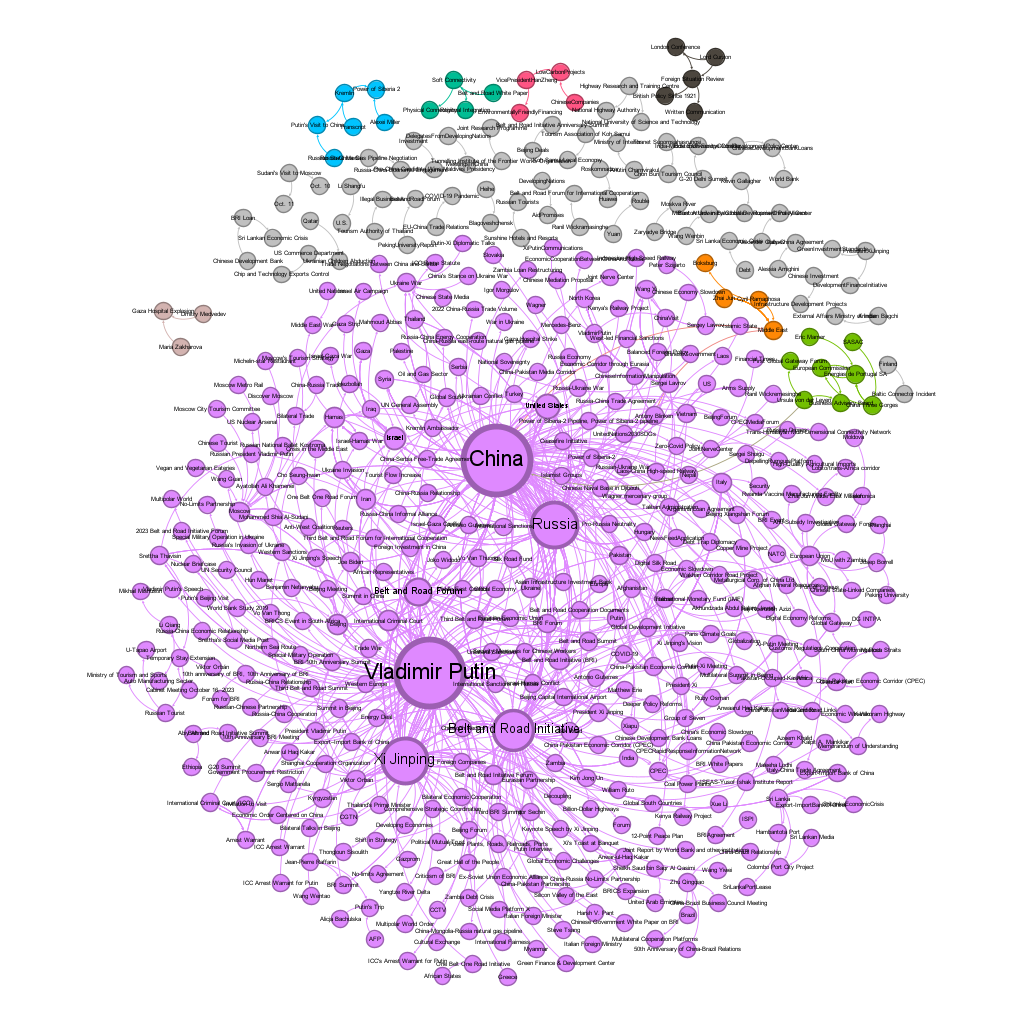}
\caption{Core reference ontology of t751}\label{fig20}
\end{figure}

Next, the created ontologies were analyzed. While object-class and object-object consistence were high with $\approx$99.7\% and 100\%, the object-relation consistence was lower, measuring $\approx$83.5\%. 232 out of 1414 relations failed the consistency check. One additional relation was removed since it was in a bad format. The core reference ontology of t751, representing all articles within it, is shown in Fig. \ref{fig20}. 

Again, nodes were sized based on their degree and colored by their inferred classes. The issue of different nodes with the same semantic was less pronounced here, yet it persisted. In addition, due to the higher number of articles, the core reference ontology was more complex and, hence, more difficult to analyze. Also, assorted communities did not prove to be valuable. For example, the assorted community of the orange-colored nodes is about the middle east. They appear in the domain ontologies of South China Morning Post, Al Jazeera, Fox News, France24, India Today, Jerusalem Post, NDTV, and Korea Times. If the topic of the Middle East had only occurred in this community, this would have been a valuable insight and indicate an aspect of the topic that only some newspapers mention. However, multiple nodes in the Middle East were in the most prominent, purple-colored community.

\subsection{Cross-topic analysis}\label{subsec45}

\begin{figure}[h]%
\centering
\includegraphics[width=0.7\textwidth]{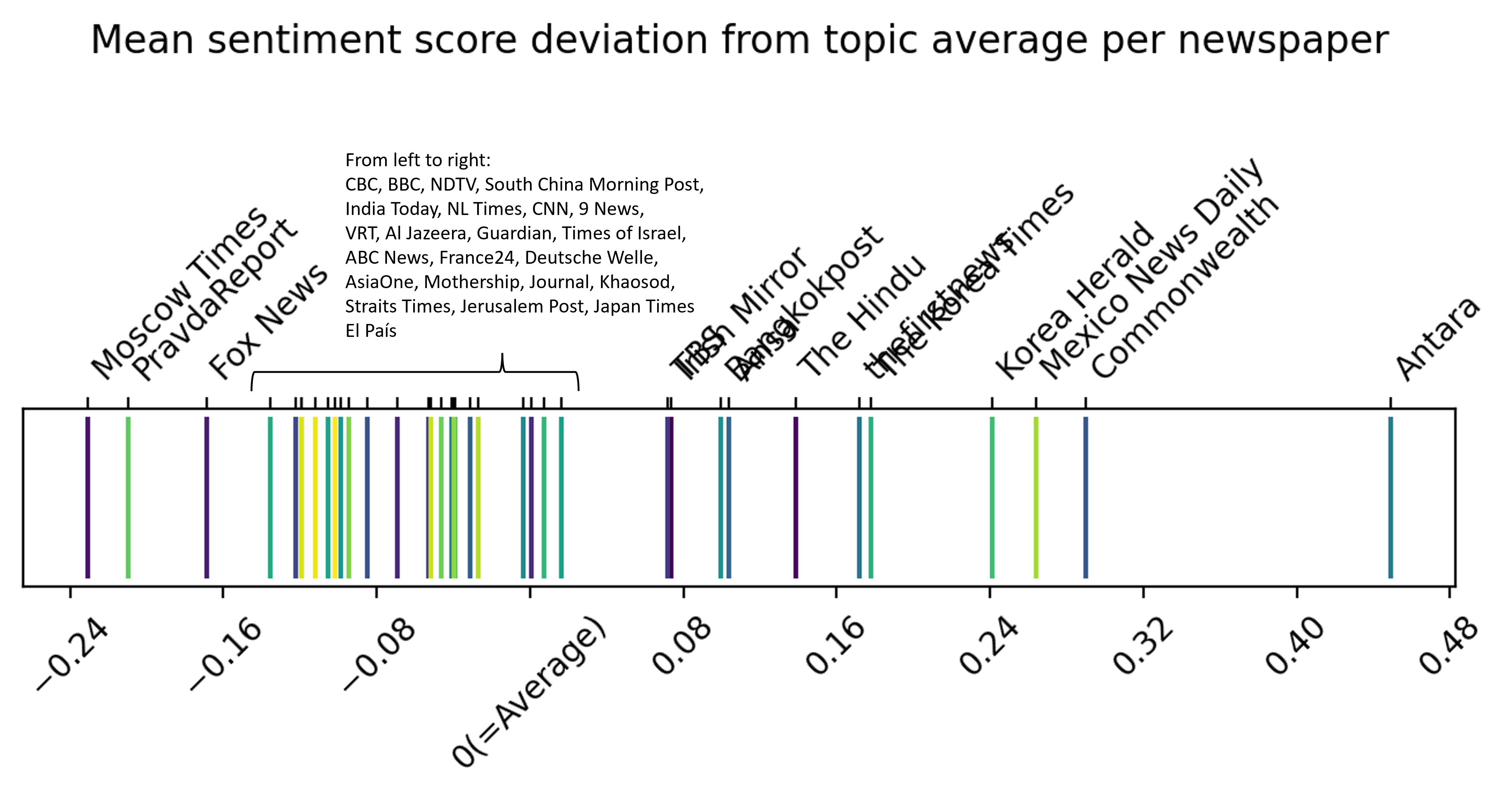}
\caption{RoBERTa title sentiment deviation across all articles}\label{fig21}
\end{figure}

As mentioned in Section \ref{subsec42}, the ontology creation was not done for the whole dataset. Also, the proposed measure for event selection bias is not eligible when looking at all articles of the dataset since it will always be 100\%. However, the labeling and word choice bias was examined for the whole dataset. The average spaCy text sentiment score across all newspapers was $\approx$0.05. The largest deviation to this score in absolute terms was the average sentiment score from Irish Mirror with an amount of 0.0572. Due to this small deviation, no further analysis was conducted, but the RoBERTa title sentiment score was analyzed. The average across all newspaper was $\approx$-0.06, so again close to completely neutral. However, the deviation showed some more significant differences, as shown in Fig. \ref{fig21}. While Antara has clearly more positive titles than average, e.g. Moscow Times and PravdaReport have more negative titles than average. The underlying data can be found in Table \ref{tabA9} in Appendix \ref{secA4}.

\begin{figure}[h]%
\centering
\includegraphics[width=0.8\textwidth]{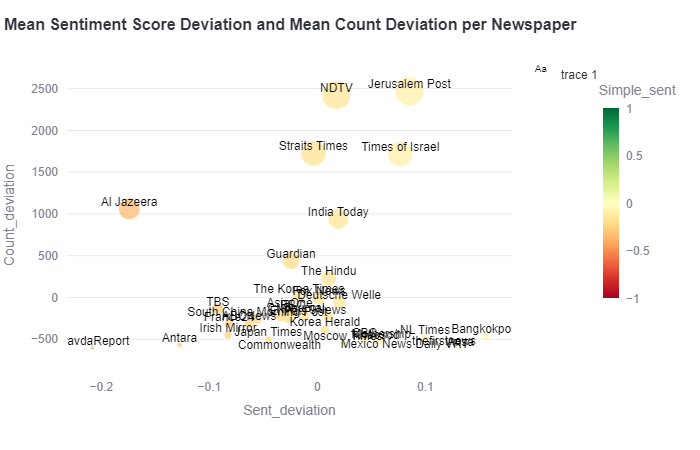}
\caption{Media bias spectrum of the entity "Israel" in whole dataset}\label{fig22}
\end{figure}

Also, analyzing entities in the whole dataset is possible. As mentioned, not all entities have been analyzed in each topic; they were only analyzed in the topics in which they were the top entities. For example, Fig. \ref{fig22} shows the media bias spectrum of the entity "Israel". Of the newspapers that mention this entity more often than average, The Jerusalem Post and Times of Israel have the most positive sentiment towards Israel, which is plausible. It is also striking that Al Jazeera has a more negative sentiment towards "Israel" than average. The underlying data can be found in Table \ref{tabA10} in Appendix \ref{secA4}.

\section{Results and Discussion}\label{sec5}

This section provides a summary of our case study findings and an assessment of the methodology we have used. At first, the case study results are summarized. The results are of two kinds: the media bias analysis of the topic, meaning the findings in event selection, labeling and word choice, and commission and omission bias. These results are only valid for the given dataset and not generalized. The other kind is the implications of the first kind of results on whether the proposed methodology is suitable to analyze media bias in articles. These results are then abstracted and generalized. Second, the proposed methodology is compared to other studies. Lastly, we outline the limitations of our research and show potential areas for future research.

\subsection{Case study results}\label{subsec51}

In the first case study in Section \ref{subsec43}, articles about the Hamas attack on Israel on October 7, 2023, were analyzed. The media bias spectrum of the titles showcases that the average title sentiment of different newspapers spans a wide range, starting from $\approx$-0.87 for CBC and ending at $\approx$0.92 for The Jerusalem Post. It is caused by either varying emotionality in titles or the mention of different aspects of the attack. The RoBERTa model proved to be well-suited for sentiment analysis. However, the spaCy sentiment analysis of the text bodies was found to hold no valuable result. The entity analysis showed that all newspapers wrote negatively about "Hamas", resulting in an overall average target-dependent sentiment score of $\approx$-0.9199. The sentiment towards the entity "Israeli" on the other side was more diverse, and the Jerusalem Post was the most positive newspaper. It showcases how the proposed methodology can capture the most relevant entities of a topic and highlight similarities and differences in newspapers writing about them. The created ontologies enabled a comparison of omitted or committed facts first by highlighting completely different clusters through assortative communities, which allows the detect different aspects mentioned at first glance due to the node color. For instance, the domain ontology of Jerusalem Post revealed that while writing about the same topic, The Jerusalem Post mentioned entirely different facts or aspects. They hardly wrote about the attack and instead focused on music performances in memory of the victims. And secondly, through enabling detailed comparisons of nodes. NDTV and the Times of Israel had most nodes in the same assortative community and shared many nodes; however, NDTV mentioned more people who were killed or abducted, while Times of Israel mentioned more people who survived the attack. Hence, ontology learning provided valuable insights into omission and commission bias. However, different nodes of the same semantic meaning compromised the quality of the ontologies.

In the second case study in Section \ref{subsec44}, articles about the BRI in 2023 were analyzed. The topic is on level two of the topic hierarchy. Nevertheless, all further steps were conducted as before, showcasing the proposed methodology’s capability to climb the hierarchy tree of topics up and down, with no additional steps needed. The media bias spectrum of the titles indicates that newspapers write relatively positively about this topic, with an overall average title sentiment score of $\approx$0.26. It highlights that three newspapers publish relatively more articles about this topic than others, providing an excellent example of how event selection bias is presented in the media bias spectrum. Again, the sentiment analysis of the text bodies with spaCy proved insufficient to grasp their sentiment. The entity analysis revealed that even though the topic is about the BRI, the conflicts in Ukraine and Israel are influenced, as they are often mentioned in the articles. The target-dependent sentiment score was again represented well in the media bias spectrums of "Xi Jinping" and "Israel". However, it was found that sometimes the given context of one sentence is not always enough context. For commission and omission, the issue of different nodes with the same semantic meaning combined with more articles on this topic resulted in the ontologies being more prominent and more complex than the ones of the first case study. Here, the proposed methodology failed to provide insights into omission and commission bias.

The third and last case study in Section \ref{subsec45} analyzed all articles except those on the noise topic. It showed the differences in average title sentiment scores across newspapers, ranging from the Moscow Times with a score of $\approx$-0.29 to Antara with a score of $\approx$0.39. This result explores the overall labeling and word choice bias of newspapers. The spaCy sentiment analysis held no valuable insights. The media bias spectrum of the entity "Israel" shows that Al Jazeera has a more negative sentiment on average toward this entity than most newspapers while mentioning it more than average. The possibility of conducting entity analysis across all topics is valuable since it removes the context of a unique topic and analyses the entity mentioned in general. As omission and commission bias are news event-specific per definition, a cross-topic analysis is not sensible.

\subsection{Case study implications on the proposed methodology}\label{subsec52}

The methodology proposed herein allows us to successfully analyze the potential media bias in event selection within news. The proportion of articles on a specific topic within the total output of a newspaper is an empirical fact independent of human bias. Comparing this ratio across different newspapers can reveal if a particular newspaper tends to write more or less about certain news events. The KPop Topic shown in Fig. \ref{fig6} is a good example for this: Korea Herald and The Korea Times wrote the most about it. However, unsupervised learning methods like BERTopic produce mistakes, as shown in the first case study in Sections \ref{subsec43} and hence, there is an inaccuracy in the resulting scores. 

The suggested approach also proves suitable for analyzing labeling and word choice media bias regarding sentiment. Especially sentiment in article titles and towards entities was found to represent this form of media bias well. Independent of the topic and entity, the results in the first and the second case studies represented the articles well. Besides the plain sentiment score, comparing different newspapers helps classify the labeling and word choice bias across them and sets it into perspective. This enables viewers to understand the range of views, for instance, negative and positive sentiment scores towards Xi Jinping in Section \ref{subsec44}, or no variations in sentiment scores like towards Hamas in Section \ref{subsec43}. However, the sentiment analysis of the main text body yielded no valuable insights. This limitation was noticeable in all three case studies. The primary scoring method employed by spaCy fails to comprehend the nuanced meanings of words in their contextual settings and fails to capture the subtleties present "between the lines". Nevertheless, given the necessary resources, a state-of-the-art LLM that can take more tokens as input than RoBERTa could do a valuable sentiment analysis, as RoBERTa did for the titles. The proposed methodology allows to freely interchange models and analyzed documents, hence even offering comparisons of different models and article elements.

The media bias spectrum we present in this study offers a practical approach for scrutinizing and contrasting the media bias exhibited by various newspapers. While it does rely on sentiment classification, which usually takes labeled data as input, it does not label data as "biased" or not, as it relies on comparing the scores of multiple newspapers. Hence there is no inherent bias in the methodology itself, although the sentiment classification models are prone to bias. This effect can, however, be reduced by comparing the results of different models. The media bias spectrum visualizes two types of media bias, event selection as well as labeling and word choice, and can be applied to certain aspects or entities as well as article titles and text bodies. It is also universally applicable to any dataset in contrast to former qualitative methods. 

The commission and omission could not be consistently and successfully analyzed using the proposed methodology. While prompting Sequence-to-Sequence models for creating ontologies worked well, the different notations of the same entities resulted in too many nodes with the same semantic, which gets worse as more articles are analyzed, as shown in Section \ref{subsec44}. Nevertheless, ontologies offer three key advantages: First, they abstract objects into classes, enabling comparisons of objects mentioned per class. Second, they show the relations between the objects and how they depend on newspapers. Third, filtering a core reference ontology on newspapers or articles allows for meaningful and comprehensive comparisons, as the first case study indicates.

\subsection{Comparison with different studies}\label{subsec53}

For analyzing event selection bias, GDELT was not used in contrast to \cite{Bourgeois2018, Saez-Trumper2013}. Instead, news events were retrieved from the dataset itself. Although this approach might omit some topics, it is not a concern as our method measures media bias through comparison, rendering unrepresented topics in the dataset irrelevant. In \cite{Hamborg2020} and \cite{Hamborg2017}, LDA was used for topic modeling. BERTopic is the more state-of-the-art approach to topic modeling and performs better \cite{Gan2024}, and it adds increased comparison capabilities through the hierarchical nature of the topics. \cite{Hamborg2020} focuses on news topics country-wise, and also \cite{Bourgeois2018} analyzed their clusters from a geographical point of view. The result of \cite{Bourgeois2018} would allow displaying different biases and opinions in the same country, contrary to \cite{Hamborg2020}. The second case study in Section \ref{subsec44} shows that our methodology offers a regional comparison of bias and allows for different biases and opinions in the same country. \cite{Hamborg2020} offers a user interface, however, it has a different aim, hence the comparison to the media bias spectrum is not eligible.

To our knowledge, commission and omission bias have not been analyzed using ontologies. Ontologies, as they are used here, offer advantages compared to aspect-based approaches like \cite{Park2009} and \cite{Ehrhardt2021}. In \cite{Ehrhardt2021}, entities were extracted, which are similar to objects in the ontologies. However, these only represent a single element of the formed ontologies instead of objects along with their class and relations and, therefore, carry less information. The entity analysis of our proposed ontology, on the other hand, not only covers the entity extraction done in \cite{Ehrhardt2021}, but also enables comparisons of entity occurrences in various newspapers by topic. Still, this method is viewed as not thorough enough when compared to ontologies. Aspects as they are used in \cite{Park2009} contain an "agent", an acting entity, and an "action" done by the agent. This aspect definition can best be compared with one object, an object pair, or sometimes even an object group, along with their ontological relationships. Ontologies cover this approach and improve it by allowing more than one agent in an aspect and more than one relation of an aspect. They also show how different aspects come into play with each other.

Labeling and word choice bias is often times analyzed using sentiment classification \cite{Hamborg2019}. While this is dependent on pre-labeled datasets, it is less exposed to human bias than classification datasets which explicitly label bias information such as left-right-bias. In \cite{Hamborg2019a}, a dataset and model for a more sophisticated classification of labeling and word choice bias than sentiment classification is introduced, with classes such as "Easiness", "Reason" or "Honor". They also enable the analysis of targets. It poses a valuable addition to the proposed methodology. However, it cannot replace standard sentiment analysis since a two-dimensional attribute is necessary in the media bias spectrum. The media bias spectrum could be plotted for each class of the classification model introduced in \cite{Hamborg2019a}. The methodology of D'Alonzo and Tegmark requires no pre-labeled data, instead, they propose an algorithm that auto-discovers bias on two axes which reflect left-right bias and establishment bias \cite{DAlonzo2022}. In terms of avoiding human intervention, their proposed methodology is superior to sentiment classification as our methodology relies on sentiment classification, which requires a pre-labeled dataset. Nevertheless, they had to manually purge and merge phrases in their work. Also, the aim of the work D'Alonzo and Tegmark is slightly different. They focus on different phrases for the same, one could argue the choice of the phrase depends on the sentiment towards it. For example, they find that in the context of US immigration, left-oriented newspapers use the phrase "undocumented immigrants", while right-oriented newspapers use the phrase "illegal immigrants" \cite{DAlonzo2022}. Our methodology differs in that, as the target-dependent sentiment towards the extracted entity "immigrants" would be analyzed in the context of their sentence. In the given example, the resulting score would be more negative for "illegal immigrants". Ultimately, different aspects of labeling and word choice bias are analyzed in \cite{DAlonzo2022}, hence the work provides a valuable addition to the proposed methodology.

\subsection{Limitations and future research}\label{subsec54}

As mentioned in Section \ref{sec4}, the results of the case studies are only valid for the given data set, which consists of articles on RSS feeds, as no other articles were included. RSS feeds frequently include only a fraction of all articles published by a newspaper. For a thorough comparison, every article analyzed by the respective newspaper must be included in the data set. Fundamentally, the significance of the methods depends on ample data and successful topic modeling.

To improve the media bias spectrum, incorporating a time dimension could reveal crucial insights, such as a general or newspaper-specific sentiment shift or a particular topic's emergence. Explainable AI methods could be used to improve the quality assurance of the results. Outlier detection algorithms could be applied to scan all topics for outliers, leading human analysts toward striking data points and increasing usability when analyzing a whole dataset. Additionally, we might consider using crosses instead of points in the media bias spectrum. It would represent the range of sentiment scores across different articles from the same newspaper.

To enhance our understanding of commission and omission bias, it is crucial to consolidate nodes that share the same semantic properties. However, this process can be labor-intensive if done manually, which necessitates the development of an automated method for combining these nodes.

Despite the quality of the ontology, visualizing these ontologies poses significant challenges. \textit{Gephi}, a tool designed for graph visualization, fails to present multiple relations between two nodes effectively. Moreover, it has not been used to depict object clusters, their relations, and their hierarchy. One potential solution could be to use the classes to assign colors to the nodes, but this approach risks losing information about assortative communities. Hence, there is an apparent demand for more innovative visualization techniques in the future. These would allow news consumers to comprehend better and engage with the data.

\section{Conclusion}\label{sec6}

In the beginning, the social relevance of media bias was highlighted, and that it is critical to illuminate biased news in order for citizens to get informed and democracies to flourish. Still, the central issue of the concept of bias remains - the fact that it is always relative, and there is no null value or baseline. Furthermore, a pluralism of opinions is even desirable to get the whole picture of news events. This balancing act of showing where news strays from the truth and accepting where they have different opinions is reflected in the proposed methodology. In acceptance of this tension, we have made three key contributions to identifying three forms of media bias.

The Media Bias Spectrum is an all-encompassing and user-friendly instrument for assessing two types of media bias: event selection bias and labeling and word choice bias. It allows users to evaluate newspapers on distinct levels, including entities, text content, and article headings. Given the societal significance of media bias, developing tools that communicate effectively with users is crucial. The Media Bias Spectrum fulfills this role, recognizing the relative nature of media bias and even capitalizing on it to illustrate the range of viewpoints on a particular subject or towards a specific entity. It can identify anomalies and quantify the extent of their divergence. In all three case studies, this tool provided valuable insights into the examined data set.

The utilization of hierarchical clustering of news topics allows the analysis of event selection bias and changes the scope of bias analysis to any level. It is possible to compare all articles altogether, deep dive into a small topic consisting of only a few articles, or anything in between. Thanks to the described methodology, the change in the context range can happen without additional steps.

Studying commission and omission bias through ontologies can give users a comprehensive view of a subject, making it easier to delve into specific areas for further analysis and comparisons. We have introduced consistency measures to ensure the reliability of results when prompting LLMs for ontologies. We have demonstrated the potential of this method through two case studies. However, the presence of nodes with identical meanings diminishes the utility of the ontologies, indicating a need for optimization. Moreover, enhancements in the visualization and user interaction with the results are necessary for society to realize the benefits fully. Looking ahead, there is considerable potential in further refining LLMs for ontology learning in media bias analysis and other areas.

\section*{Statements and Declarations}

\subsection*{Data availability}
The datasets analyzed during the current study are available from the corresponding author upon reasonable request.

\subsection*{Competing Interests}
The authors have no competing interests to declare that are relevant to the content of this article.









\pagebreak

\begin{appendices}

\section{Data of case study "Hamas attack on Israel"}\label{secA2}

\begin{table}[h]
\caption{Media bias spectrum of t138 based on RoBERTa title sentiment score, sorted by "Sent\_deviation"}\label{tabA1}%
\begin{tabular}{@{}lllllll@{}}
\toprule
Newspaper & Count\_deviation & Sent\_deviation & Part\_of\_total & Sent & Count\\
\midrule
CBC & 0.0042 & -0.5907 & 0.0085 & -0.8659 & 1\\
Mothership & -0.0009 & -0.5692 & 0.0034 & -0.8445 & 1\\
France24 & -0.0007 & -0.4524 & 0.0036 & -0.7276 & 1\\
India Today & 0.0057 & -0.3627 & 0.0101 & -0.6379 & 5\\
Al Jazeera & -0.0029 & -0.3422 & 0.0014 & -0.6174 & 1\\
NDTV & 0.003 & -0.096 & 0.0073 & -0.3713 & 8\\
9News & 0.0022 & -0.0772 & 0.0066 & -0.3525 & 3\\
El Pais & -0.0028 & -0.0601 & 0.0015 & -0.3353 & 1\\
Straits Times & -0.0036 & -0.0166 & 0.0007 & -0.2918 & 1\\
Fox News & 0.0039 & 0.0013 & 0.0082 & -0.274 & 3\\
Guardian & -0.0024 & 0.0386 & 0.002 & -0.2367 & 3\\
The Hindu & -0.0039 & 0.0577 & 0.0004 & -0.2175 & 2\\
Japan Times & -0.0017 & 0.1215 & 0.0026 & -0.1538 & 1\\
Journal & -0.0015 & 0.1378 & 0.0028 & -0.1375 & 2\\
BBC & -0.0021 & 0.1968 & 0.0022 & -0.0784 & 1\\
CNN & -0.0006 & 0.2054 & 0.0037 & -0.0698 & 3\\
South China & 0.0036 & 0.282 & 0.0079 & 0.0068 & 3\\
\quad Morning Post & & & & & \\
Times of Israel & 0.0025 & 0.3282 & 0.0069 & 0.053 & 5\\
Jerusalem Post & -0.0021 & 1.1978 & 0.0022 & 0.9225 & 2\\
\botrule
\end{tabular}
\end{table}

\begin{table}[h]
\caption{Media bias spectrum of t138 based on spaCy text sentiment score, sorted by "Sent\_deviation"}\label{tabA2}%
\begin{tabular}{@{}lllllll@{}}
\toprule
Newspaper & Count\_deviation & Sent\_deviation & Part\_of\_total & Sent & Count\\
\midrule
Japan Times & -0.0017 & -0.2401 & 0.0026 & -0.2411 & 1\\
The Hindu & -0.0039 & -0.1084 & 0.0004 & -0.1094 & 2\\
Journal & -0.0015 & -0.0497 & 0.0028 & -0.0507 & 2\\
India Today & 0.0057 & -0.0417 & 0.0101 & -0.0427 & 5\\
Mothership & -0.0009 & -0.0372 & 0.0034 & -0.0382 & 1\\
France24 & -0.0007 & -0.0303 & 0.0036 & -0.0313 & 1\\
Al Jazeera & -0.0029 & -0.0072 & 0.0014 & -0.0082 & 1\\
NDTV & 0.003 & -0.0046 & 0.0073 & -0.0057 & 8\\
South China & 0.0036 & -0.0043 & 0.0079 & -0.0053 & 3\\
\quad Morning Post & & & & & \\
9News & 0.0022 & 0.0157 & 0.0066 & 0.0147 & 3\\
CNN & -0.0006 & 0.0164 & 0.0037 & 0.0153 & 3\\
BBC & -0.0021 & 0.0197 & 0.0022 & 0.0187 & 1\\
Times of Israel & 0.0025 & 0.0275 & 0.0069 & 0.0265 & 5\\
Fox News & 0.0039 & 0.0309 & 0.0082 & 0.0299 & 3\\
Guardian & -0.0024 & 0.0583 & 0.002 & 0.0573 & 3\\
CBC & 0.0042 & 0.0659 & 0.0085 & 0.0648 & 1\\
El Pais & -0.0028 & 0.0665 & 0.0015 & 0.0654 & 1\\
Straits Times & -0.0036 & 0.1021 & 0.0007 & 0.101 & 1\\
Jerusalem Post & -0.0021 & 0.1204 & 0.0022 & 0.1194 & 2\\
\botrule
\end{tabular}
\end{table}

\begin{table}[h]
\caption{Media bias spectrum of "Hamas" in t138 based on target-dependent sentiment score, sorted by "Sent\_deviation"}\label{tabA3}%
\begin{tabular}{@{}lllllll@{}}
\toprule
Newspaper & Simple\_sent & Count & Sent\_deviation & Count\_deviation\\
\midrule
Japan Times & -0.9685 & 1 & -0.0485 & -6.8235\\
CBC & -0.9668 & 6 & -0.0469 & -1.8235\\
Mothership & -0.9616 & 1 & -0.0416 & -6.8235\\
India Today & -0.9537 & 22 & -0.0337 & 14.1765\\
9News & -0.9531 & 8 & -0.0331 & 0.1765\\
Fox News & -0.9514 & 7 & -0.0315 & -0.8235\\
Al Jazeera & -0.9464 & 3 & -0.0264 & -4.8235\\
CNN & -0.9419 & 1 & -0.0219 & -6.8235\\
BBC & -0.9303 & 1 & -0.0103 & -6.8235\\
France24 & -0.9198 & 1 & 0.0001 & -6.8235\\
Journal & -0.9144 & 4 & 0.0056 & -3.8235\\
South China & -0.9131 & 6 & 0.0069 & -1.8235\\
\quad Morning Post & & & & \\
Times of Israel & -0.9114 & 12 & 0.0086 & 4.1765\\
NDTV & -0.9109 & 42 & 0.0091 & 34.1765\\
Jerusalem Post & -0.8554 & 2 & 0.0646 & -5.8235\\
Guardian & -0.8373 & 12 & 0.0826 & 4.1765\\
The Hindu & -0.8034 & 4 & 0.1166 & -3.8235\\
\botrule
\end{tabular}
\end{table}

\begin{table}[h]
\caption{Media bias spectrum of "Israeli" in t138 based on target-dependent sentiment score, sorted by "Sent\_deviation"}\label{tabA4}%
\begin{tabular}{@{}lllllll@{}}
\toprule
Newspaper & Simple\_sent & Count & Sent\_deviation & Count\_deviation\\
\midrule
Mothership & -0.4154 & 3 & -0.3394 & -1.8824\\
Journal & -0.3581 & 4 & -0.2821 & -0.8824\\
South China & -0.1898 & 4 & -0.1138 & -0.8824\\
\quad Morning Post & & & & \\
9News & -0.1621 & 9 & -0.0861 & 4.1176\\
Guardian & -0.1521 & 6 & -0.0761 & 1.1176\\
NDTV & -0.1484 & 11 & -0.0724 & 6.1176\\
BBC & -0.0942 & 1 & -0.0182 & -3.8824\\
India Today & -0.0795 & 9 & -0.0035 & 4.1176\\
Fox News & -0.0711 & 7 & 0.0049 & 2.1176\\
Al Jazeera & -0.0652 & 7 & 0.0108 & 2.1176\\
The Hindu & -0.0567 & 4 & 0.0193 & -0.8824\\
Straits Times & -0.0412 & 2 & 0.0348 & -2.8824\\
Times of Israel & -0.0365 & 7 & 0.0395 & 2.1176\\
CBC & -0.0291 & 1 & 0.0469 & -3.8824\\
France24 & 0.0061 & 1 & 0.0821 & -3.8824\\
Japan Times & 0.1164 & 1 & 0.1924 & -3.8824\\
Jerusalem Post & 0.4848 & 6 & 0.5608 & 1.1176\\
\botrule
\end{tabular}
\end{table}

\afterpage{\clearpage}
\pagebreak

\section{Data of case study "Thirt Beld and Road Forum"}\label{secA3}

\begin{table}[h]
\caption{Media bias spectrum of t751 based on RoBERTa title sentiment score, sorted by "Sent\_deviation"}\label{tabA5}%
\begin{tabular}{@{}lllllll@{}}
\toprule
Newspaper & Count\_deviation & Sent\_deviation & Part\_of\_total & Sent & Count\\
\midrule
Bangkokpost & -0.0044 & -0.5918 & 0.0063 & -0.3368 & 1\\
Straits Times & -0.0086 & -0.5225 & 0.0022 & -0.2675 & 3\\
Fox News & -0.0053 & -0.298 & 0.0054 & -0.043 & 2\\
Moscow Times & 0.0222 & -0.2601 & 0.033 & -0.0051 & 6\\
Japan Times & 0.0099 & -0.2505 & 0.0207 & 0.0045 & 8\\
Al Jazeera & -0.0009 & -0.2299 & 0.0098 & 0.0251 & 7\\
Mothership & -0.0074 & -0.201 & 0.0034 & 0.054 & 1\\
Jerusalem Post & -0.0064 & -0.1251 & 0.0044 & 0.1299 & 4\\
Deutsche Welle & -0.0008 & -0.1021 & 0.01 & 0.1529 & 6\\
ABCNews & 0.0025 & -0.0709 & 0.0133 & 0.1841 & 6\\
BBC & 0.0001 & -0.0479 & 0.0109 & 0.2071 & 5\\
PravdaReport & 0.0026 & -0.0442 & 0.0133 & 0.2108 & 1\\
The Hindu & -0.0077 & -0.0253 & 0.0031 & 0.2297 & 14\\
NDTV & 0.0021 & -0.0012 & 0.0128 & 0.2538 & 14\\
El Pais & -0.0092 & 0.001 & 0.0015 & 0.256 & 1\\
France24 & 0.0001 & 0.0126 & 0.0109 & 0.2676 & 3\\
The Korea Times & -0.0033 & 0.0625 & 0.0075 & 0.3175 & 8\\
India Today & 0.0073 & 0.1259 & 0.0181 & 0.3809 & 9\\
Guardian & -0.0082 & 0.1279 & 0.0026 & 0.383 & 4\\
Khaosod & 0.0237 & 0.1669 & 0.0345 & 0.4219 & 3\\
TBS & -0.0016 & 0.1988 & 0.0092 & 0.4538 & 9\\
South China & 0.0208 & 0.2 & 0.0316 & 0.455 & 12\\
\quad Morning Post & & & & & \\
9News & -0.0064 & 0.2388 & 0.0044 & 0.4938 & 2\\
AsiaOne & 0.0045 & 0.2769 & 0.0153 & 0.5319 & 5\\
CNN & -0.007 & 0.3367 & 0.0037 & 0.5917 & 3\\
Journal & -0.0094 & 0.3477 & 0.0014 & 0.6027 & 1\\
Korea Herald & -0.0094 & 0.6749 & 0.0014 & 0.9299 & 1\\
\botrule
\end{tabular}
\end{table}

\begin{table}[h]
\caption{Media bias spectrum of 751 based on spaCy text sentiment score, sorted by "Sent\_deviation"}\label{tabA6}%
\begin{tabular}{@{}lllllll@{}}
\toprule
Newspaper & Count\_deviation & Sent\_deviation & Part\_of\_total & Sent & Count\\
\midrule
Fox News & -0.0053 & -0.0854 & 0.0054 & -0.0221 & 2\\
Jerusalem Post & -0.0064 & -0.043 & 0.0044 & 0.0203 & 4\\
El Pais & -0.0092 & -0.0255 & 0.0015 & 0.0379 & 1\\
Bangkokpost & -0.0044 & -0.0227 & 0.0063 & 0.0406 & 1\\
AsiaOne & 0.0045 & -0.0215 & 0.0153 & 0.0418 & 5\\
9News & -0.0064 & -0.0137 & 0.0044 & 0.0496 & 2\\
Moscow Times & 0.0222 & -0.0092 & 0.033 & 0.0541 & 6\\
South China & 0.0208 & -0.0087 & 0.0316 & 0.0546 & 12\\
\quad Morning Post & & & & & \\
ABCNews & 0.0025 & -0.0082 & 0.0133 & 0.0551 & 6\\
Mothership & -0.0074 & -0.0051 & 0.0034 & 0.0582 & 1\\
Journal & -0.0094 & -0.0042 & 0.0014 & 0.0591 & 1\\
Japan Times & 0.0099 & 0.0003 & 0.0207 & 0.0636 & 8\\
CNN & -0.007 & 0.0021 & 0.0037 & 0.0654 & 3\\
Khaosod & 0.0237 & 0.0035 & 0.0345 & 0.0668 & 3\\
Al Jazeera & -0.0009 & 0.0036 & 0.0098 & 0.0669 & 7\\
The Korea Times & -0.0033 & 0.0037 & 0.0075 & 0.067 & 8\\
The Hindu & -0.0077 & 0.0065 & 0.0031 & 0.0698 & 14\\
Guardian & -0.0082 & 0.0103 & 0.0026 & 0.0736 & 4\\
Deutsche Welle & -0.0008 & 0.0122 & 0.01 & 0.0755 & 6\\
TBS & -0.0016 & 0.0137 & 0.0092 & 0.077 & 9\\
NDTV & 0.0021 & 0.0173 & 0.0128 & 0.0806 & 14\\
Straits Times & -0.0086 & 0.0177 & 0.0022 & 0.081 & 3\\
BBC & 0.0001 & 0.0179 & 0.0109 & 0.0812 & 5\\
India Today & 0.0073 & 0.0184 & 0.0181 & 0.0817 & 9\\
PravdaReport & 0.0026 & 0.025 & 0.0133 & 0.0883 & 1\\
Korea Herald & -0.0094 & 0.0419 & 0.0014 & 0.1052 & 1\\
France24 & 0.0001 & 0.0532 & 0.0109 & 0.1165 & 3\\
\botrule
\end{tabular}
\end{table}

\begin{table}[h]
\caption{Media bias spectrum of "Xi Jinping" in t751 based on target-dependent sentiment score, sorted by "Sent\_deviation"}\label{tabA7}%
\begin{tabular}{@{}lllllll@{}}
\toprule
Newspaper & Simple\_sent & Count & Sent\_deviation & Count\_deviation\\
\midrule
The Korea Times & -0.2171 & 7 & -0.6717 & 3\\
Al Jazeera & -0.0408 & 1 & -0.4954 & -3\\
TBS & 0.0285 & 4 & -0.4261 & 0\\
NDTV & 0.1445 & 5 & -0.3101 & 1\\
India Today & 0.4307 & 7 & -0.0238 & 3\\
The Hindu & 0.5442 & 7 & 0.0897 & 3\\
Korea Herald & 0.5581 & 2 & 0.1035 & -2\\
Guardian & 0.7115 & 1 & 0.2569 & -3\\
South China Morning Post & 0.7172 & 2 & 0.2627 & -2\\
\quad Morning Post & & & & \\
ABCNews & 0.7912 & 4 & 0.3366 & 0\\
Japan Times & 0.8755 & 5 & 0.421 & 1\\
BBC & 0.9111 & 3 & 0.4566 & -1\\
\botrule
\end{tabular}
\end{table}

\begin{table}[h]
\caption{Media bias spectrum of "Israel" in t751 based on target-dependent sentiment score, sorted by "Sent\_deviation"}\label{tabA8}%
\begin{tabular}{@{}lllllll@{}}
\toprule
Newspaper & Simple\_sent & Count & Sent\_deviation & Count\_deviation\\
\midrule
Japan Times & -0.8896 & 2 & -0.3908 & -4.4706\\
ABCNews & -0.841 & 4 & -0.3422 & -2.4706\\
Khaosod & -0.8319 & 2 & -0.3331 & -4.4706\\
9News & -0.7292 & 4 & -0.2304 & -2.4706\\
South China Morning Post & -0.6156 & 1 & -0.1169 & -5.4706\\
\quad Morning Post & & & & \\
The Hindu & -0.5873 & 4 & -0.0886 & -2.4706\\
The Korea Times & -0.5431 & 12 & -0.0443 & 5.5294\\
Guardian & -0.489 & 2 & 0.0097 & -4.4706\\
Jerusalem Post & -0.4592 & 16 & 0.0395 & 9.5294\\
France24 & -0.456 & 11 & 0.0428 & 4.5294\\
India Today & -0.4342 & 10 & 0.0645 & 3.5294\\
Fox News & -0.4299 & 7 & 0.0689 & 0.5294\\
NDTV & -0.3913 & 21 & 0.1075 & 14.5294\\
Deutsche Welle & -0.371 & 3 & 0.1278 & -3.4706\\
Straits Times & -0.2853 & 3 & 0.2134 & -3.4706\\
TBS & -0.063 & 2 & 0.4357 & -4.4706\\
CNN & -0.0622 & 6 & 0.4365 & -0.4706\\
\botrule
\end{tabular}
\end{table}

\afterpage{\clearpage}
\pagebreak
\section{Data of case study "Cross-topic analysis"}\label{secA4}

\begin{table}[h]
\caption{RoBERTa title sentiment deviation in all articles, sorted by "Sent\_deviation"}\label{tabA9}%
\begin{tabular}{@{}lllllll@{}}
\toprule
Newspaper & Sent\_deviation & Sent\\
\midrule
Moscow Times & -0.2305 & -0.2935\\
PravdaReport & -0.2099 & -0.2729\\
Fox News & -0.1688 & -0.2318\\
CBC & -0.1356 & -0.1986\\
BBC & -0.1226 & -0.1856\\
NDTV & -0.1194 & -0.1824\\
South China Morning Post & -0.1191 & -0.1821\\
India Today & -0.1118 & -0.1748\\
NL Times & -0.1055 & -0.1685\\
CNN & -0.1018 & -0.1648\\
9News & -0.0986 & -0.1616\\
VRT & -0.0946 & -0.1576\\
Al Jazeera & -0.0849 & -0.1479\\
Guardian & -0.0694 & -0.1324\\
Times of Israel & -0.0532 & -0.1162\\
ABCNews & -0.052 & -0.115\\
France24 & -0.0464 & -0.1094\\
Deutsche Welle & -0.0406 & -0.1036\\
AsiaOne & -0.0397 & -0.1027\\
Mothership & -0.0391 & -0.1021\\
Journal & -0.0312 & -0.0942\\
Khaosod & -0.0272 & -0.0902\\
Straits Times & -0.0038 & -0.0668\\
Jerusalem Post & 0.0008 & -0.0622\\
Japan Times & 0.0071 & -0.0559\\
El Pais & 0.0165 & -0.0465\\
TBS & 0.0719 & 0.0089\\
Irish Mirror & 0.0739 & 0.0109\\
Bangkokpost & 0.0993 & 0.0364\\
Ansa & 0.1038 & 0.0408\\
The Hindu & 0.1386 & 0.0756\\
thefirstnews & 0.1719 & 0.1089\\
The Korea Times & 0.1779 & 0.1149\\
Korea Herald & 0.2412 & 0.1782\\
Mexico News Daily & 0.264 & 0.201\\
Commonwealth & 0.2901 & 0.2271\\
Antara & 0.4488 & 0.3858\\
\botrule
\end{tabular}
\end{table}

\begin{table}[h]
\caption{Media bias spectrum of "Israel" in all articles, based on target-dependent sentiment score, sorted by "Sent\_deviation"}\label{tabA10}%
\begin{tabular}{@{}lllllll@{}}
\toprule
Newspaper & Simple\_sent & Count & Sent\_deviation & Count\_deviation\\
\midrule
PravdaReport & -0.398 & 49 & -0.2084 & -605.9459\\
Al Jazeera & -0.3638 & 1,717 & -0.1742 & 1,062.0541\\
Antara & -0.3172 & 88 & -0.1276 & -566.9459\\
TBS & -0.2816 & 518 & -0.092 & -136.9459\\
Irish Mirror & -0.2723 & 207 & -0.0827 & -447.9459\\
France24 & -0.2713 & 341 & -0.0817 & -313.9459\\
ABCNews & -0.2528 & 346 & -0.0632 & -308.9459\\
South China Morning Post & -0.2453 & 394 & -0.0556 & -260.9459\\
Japan Times & -0.235 & 156 & -0.0453 & -498.9459\\
Commonwealth & -0.2247 & 8 & -0.0351 & -646.9459\\
CNN & -0.2239 & 434 & -0.0343 & -220.9459\\
El Pais & -0.2165 & 410 & -0.0269 & -244.9459\\
AsiaOne & -0.2153 & 503 & -0.0257 & -151.9459\\
Guardian & -0.214 & 1,095 & -0.0244 & 440.0541\\
BBC & -0.2128 & 483 & -0.0232 & -171.9459\\
The Korea Times & -0.2067 & 674 & -0.0171 & 19.0541\\
Journal & -0.2013 & 446 & -0.0117 & -208.9459\\
Straits Times & -0.1932 & 2,380 & -0.0036 & 1,725.0541\\
Fox News & -0.1879 & 652 & 0.0017 & -2.9459\\
Korea Herald & -0.1827 & 272 & 0.0069 & -382.9459\\
9News & -0.1797 & 425 & 0.01 & -229.9459\\
The Hindu & -0.1791 & 883 & 0.0105 & 228.0541\\
NDTV & -0.172 & 3,077 & 0.0176 & 2,422.0541\\
India Today & -0.1702 & 1,596 & 0.0194 & 941.0541\\
Deutsche Welle & -0.1695 & 602 & 0.0201 & -52.9459\\
Moscow Times & -0.1659 & 104 & 0.0237 & -550.9459\\
CBC & -0.1459 & 151 & 0.0437 & -503.9459\\
Khaosod & -0.1355 & 125 & 0.0542 & -529.9459\\
Mothership & -0.1305 & 134 & 0.0591 & -520.9459\\
Mexico News Daily & -0.1211 & 10 & 0.0685 & -644.9459\\
Times of Israel & -0.1129 & 2,373 & 0.0768 & 1,718.0541\\
Jerusalem Post & -0.1044 & 3,123 & 0.0853 & 2,468.0541\\
NL Times & -0.0898 & 179 & 0.0998 & -475.9459\\
thefirstnews & -0.0723 & 44 & 0.1174 & -610.9459\\
VRT & -0.0605 & 10 & 0.1292 & -644.9459\\
Ansa & -0.0571 & 33 & 0.1326 & -621.9459\\
Bangkokpost & -0.0336 & 191 & 0.156 & -463.9459\\
\botrule
\end{tabular}
\end{table}




\afterpage{\clearpage}
\pagebreak

\end{appendices}


\bibliography{media_bias-bibliography}


\begin{thebibliography}{101}
\ifx \bisbn   \undefined \def \bisbn  #1{ISBN #1}\fi
\ifx \binits  \undefined \def \binits#1{#1}\fi
\ifx \bauthor  \undefined \def \bauthor#1{#1}\fi
\ifx \batitle  \undefined \def \batitle#1{#1}\fi
\ifx \bjtitle  \undefined \def \bjtitle#1{#1}\fi
\ifx \bvolume  \undefined \def \bvolume#1{\textbf{#1}}\fi
\ifx \byear  \undefined \def \byear#1{#1}\fi
\ifx \bissue  \undefined \def \bissue#1{#1}\fi
\ifx \bfpage  \undefined \def \bfpage#1{#1}\fi
\ifx \blpage  \undefined \def \blpage #1{#1}\fi
\ifx \burl  \undefined \def \burl#1{\textsf{#1}}\fi
\ifx \doiurl  \undefined \def \doiurl#1{\url{https://doi.org/#1}}\fi
\ifx \betal  \undefined \def \betal{\textit{et al.}}\fi
\ifx \binstitute  \undefined \def \binstitute#1{#1}\fi
\ifx \binstitutionaled  \undefined \def \binstitutionaled#1{#1}\fi
\ifx \bctitle  \undefined \def \bctitle#1{#1}\fi
\ifx \beditor  \undefined \def \beditor#1{#1}\fi
\ifx \bpublisher  \undefined \def \bpublisher#1{#1}\fi
\ifx \bbtitle  \undefined \def \bbtitle#1{#1}\fi
\ifx \bedition  \undefined \def \bedition#1{#1}\fi
\ifx \bseriesno  \undefined \def \bseriesno#1{#1}\fi
\ifx \blocation  \undefined \def \blocation#1{#1}\fi
\ifx \bsertitle  \undefined \def \bsertitle#1{#1}\fi
\ifx \bsnm \undefined \def \bsnm#1{#1}\fi
\ifx \bsuffix \undefined \def \bsuffix#1{#1}\fi
\ifx \bparticle \undefined \def \bparticle#1{#1}\fi
\ifx \barticle \undefined \def \barticle#1{#1}\fi
\bibcommenthead
\ifx \bconfdate \undefined \def \bconfdate #1{#1}\fi
\ifx \botherref \undefined \def \botherref #1{#1}\fi
\ifx \url \undefined \def \url#1{\textsf{#1}}\fi
\ifx \bchapter \undefined \def \bchapter#1{#1}\fi
\ifx \bbook \undefined \def \bbook#1{#1}\fi
\ifx \bcomment \undefined \def \bcomment#1{#1}\fi
\ifx \oauthor \undefined \def \oauthor#1{#1}\fi
\ifx \citeauthoryear \undefined \def \citeauthoryear#1{#1}\fi
\ifx \endbibitem  \undefined \def \endbibitem {}\fi
\ifx \bconflocation  \undefined \def \bconflocation#1{#1}\fi
\ifx \arxivurl  \undefined \def \arxivurl#1{\textsf{#1}}\fi
\csname PreBibitemsHook\endcsname

\bibitem[\protect\citeauthoryear{Hamborg et~al.}{2019}]{Hamborg2019}
\begin{barticle}
\bauthor{\bsnm{Hamborg}, \binits{F.}},
\bauthor{\bsnm{Donnay}, \binits{K.}},
\bauthor{\bsnm{Gipp}, \binits{B.}}:
\batitle{{Automated identification of media bias in news articles: an interdisciplinary literature review}}.
\bjtitle{\textit{International Journal on Digital Libraries}}
\bvolume{20}(\bissue{4}),
\bfpage{391}--\blpage{415}
(\byear{2019})
\doiurl{10.1007/s00799-018-0261-y}
\end{barticle}
\endbibitem

\bibitem[\protect\citeauthoryear{Kroon and van~der Meer}{2023}]{Kroon2023}
\begin{barticle}
\bauthor{\bsnm{Kroon}, \binits{A.C.}},
\bauthor{\bsnm{Meer}, \binits{T.G.L.A.}}:
\batitle{{Who's to Fear? Implicit Sexual Threat Pre and Post the “Refugee Crisis”}}.
\bjtitle{\textit{Journalism Practice}}
\bvolume{17}(\bissue{2}),
\bfpage{319}--\blpage{335}
(\byear{2023})
\doiurl{10.1080/17512786.2021.1916401}
\end{barticle}
\endbibitem

\bibitem[\protect\citeauthoryear{Opperhuizen et~al.}{2019}]{Opperhuizen2019}
\begin{barticle}
\bauthor{\bsnm{Opperhuizen}, \binits{A.E.}},
\bauthor{\bsnm{Schouten}, \binits{K.}},
\bauthor{\bsnm{Klijn}, \binits{E.H.}}:
\batitle{{Framing a Conflict! How Media Report on Earthquake Risks Caused by Gas Drilling: A Longitudinal Analysis Using Machine Learning Techniques of Media Reporting on Gas Drilling from 1990 to 2015}}.
\bjtitle{\textit{Journalism Studies}}
\bvolume{20}(\bissue{5}),
\bfpage{714}--\blpage{734}
(\byear{2019})
\doiurl{10.1080/1461670X.2017.1418672}
\end{barticle}
\endbibitem

\bibitem[\protect\citeauthoryear{Galgoczy et~al.}{2022}]{Galgoczy2022}
\begin{barticle}
\bauthor{\bsnm{Galgoczy}, \binits{M.C.}},
\bauthor{\bsnm{Phatak}, \binits{A.}},
\bauthor{\bsnm{Vinson}, \binits{D.}},
\bauthor{\bsnm{Mago}, \binits{V.K.}},
\bauthor{\bsnm{Giabbanelli}, \binits{P.J.}}:
\batitle{{(Re)shaping online narratives: when bots promote the message of President Trump during his first impeachment}}.
\bjtitle{\textit{PeerJ Computer Science}}
\bvolume{8},
\bfpage{1}--\blpage{26}
(\byear{2022})
\doiurl{10.7717/peerj-cs.947}
\end{barticle}
\endbibitem

\bibitem[\protect\citeauthoryear{Bernhardt et~al.}{2008}]{Bernhardt2008}
\begin{barticle}
\bauthor{\bsnm{Bernhardt}, \binits{D.}},
\bauthor{\bsnm{Krasa}, \binits{S.}},
\bauthor{\bsnm{Polborn}, \binits{M.}}:
\batitle{{Political polarization and the electoral effects of media bias}}.
\bjtitle{\textit{Journal of Public Economics}}
\bvolume{92}(\bissue{5-6}),
\bfpage{1092}--\blpage{1104}
(\byear{2008})
\doiurl{10.1016/j.jpubeco.2008.01.006}
\end{barticle}
\endbibitem

\bibitem[\protect\citeauthoryear{Wilson et~al.}{2020}]{Wilson2020}
\begin{barticle}
\bauthor{\bsnm{Wilson}, \binits{A.E.}},
\bauthor{\bsnm{Parker}, \binits{V.}},
\bauthor{\bsnm{Feinberg}, \binits{M.}}:
\batitle{{Polarization in the contemporary political and media landscape}}.
\bjtitle{\textit{Current Opinion in Behavioral Sciences}}
\bvolume{34}(\bissue{October}),
\bfpage{223}--\blpage{228}
(\byear{2020})
\doiurl{10.1016/j.cobeha.2020.07.005}
\end{barticle}
\endbibitem

\bibitem[\protect\citeauthoryear{Baron}{2006}]{Baron2006}
\begin{barticle}
\bauthor{\bsnm{Baron}, \binits{D.P.}}:
\batitle{{Persistent media bias}}.
\bjtitle{\textit{Journal of Public Economics}}
\bvolume{90}(\bissue{1-2}),
\bfpage{1}--\blpage{36}
(\byear{2006})
\doiurl{10.1016/j.jpubeco.2004.10.006}
\end{barticle}
\endbibitem

\bibitem[\protect\citeauthoryear{D'Alonzo and Tegmark}{2022}]{DAlonzo2022}
\begin{barticle}
\bauthor{\bsnm{D'Alonzo}, \binits{S.}},
\bauthor{\bsnm{Tegmark}, \binits{M.}}:
\batitle{{Machine-learning media bias}}.
\bjtitle{\textit{PLoS ONE}}
\bvolume{17}(\bissue{8 August}),
\bfpage{1}--\blpage{24}
(\byear{2022})
\doiurl{10.1371/journal.pone.0271947}
{\href{https://arxiv.org/abs/2109.00024}{{arXiv:2109.00024}}}
\end{barticle}
\endbibitem

\bibitem[\protect\citeauthoryear{Park et~al.}{2009}]{Park2009}
\begin{botherref}
\oauthor{\bsnm{Park}, \binits{S.}},
\oauthor{\bsnm{Kang}, \binits{S.}},
\oauthor{\bsnm{Chung}, \binits{S.}},
\oauthor{\bsnm{Song}, \binits{J.}}:
{NewsCube: Delivering multiple aspects of news to mitigate media bias}.
\textit{Conference on Human Factors in Computing Systems - Proceedings},
443--452
(2009)
\doiurl{10.1145/1518701.1518772}
\end{botherref}
\endbibitem

\bibitem[\protect\citeauthoryear{Naredla and Adedoyin}{2022}]{Naredla2022}
\begin{barticle}
\bauthor{\bsnm{Naredla}, \binits{N.R.}},
\bauthor{\bsnm{Adedoyin}, \binits{F.F.}}:
\batitle{{Detection of hyperpartisan news articles using natural language processing technique}}.
\bjtitle{\textit{International Journal of Information Management Data Insights}}
\bvolume{2}(\bissue{1}),
\bfpage{100064}
(\byear{2022})
\doiurl{10.1016/j.jjimei.2022.100064}
\end{barticle}
\endbibitem

\bibitem[\protect\citeauthoryear{Shultziner and Stukalin}{2021}]{Shultziner2021}
\begin{barticle}
\bauthor{\bsnm{Shultziner}, \binits{D.}},
\bauthor{\bsnm{Stukalin}, \binits{Y.}}:
\batitle{{Distorting the News? The Mechanisms of Partisan Media Bias and Its Effects on News Production}}.
\bjtitle{\textit{Political Behavior}}
\bvolume{43}(\bissue{1}),
\bfpage{201}--\blpage{222}
(\byear{2021})
\doiurl{10.1007/s11109-019-09551-y}
\end{barticle}
\endbibitem

\bibitem[\protect\citeauthoryear{N{\'{e}}meth}{2022}]{Nemeth2022}
\begin{barticle}
\bauthor{\bsnm{N{\'{e}}meth}, \binits{R.}}:
\batitle{{A scoping review on the use of natural language processing in research on political polarization: trends and research prospects}}.
\bjtitle{\textit{Journal of Computational Social Science}}
\bvolume{6}(\bissue{1}),
\bfpage{289}--\blpage{313}
(\byear{2022})
\doiurl{10.1007/s42001-022-00196-2}
\end{barticle}
\endbibitem

\bibitem[\protect\citeauthoryear{Puglisi and Snyder}{2015}]{Puglisi2015}
\begin{bchapter}
\bauthor{\bsnm{Puglisi}, \binits{R.}},
\bauthor{\bsnm{Snyder}, \binits{J.M.}}:
\bctitle{{Empirical Studies of Media Bias}}.
In: \bbtitle{\textit{Handbook of Media Economics}},
pp. \bfpage{647}--\blpage{667}
(\byear{2015}).
\doiurl{10.1016/B978-0-444-63685-0.00015-2} .
\burl{https://linkinghub.elsevier.com/retrieve/pii/B9780444636850000152}
\end{bchapter}
\endbibitem

\bibitem[\protect\citeauthoryear{Brandtzaeg and F{\o}lstad}{2017}]{Brandtzaeg2017}
\begin{barticle}
\bauthor{\bsnm{Brandtzaeg}, \binits{P.B.}},
\bauthor{\bsnm{F{\o}lstad}, \binits{A.}}:
\batitle{{Trust and distrust in online fact-checking services}}.
\bjtitle{\textit{Communications of the ACM}}
\bvolume{60}(\bissue{9}),
\bfpage{65}--\blpage{71}
(\byear{2017})
\doiurl{10.1145/3122803}
\end{barticle}
\endbibitem

\bibitem[\protect\citeauthoryear{Kang and Yang}{2022}]{Kang2022}
\begin{barticle}
\bauthor{\bsnm{Kang}, \binits{H.}},
\bauthor{\bsnm{Yang}, \binits{J.}}:
\batitle{{Quantifying Perceived Political Bias of Newspapers through a Document Classification Technique}}.
\bjtitle{\textit{Journal of Quantitative Linguistics}}
\bvolume{29}(\bissue{2}),
\bfpage{127}--\blpage{150}
(\byear{2022})
\doiurl{10.1080/09296174.2020.1771136}
\end{barticle}
\endbibitem

\bibitem[\protect\citeauthoryear{D'Alessio and Allen}{2000}]{DAlessio2000}
\begin{barticle}
\bauthor{\bsnm{D'Alessio}, \binits{D.}},
\bauthor{\bsnm{Allen}, \binits{M.}}:
\batitle{{Media Bias in Presidential Elections: A Meta-Analysis}}.
\bjtitle{\textit{Journal of Communication}}
\bvolume{50}(\bissue{4}),
\bfpage{133}--\blpage{156}
(\byear{2000})
\doiurl{10.1111/j.1460-2466.2000.tb02866.x}
\end{barticle}
\endbibitem

\bibitem[\protect\citeauthoryear{Mullainathan and Shleifer}{2002}]{Mullainathan2002}
\begin{botherref}
\oauthor{\bsnm{Mullainathan}, \binits{S.}},
\oauthor{\bsnm{Shleifer}, \binits{A.}}:
{Media Bias}.
Technical report,
National Bureau of Economic Research,
Cambridge, MA
(oct 2002).
\doiurl{10.3386/w9295} .
\url{http://www.nber.org/papers/w9295.pdf}
\end{botherref}
\endbibitem

\bibitem[\protect\citeauthoryear{Rodrigo-Gin{\'{e}}s et~al.}{2024}]{Rodrigo-Gines2024}
\begin{barticle}
\bauthor{\bsnm{Rodrigo-Gin{\'{e}}s}, \binits{F.-J.}},
\bauthor{\bsnm{Carrillo-de-Albornoz}, \binits{J.}},
\bauthor{\bsnm{Plaza}, \binits{L.}}:
\batitle{{A systematic review on media bias detection: What is media bias, how it is expressed, and how to detect it}}.
\bjtitle{\textit{Expert Systems with Applications}}
\bvolume{237},
\bfpage{121641}
(\byear{2024})
\doiurl{10.1016/j.eswa.2023.121641}
\end{barticle}
\endbibitem

\bibitem[\protect\citeauthoryear{Vaswani et~al.}{2017}]{Vaswani2017}
\begin{botherref}
\oauthor{\bsnm{Vaswani}, \binits{A.}},
\oauthor{\bsnm{Shazeer}, \binits{N.}},
\oauthor{\bsnm{Parmar}, \binits{N.}},
\oauthor{\bsnm{Uszkoreit}, \binits{J.}},
\oauthor{\bsnm{Jones}, \binits{L.}},
\oauthor{\bsnm{Gomez}, \binits{A.N.}},
\oauthor{\bsnm{Kaiser}, \binits{L.}},
\oauthor{\bsnm{Polosukhin}, \binits{I.}}:
{Attention Is All You Need}.
\textit{arXiv: 1706.03762}
(2017)
{\href{https://arxiv.org/abs/1706.03762}{{arXiv:1706.03762}}}
\end{botherref}
\endbibitem

\bibitem[\protect\citeauthoryear{Baly et~al.}{2020}]{Baly2020a}
\begin{bchapter}
\bauthor{\bsnm{Baly}, \binits{R.}},
\bauthor{\bsnm{{Da San Martino}}, \binits{G.}},
\bauthor{\bsnm{Glass}, \binits{J.}},
\bauthor{\bsnm{Nakov}, \binits{P.}}:
\bctitle{{We Can Detect Your Bias: Predicting the Political Ideology of News Articles}}.
In: \bbtitle{\textit{Proceedings of the 2020 Conference on Empirical Methods in Natural Language Processing (EMNLP)}},
pp. \bfpage{4982}--\blpage{4991}.
\bpublisher{Association for Computational Linguistics},
\blocation{Stroudsburg, PA, USA}
(\byear{2020}).
\doiurl{10.18653/v1/2020.emnlp-main.404} .
\burl{https://www.aclweb.org/anthology/2020.emnlp-main.404}
\end{bchapter}
\endbibitem

\bibitem[\protect\citeauthoryear{Quijote et~al.}{2019}]{Quijote2019}
\begin{barticle}
\bauthor{\bsnm{Quijote}, \binits{T.A.}},
\bauthor{\bsnm{Zamoras}, \binits{A.D.}},
\bauthor{\bsnm{Ceniza}, \binits{A.}}:
\batitle{{Bias detection in Philippine political news articles using SentiWordNet and inverse reinforcement model}}.
\bjtitle{\textit{IOP Conference Series: Materials Science and Engineering}}
\bvolume{482},
\bfpage{012036}
(\byear{2019})
\doiurl{10.1088/1757-899X/482/1/012036}
\end{barticle}
\endbibitem

\bibitem[\protect\citeauthoryear{Agrawal et~al.}{2022}]{Agrawal2022}
\begin{bchapter}
\bauthor{\bsnm{Agrawal}, \binits{S.}},
\bauthor{\bsnm{Gupta}, \binits{K.}},
\bauthor{\bsnm{Gautam}, \binits{D.}},
\bauthor{\bsnm{Mamidi}, \binits{R.}}:
\bctitle{{Towards Detecting Political Bias in Hindi News Articles}}.
In: \bbtitle{\textit{Proceedings of the 60th Annual Meeting of the Association for Computational Linguistics: Student Research Workshop}},
pp. \bfpage{239}--\blpage{244}.
\bpublisher{Association for Computational Linguistics},
\blocation{Stroudsburg, PA, USA}
(\byear{2022}).
\doiurl{10.18653/v1/2022.acl-srw.17} .
\burl{https://aclanthology.org/2022.acl-srw.17}
\end{bchapter}
\endbibitem

\bibitem[\protect\citeauthoryear{Sinha and Dasgupta}{2021}]{Sinha2021}
\begin{bchapter}
\bauthor{\bsnm{Sinha}, \binits{M.}},
\bauthor{\bsnm{Dasgupta}, \binits{T.}}:
\bctitle{{Determining Subjective Bias in Text through Linguistically Informed Transformer based Multi-Task Network}}.
In: \bbtitle{\textit{Proceedings of the 30th ACM International Conference on Information {\&} Knowledge Management}},
pp. \bfpage{3418}--\blpage{3422}.
\bpublisher{ACM},
\blocation{New York, NY, USA}
(\byear{2021}).
\doiurl{10.1145/3459637.3482084} .
\burl{https://dl.acm.org/doi/10.1145/3459637.3482084}
\end{bchapter}
\endbibitem

\bibitem[\protect\citeauthoryear{Chen et~al.}{2018}]{Chen2018}
\begin{bchapter}
\bauthor{\bsnm{Chen}, \binits{W.-F.}},
\bauthor{\bsnm{Wachsmuth}, \binits{H.}},
\bauthor{\bsnm{Al-Khatib}, \binits{K.}},
\bauthor{\bsnm{Stein}, \binits{B.}}:
\bctitle{{Learning to Flip the Bias of News Headlines}}.
In: \bbtitle{\textit{Proceedings of the 11th International Conference on Natural Language Generation}},
pp. \bfpage{79}--\blpage{88}.
\bpublisher{Association for Computational Linguistics},
\blocation{Stroudsburg, PA, USA}
(\byear{2018}).
\doiurl{10.18653/v1/W18-6509} .
\burl{http://aclweb.org/anthology/W18-6509}
\end{bchapter}
\endbibitem

\bibitem[\protect\citeauthoryear{Fan et~al.}{2019}]{Fan2019}
\begin{bchapter}
\bauthor{\bsnm{Fan}, \binits{L.}},
\bauthor{\bsnm{White}, \binits{M.}},
\bauthor{\bsnm{Sharma}, \binits{E.}},
\bauthor{\bsnm{Su}, \binits{R.}},
\bauthor{\bsnm{Choubey}, \binits{P.K.}},
\bauthor{\bsnm{Huang}, \binits{R.}},
\bauthor{\bsnm{Wang}, \binits{L.}}:
\bctitle{{In Plain Sight: Media Bias Through the Lens of Factual Reporting}}.
In: \bbtitle{\textit{Proceedings of the 2019 Conference on Empirical Methods in Natural Language Processing and the 9th International Joint Conference on Natural Language Processing (EMNLP-IJCNLP)}},
pp. \bfpage{6342}--\blpage{6348}.
\bpublisher{Association for Computational Linguistics},
\blocation{Stroudsburg, PA, USA}
(\byear{2019}).
\doiurl{10.18653/v1/D19-1664} .
\burl{https://www.aclweb.org/anthology/D19-1664}
\end{bchapter}
\endbibitem

\bibitem[\protect\citeauthoryear{Spinde et~al.}{2022}]{Spinde2022}
\begin{barticle}
\bauthor{\bsnm{Spinde}, \binits{T.}},
\bauthor{\bsnm{Krieger}, \binits{J.-D.}},
\bauthor{\bsnm{Ruas}, \binits{T.}},
\bauthor{\bsnm{Mitrovi{\'{c}}}, \binits{J.}},
\bauthor{\bsnm{G{\"{o}}tz-Hahn}, \binits{F.}},
\bauthor{\bsnm{Aizawa}, \binits{A.}},
\bauthor{\bsnm{Gipp}, \binits{B.}}:
\batitle{{Exploiting Transformer-based Multitask Learning for the Detection of Media Bias in News Articles}}.
\bjtitle{\textit{arXiv: 2211.03491}}
(\byear{2022})
\doiurl{10.1007/978-3-030-96957-8_20}
{\href{https://arxiv.org/abs/2211.03491}{{arXiv:2211.03491}}}
\end{barticle}
\endbibitem

\bibitem[\protect\citeauthoryear{Bourgeois et~al.}{2018}]{Bourgeois2018}
\begin{botherref}
\oauthor{\bsnm{Bourgeois}, \binits{D.}},
\oauthor{\bsnm{Rappaz}, \binits{J.}},
\oauthor{\bsnm{Aberer}, \binits{K.}}:
{Selection Bias in News Coverage: Learning it, Fighting it}.
\textit{The Web Conference 2018 - Companion of the World Wide Web Conference, WWW 2018},
535--543
(2018)
\doiurl{10.1145/3184558.3188724}
{\href{https://arxiv.org/abs/1904.07536}{{arXiv:1904.07536}}}
\end{botherref}
\endbibitem

\bibitem[\protect\citeauthoryear{Saez-Trumper et~al.}{2013}]{Saez-Trumper2013}
\begin{botherref}
\oauthor{\bsnm{Saez-Trumper}, \binits{D.}},
\oauthor{\bsnm{Castillo}, \binits{C.}},
\oauthor{\bsnm{Lalmas}, \binits{M.}}:
{Social media news communities: Gatekeeping, coverage, and statement bias}.
\textit{International Conference on Information and Knowledge Management, Proceedings}
(June 2014),
1679--1684
(2013)
\doiurl{10.1145/2505515.2505623}
\end{botherref}
\endbibitem

\bibitem[\protect\citeauthoryear{Ehrhardt et~al.}{2021}]{Ehrhardt2021}
\begin{botherref}
\oauthor{\bsnm{Ehrhardt}, \binits{J.}},
\oauthor{\bsnm{Spinde}, \binits{T.}},
\oauthor{\bsnm{Vardasbi}, \binits{A.}},
\oauthor{\bsnm{Hamborg}, \binits{F.}}:
{Omission of information : Identifying political slant via an analysis of co-occurring entities}.
\textit{Information between Data and Knowledge},
80--93
(2021)
\doiurl{10.5283/epub.44931}
\end{botherref}
\endbibitem

\bibitem[\protect\citeauthoryear{Gandhi et~al.}{2023}]{Gandhi2023}
\begin{barticle}
\bauthor{\bsnm{Gandhi}, \binits{A.}},
\bauthor{\bsnm{Adhvaryu}, \binits{K.}},
\bauthor{\bsnm{Poria}, \binits{S.}},
\bauthor{\bsnm{Cambria}, \binits{E.}},
\bauthor{\bsnm{Hussain}, \binits{A.}}:
\batitle{{Multimodal sentiment analysis: A systematic review of history, datasets, multimodal fusion methods, applications, challenges and future directions}}.
\bjtitle{\textit{Information Fusion}}
\bvolume{91},
\bfpage{424}--\blpage{444}
(\byear{2023})
\doiurl{10.1016/J.INFFUS.2022.09.025}
\end{barticle}
\endbibitem

\bibitem[\protect\citeauthoryear{Molenaar et~al.}{2023}]{Molenaar2023}
\begin{botherref}
\oauthor{\bsnm{Molenaar}, \binits{I.}},
\oauthor{\bsnm{Mooij}, \binits{S.}},
\oauthor{\bsnm{Azevedo}, \binits{R.}},
\oauthor{\bsnm{Bannert}, \binits{M.}},
\oauthor{\bsnm{J{\"{a}}rvel{\"{a}}}, \binits{S.}},
\oauthor{\bsnm{Ga{\v{s}}evi{\'{c}}}, \binits{D.}}:
{Measuring self-regulated learning and the role of AI: Five years of research using multimodal multichannel data}.
\textit{Computers in Human Behavior}
\textbf{139}(September 2022)
(2023)
\doiurl{10.1016/j.chb.2022.107540}
\end{botherref}
\endbibitem

\bibitem[\protect\citeauthoryear{Xu et~al.}{2023}]{Xu2023}
\begin{barticle}
\bauthor{\bsnm{Xu}, \binits{P.}},
\bauthor{\bsnm{Zhu}, \binits{X.}},
\bauthor{\bsnm{Clifton}, \binits{D.A.}}:
\batitle{{Multimodal Learning With Transformers: A Survey}}.
\bjtitle{\textit{IEEE Transactions on Pattern Analysis and Machine Intelligence}}
\bvolume{45}(\bissue{10}),
\bfpage{12113}--\blpage{12132}
(\byear{2023})
\doiurl{10.1109/TPAMI.2023.3275156}
{\href{https://arxiv.org/abs/2206.06488}{{arXiv:2206.06488}}}
\end{barticle}
\endbibitem

\bibitem[\protect\citeauthoryear{Alexandropoulos et~al.}{2019}]{Alexandropoulos2019}
\begin{botherref}
\oauthor{\bsnm{Alexandropoulos}, \binits{S.A.N.}},
\oauthor{\bsnm{Kotsiantis}, \binits{S.B.}},
\oauthor{\bsnm{Vrahatis}, \binits{M.N.}}:
{Data preprocessing in predictive data mining}.
\textit{Knowledge Engineering Review}
\textbf{34}(April 2020)
(2019)
\doiurl{10.1017/S026988891800036X}
\end{botherref}
\endbibitem

\bibitem[\protect\citeauthoryear{Siino et~al.}{2024}]{Siino2024a}
\begin{barticle}
\bauthor{\bsnm{Siino}, \binits{M.}},
\bauthor{\bsnm{Tinnirello}, \binits{I.}},
\bauthor{\bsnm{{La Cascia}}, \binits{M.}}:
\batitle{{Is text preprocessing still worth the time? A comparative survey on the influence of popular preprocessing methods on Transformers and traditional classifiers}}.
\bjtitle{\textit{Information Systems}}
\bvolume{121}(\bissue{March 2023}),
\bfpage{102342}
(\byear{2024})
\doiurl{10.1016/j.is.2023.102342}
\end{barticle}
\endbibitem

\bibitem[\protect\citeauthoryear{Symeonidis et~al.}{2018}]{Symeonidis2018}
\begin{barticle}
\bauthor{\bsnm{Symeonidis}, \binits{S.}},
\bauthor{\bsnm{Effrosynidis}, \binits{D.}},
\bauthor{\bsnm{Arampatzis}, \binits{A.}}:
\batitle{{A comparative evaluation of pre-processing techniques and their interactions for twitter sentiment analysis}}.
\bjtitle{\textit{Expert Systems with Applications}}
\bvolume{110},
\bfpage{298}--\blpage{310}
(\byear{2018})
\doiurl{10.1016/j.eswa.2018.06.022}
\end{barticle}
\endbibitem

\bibitem[\protect\citeauthoryear{Hickman et~al.}{2022}]{Hickman2022}
\begin{barticle}
\bauthor{\bsnm{Hickman}, \binits{L.}},
\bauthor{\bsnm{Thapa}, \binits{S.}},
\bauthor{\bsnm{Tay}, \binits{L.}},
\bauthor{\bsnm{Cao}, \binits{M.}},
\bauthor{\bsnm{Srinivasan}, \binits{P.}}:
\batitle{{Text Preprocessing for Text Mining in Organizational Research: Review and Recommendations}}.
\bjtitle{\textit{Organizational Research Methods}}
\bvolume{25}(\bissue{1}),
\bfpage{114}--\blpage{146}
(\byear{2022})
\doiurl{10.1177/1094428120971683}
\end{barticle}
\endbibitem

\bibitem[\protect\citeauthoryear{Goyal et~al.}{2021}]{Goyal2021}
\begin{bchapter}
\bauthor{\bsnm{Goyal}, \binits{N.}},
\bauthor{\bsnm{Du}, \binits{J.}},
\bauthor{\bsnm{Ott}, \binits{M.}},
\bauthor{\bsnm{Anantharaman}, \binits{G.}},
\bauthor{\bsnm{Conneau}, \binits{A.}}:
\bctitle{{Larger-Scale Transformers for Multilingual Masked Language Modeling}}.
In: \bbtitle{\textit{RepL4NLP 2021 - 6th Workshop on Representation Learning for NLP, Proceedings of the Workshop}}
(\byear{2021}).
\doiurl{10.18653/v1/2021.repl4nlp-1.4}
\end{bchapter}
\endbibitem

\bibitem[\protect\citeauthoryear{Beauchemin}{2021}]{Beauchemin2021}
\begin{botherref}
\oauthor{\bsnm{Beauchemin}, \binits{D.}}:
spacy-language-detection 0.2.1
(2021).
\url{https://pypi.org/project/spacy-language-detection/}
Accessed 2024-02-17
\end{botherref}
\endbibitem

\bibitem[\protect\citeauthoryear{Buchkremer et~al.}{2019}]{Buchkremer2019}
\begin{barticle}
\bauthor{\bsnm{Buchkremer}, \binits{R.}},
\bauthor{\bsnm{Demund}, \binits{A.}},
\bauthor{\bsnm{Ebener}, \binits{S.}},
\bauthor{\bsnm{Gampfer}, \binits{F.}},
\bauthor{\bsnm{Jägering}, \binits{D.}},
\bauthor{\bsnm{Jürgens}, \binits{A.}},
\bauthor{\bsnm{Klenke}, \binits{S.}},
\bauthor{\bsnm{Krimpmann}, \binits{D.}},
\bauthor{\bsnm{Schmank}, \binits{J.}},
\bauthor{\bsnm{Spiekermann}, \binits{M.}},
\bauthor{\bsnm{Wahlers}, \binits{M.}},
\bauthor{\bsnm{Wiepke}, \binits{M.}}:
\batitle{The application of artificial intelligence technologies as a substitute for reading and to support and enhance the authoring of scientific review articles}.
\bjtitle{IEEE Access}
\bvolume{7},
\bfpage{65263}--\blpage{65276}
(\byear{2019})
\doiurl{10.1109/ACCESS.2019.2917719}
\end{barticle}
\endbibitem

\bibitem[\protect\citeauthoryear{Grootendorst}{2020}]{Grootendorst2020}
\begin{botherref}
\oauthor{\bsnm{Grootendorst}, \binits{M.}}:
{BERTopic: Neural topic modeling with a class-based TF-IDF procedure}.
\textit{arXiv preprint arXiv:2203.05794}
(2020)
{\href{https://arxiv.org/abs/2203.05794v1}{{arXiv:2203.05794v1}}}
\end{botherref}
\endbibitem

\bibitem[\protect\citeauthoryear{Grootendorst}{2024}]{Grootendorst2024a}
\begin{botherref}
\oauthor{\bsnm{Grootendorst}, \binits{M.}}:
{The Algorithm}
(2024).
\url{https://maartengr.github.io/BERTopic/algorithm/algorithm.html}
Accessed 2024-01-19
\end{botherref}
\endbibitem

\bibitem[\protect\citeauthoryear{Reimers and Gurevych}{2019}]{Reimers2019}
\begin{botherref}
\oauthor{\bsnm{Reimers}, \binits{N.}},
\oauthor{\bsnm{Gurevych}, \binits{I.}}:
{Sentence-BERT: Sentence embeddings using siamese BERT-networks}.
\textit{EMNLP-IJCNLP 2019 - 2019 Conference on Empirical Methods in Natural Language Processing and 9th International Joint Conference on Natural Language Processing, Proceedings of the Conference},
3982--3992
(2019)
\doiurl{10.18653/v1/d19-1410}
{\href{https://arxiv.org/abs/1908.10084}{{arXiv:1908.10084}}}
\end{botherref}
\endbibitem

\bibitem[\protect\citeauthoryear{Mcinnes et~al.}{2020}]{Mcinnes2020}
\begin{botherref}
\oauthor{\bsnm{Mcinnes}, \binits{L.}},
\oauthor{\bsnm{Healy}, \binits{J.}},
\oauthor{\bsnm{Melville}, \binits{J.}}:
{UMAP : Uniform Manifold Approximation and Projection for Dimension Reduction}.
\textit{arXiv:1802.03426}
(2020)
{\href{https://arxiv.org/abs/1802.03426v3}{{arXiv:1802.03426v3}}}
\end{botherref}
\endbibitem

\bibitem[\protect\citeauthoryear{Campello et~al.}{2013}]{Campello2013}
\begin{botherref}
\oauthor{\bsnm{Campello}, \binits{R.J.G.B.}},
\oauthor{\bsnm{Moulavi}, \binits{D.}},
\oauthor{\bsnm{Sander}, \binits{J.}}:
{Density-Based Clustering Based on Hierarchical Density Estimates}.
\textit{Advances in Knowledge Discovery and Data Mining},
160--172
(2013)
\doiurl{10.1007/978-3-642-37456-2_14}
\end{botherref}
\endbibitem

\bibitem[\protect\citeauthoryear{}{2023}]{Scikitlearn}
\begin{botherref}
{sklearn.feature{\_}extraction.text.CountVectorizer}
(2023).
\url{https://scikit-learn.org/stable/modules/generated/sklearn.feature{\_}extraction.text.CountVectorizer.html}
Accessed 2024-01-18
\end{botherref}
\endbibitem

\bibitem[\protect\citeauthoryear{Schubert et~al.}{2017}]{Schubert2017}
\begin{botherref}
\oauthor{\bsnm{Schubert}, \binits{E.}},
\oauthor{\bsnm{Sander}, \binits{J.}},
\oauthor{\bsnm{Ester}, \binits{M.}},
\oauthor{\bsnm{Kriegel}, \binits{H.P.}},
\oauthor{\bsnm{Xu}, \binits{X.}}:
{DBSCAN revisited, revisited: Why and how you should (still) use DBSCAN}.
\textit{ACM Transactions on Database Systems}
\textbf{42}(3)
(2017)
\doiurl{10.1145/3068335}
\end{botherref}
\endbibitem

\bibitem[\protect\citeauthoryear{Mckinney}{2011}]{Mckinney2011}
\begin{botherref}
\oauthor{\bsnm{Mckinney}, \binits{W.}}:
{pandas: a Foundational Python Library for Data Analysis and Statistics}
(2011)
\end{botherref}
\endbibitem

\bibitem[\protect\citeauthoryear{Grootendorst}{2024}]{Grootendorst2024}
\begin{botherref}
\oauthor{\bsnm{Grootendorst}, \binits{M.}}:
{BERTopic}
(2024).
\url{https://maartengr.github.io/BERTopic/api/bertopic.html}
Accessed 2024-01-18
\end{botherref}
\endbibitem

\bibitem[\protect\citeauthoryear{Liu et~al.}{2019}]{Liu2019}
\begin{botherref}
\oauthor{\bsnm{Liu}, \binits{Y.}},
\oauthor{\bsnm{Ott}, \binits{M.}},
\oauthor{\bsnm{Goyal}, \binits{N.}},
\oauthor{\bsnm{Du}, \binits{J.}},
\oauthor{\bsnm{Joshi}, \binits{M.}},
\oauthor{\bsnm{Chen}, \binits{D.}},
\oauthor{\bsnm{Levy}, \binits{O.}},
\oauthor{\bsnm{Lewis}, \binits{M.}},
\oauthor{\bsnm{Zettlemoyer}, \binits{L.}},
\oauthor{\bsnm{Stoyanov}, \binits{V.}}:
{RoBERTa: A Robustly Optimized BERT Pretraining Approach}.
\textit{arXiv: 1907.11692}
(1)
(2019)
{\href{https://arxiv.org/abs/1907.11692}{{arXiv:1907.11692}}}
\end{botherref}
\endbibitem

\bibitem[\protect\citeauthoryear{Edwardes}{2024}]{Edwardes2024}
\begin{botherref}
\oauthor{\bsnm{Edwardes}, \binits{S.}}:
spacytextblob
(2024).
\url{https://spacy.io/universe/project/spacy-textblob}
Accessed 2024-02-17
\end{botherref}
\endbibitem

\bibitem[\protect\citeauthoryear{Explosion}{2023}]{Explosion2023}
\begin{botherref}
\oauthor{\bsnm{Explosion}}:
en{\_}core{\_}web{\_}sm
(2023).
\url{https://spacy.io/models/en{\#}en{\_}core{\_}web{\_}sm}
Accessed 2024-02-17
\end{botherref}
\endbibitem

\bibitem[\protect\citeauthoryear{Weischedel et~al.}{2013}]{Weischedel2013}
\begin{botherref}
\oauthor{\bsnm{Weischedel}, \binits{R.}},
\oauthor{\bsnm{Palmer}, \binits{M.}},
\oauthor{\bsnm{Marcus}, \binits{M.}},
\oauthor{\bsnm{Hovy}, \binits{E.}},
\oauthor{\bsnm{Pradhan}, \binits{S.}},
\oauthor{\bsnm{Ramshaw}, \binits{L.}},
\oauthor{\bsnm{Xue}, \binits{N.}},
\oauthor{\bsnm{Taylor}, \binits{A.}},
\oauthor{\bsnm{Kaufman}, \binits{J.}},
\oauthor{\bsnm{Franchini}, \binits{M.}},
\oauthor{\bsnm{Others}}:
{Ontonotes release 5.0}.
Linguistic Data Consortium
(2013).
\doiurl{10.35111/xmhb-2b84} .
\url{https://catalog.ldc.upenn.edu/LDC2013T19}
\end{botherref}
\endbibitem

\bibitem[\protect\citeauthoryear{SpaCy}{2023}]{SpaCy2023}
\begin{botherref}
\oauthor{\bsnm{SpaCy}}:
{English}
(2023).
\url{https://github.com/explosion/spacy-models/releases/tag/en{\_}core{\_}web{\_}sm-3.7.1}
\end{botherref}
\endbibitem

\bibitem[\protect\citeauthoryear{Jadhav}{2020}]{Jadhav2020}
\begin{botherref}
\oauthor{\bsnm{Jadhav}, \binits{S.A.}}:
{Detecting Potential Topics In News Using BERT, CRF and Wikipedia}.
\textit{arXiv: 2002.11402}
(2020)
{\href{https://arxiv.org/abs/2002.11402}{{arXiv:2002.11402}}}
\end{botherref}
\endbibitem

\bibitem[\protect\citeauthoryear{Vychegzhanin and Kotelnikov}{2019}]{Vychegzhanin2019}
\begin{botherref}
\oauthor{\bsnm{Vychegzhanin}, \binits{S.}},
\oauthor{\bsnm{Kotelnikov}, \binits{E.}}:
{Comparison of named entity recognition tools applied to news articles}.
\textit{Proceedings - 2019 Ivannikov Ispras Open Conference, ISPRAS 2019}
(February),
72--77
(2019)
\doiurl{10.1109/ISPRAS47671.2019.00017}
\end{botherref}
\endbibitem

\bibitem[\protect\citeauthoryear{Kamel et~al.}{2019}]{Kamel2019}
\begin{barticle}
\bauthor{\bsnm{Kamel}, \binits{M.}},
\bauthor{\bsnm{Siuky}, \binits{F.N.}},
\bauthor{\bsnm{Yazdi}, \binits{H.S.}}:
\batitle{{Robust sentiment fusion on distribution of news}}.
\bjtitle{\textit{Multimedia Tools and Applications}}
\bvolume{78}(\bissue{15}),
\bfpage{21917}--\blpage{21942}
(\byear{2019})
\doiurl{10.1007/s11042-019-7505-8}
\end{barticle}
\endbibitem

\bibitem[\protect\citeauthoryear{Mandalapu et~al.}{2019}]{Mandalapu2019}
\begin{barticle}
\bauthor{\bsnm{Mandalapu}, \binits{A.C.}},
\bauthor{\bsnm{Gunabalan}, \binits{S.}},
\bauthor{\bsnm{Sadineni}, \binits{A.}},
\bauthor{\bsnm{Cai}, \binits{T.}},
\bauthor{\bsnm{{Al Hasan Haldar}}, \binits{N.}},
\bauthor{\bsnm{Li}, \binits{J.}}:
\batitle{{Correlate Influential News Article Events to Stock Quote Movement}}.
\bjtitle{\textit{Advanced Data Mining and Applications. ADMA 2019. Lecture Notes in Computer Science}}
\bvolume{11888 LNAI}(\bissue{November}),
\bfpage{331}--\blpage{342}
(\byear{2019})
\doiurl{10.1007/978-3-030-35231-8_24}
\end{barticle}
\endbibitem

\bibitem[\protect\citeauthoryear{Hamborg and Donnay}{2021}]{Hamborg2021}
\begin{bchapter}
\bauthor{\bsnm{Hamborg}, \binits{F.}},
\bauthor{\bsnm{Donnay}, \binits{K.}}:
\bctitle{{NewsMTSC: A dataset for (multi-)target-dependent sentiment classification in political news articles}}.
In: \bbtitle{\textit{EACL 2021 - 16th Conference of the European Chapter of the Association for Computational Linguistics, Proceedings of the Conference}},
pp. \bfpage{1663}--\blpage{1675}
(\byear{2021}).
\doiurl{10.18653/v1/2021.eacl-main.142}
\end{bchapter}
\endbibitem

\bibitem[\protect\citeauthoryear{Harth et~al.}{2023}]{Harth2023}
\begin{barticle}
\bauthor{\bsnm{Harth}, \binits{P.}},
\bauthor{\bsnm{J{\"{a}}hde}, \binits{O.}},
\bauthor{\bsnm{Schneider}, \binits{S.}},
\bauthor{\bsnm{Horn}, \binits{N.}},
\bauthor{\bsnm{Buchkremer}, \binits{R.}}:
\batitle{{From Data to Human-Readable Requirements: Advancing Requirements Elicitation through Language-Transformer-Enhanced Opportunity Mining}}.
\bjtitle{\textit{Algorithms}}
\bvolume{16}(\bissue{9}),
\bfpage{403}
(\byear{2023})
\doiurl{10.3390/a16090403}
\end{barticle}
\endbibitem

\bibitem[\protect\citeauthoryear{Devlin et~al.}{2019}]{Devlin2019}
\begin{barticle}
\bauthor{\bsnm{Devlin}, \binits{J.}},
\bauthor{\bsnm{Chang}, \binits{M.W.}},
\bauthor{\bsnm{Lee}, \binits{K.}},
\bauthor{\bsnm{Toutanova}, \binits{K.}}:
\batitle{{BERT: Pre-training of deep bidirectional transformers for language understanding}}.
\bjtitle{\textit{NAACL HLT 2019 - 2019 Conference of the North American Chapter of the Association for Computational Linguistics: Human Language Technologies - Proceedings of the Conference}}
\bvolume{1}(\bissue{Mlm}),
\bfpage{4171}--\blpage{4186}
(\byear{2019})
{\href{https://arxiv.org/abs/1810.04805}{{arXiv:1810.04805}}}
\end{barticle}
\endbibitem

\bibitem[\protect\citeauthoryear{Huggingface}{2024}]{Huggingface2024}
\begin{botherref}
\oauthor{\bsnm{Huggingface}}:
transformers 4.37.2
(2024).
\url{https://pypi.org/project/transformers/}
Accessed 2024-02-17
\end{botherref}
\endbibitem

\bibitem[\protect\citeauthoryear{Sang and {De Meulder}}{2003}]{sang2003}
\begin{bchapter}
\bauthor{\bsnm{Sang}, \binits{E.F.T.K.}},
\bauthor{\bsnm{{De Meulder}}, \binits{F.}}:
\bctitle{{Introduction to the CoNLL-2003 Shared Task: Language-Independent Named Entity Recognition}}.
In: \bbtitle{\textit{Proceedings of the Seventh Conference on Natural Language Learning at HLT-NAACL 2003}},
pp. \bfpage{142}--\blpage{147}
(\byear{2003}).
\burl{https://aclanthology.org/W03-0419}
\end{bchapter}
\endbibitem

\bibitem[\protect\citeauthoryear{Tedeschi and Navigli}{2022}]{Tedeschi2022}
\begin{bchapter}
\bauthor{\bsnm{Tedeschi}, \binits{S.}},
\bauthor{\bsnm{Navigli}, \binits{R.}}:
\bctitle{{MultiNERD: A Multilingual, Multi-Genre and Fine-Grained Dataset for Named Entity Recognition (and Disambiguation)}}.
In: \bbtitle{\textit{Findings of the Association for Computational Linguistics: NAACL 2022 - Findings}},
pp. \bfpage{801}--\blpage{812}
(\byear{2022}).
\doiurl{10.18653/v1/2022.findings-naacl.60}
\end{bchapter}
\endbibitem

\bibitem[\protect\citeauthoryear{Ringland et~al.}{2020}]{Ringland2020}
\begin{bchapter}
\bauthor{\bsnm{Ringland}, \binits{N.}},
\bauthor{\bsnm{Dai}, \binits{X.}},
\bauthor{\bsnm{Hachey}, \binits{B.}},
\bauthor{\bsnm{Karimi}, \binits{S.}},
\bauthor{\bsnm{Paris}, \binits{C.}},
\bauthor{\bsnm{Curran}, \binits{J.R.}}:
\bctitle{{NNE: A dataset for nested named entity recognition in English newswire}}.
In: \bbtitle{\textit{ACL 2019 - 57th Annual Meeting of the Association for Computational Linguistics, Proceedings of the Conference}},
pp. \bfpage{5176}--\blpage{5181}
(\byear{2020}).
\doiurl{10.18653/v1/p19-1510}
\end{bchapter}
\endbibitem

\bibitem[\protect\citeauthoryear{Liu and Ritter}{2023}]{Liu2023}
\begin{botherref}
\oauthor{\bsnm{Liu}, \binits{S.}},
\oauthor{\bsnm{Ritter}, \binits{A.}}:
{Do CoNLL-2003 Named Entity Taggers Still Work Well in 2023?}
\textit{arXiv: 2212.09747},
8254--8271
(2023)
\doiurl{10.18653/v1/2023.acl-long.459}
{\href{https://arxiv.org/abs/2212.09747}{{arXiv:2212.09747}}}
\end{botherref}
\endbibitem

\bibitem[\protect\citeauthoryear{Hu et~al.}{2024}]{Hu2024}
\begin{barticle}
\bauthor{\bsnm{Hu}, \binits{X.}},
\bauthor{\bsnm{Zhou}, \binits{Z.}},
\bauthor{\bsnm{Li}, \binits{H.}},
\bauthor{\bsnm{Hu}, \binits{Y.}},
\bauthor{\bsnm{Gu}, \binits{F.}},
\bauthor{\bsnm{Kersten}, \binits{J.}},
\bauthor{\bsnm{Fan}, \binits{H.}},
\bauthor{\bsnm{Klan}, \binits{F.}}:
\batitle{{Location Reference Recognition from Texts: A Survey and Comparison}}.
\bjtitle{\textit{ACM Computing Surveys}}
\bvolume{56}(\bissue{5}),
\bfpage{1}--\blpage{37}
(\byear{2024})
\doiurl{10.1145/3625819}
{\href{https://arxiv.org/abs/2207.01683}{{arXiv:2207.01683}}}
\end{barticle}
\endbibitem

\bibitem[\protect\citeauthoryear{Polignano et~al.}{2021}]{Polignano2021}
\begin{barticle}
\bauthor{\bsnm{Polignano}, \binits{M.}},
\bauthor{\bsnm{Gemmis}, \binits{M.}},
\bauthor{\bsnm{Semeraro}, \binits{G.}}:
\batitle{{Comparing transformer-based NER approaches for analysing textual medical diagnoses}}.
\bjtitle{\textit{CEUR Workshop Proceedings}}
\bvolume{2936},
\bfpage{818}--\blpage{833}
(\byear{2021})
\end{barticle}
\endbibitem

\bibitem[\protect\citeauthoryear{Zhang et~al.}{2023}]{Zhang2023}
\begin{barticle}
\bauthor{\bsnm{Zhang}, \binits{W.}},
\bauthor{\bsnm{Li}, \binits{X.}},
\bauthor{\bsnm{Deng}, \binits{Y.}},
\bauthor{\bsnm{Bing}, \binits{L.}},
\bauthor{\bsnm{Lam}, \binits{W.}}:
\batitle{{A Survey on Aspect-Based Sentiment Analysis: Tasks, Methods, and Challenges}}.
\bjtitle{\textit{IEEE Transactions on Knowledge and Data Engineering}}
\bvolume{35}(\bissue{11}),
\bfpage{11019}--\blpage{11038}
(\byear{2023})
\doiurl{10.1109/TKDE.2022.3230975}
\end{barticle}
\endbibitem

\bibitem[\protect\citeauthoryear{Wu et~al.}{2022}]{Wu2022}
\begin{barticle}
\bauthor{\bsnm{Wu}, \binits{H.}},
\bauthor{\bsnm{Zhang}, \binits{Z.}},
\bauthor{\bsnm{Shi}, \binits{S.}},
\bauthor{\bsnm{Wu}, \binits{Q.}},
\bauthor{\bsnm{Song}, \binits{H.}}:
\batitle{{Phrase dependency relational graph attention network for Aspect-based Sentiment Analysis}}.
\bjtitle{\textit{Knowledge-Based Systems}}
\bvolume{236},
\bfpage{107736}
(\byear{2022})
\doiurl{10.1016/j.knosys.2021.107736}
\end{barticle}
\endbibitem

\bibitem[\protect\citeauthoryear{Bird}{2006}]{Bird2006}
\begin{botherref}
\oauthor{\bsnm{Bird}, \binits{S.}}:
{NLTK: The natural language toolkit}.
\textit{COLING/ACL 2006 - 21st International Conference on Computational Linguistics and 44th Annual Meeting of the Association for Computational Linguistics, Proceedings of the Interactive Presentation Sessions},
69--72
(2006)
{\href{https://arxiv.org/abs/0205028}{{arXiv:0205028}}}
{[cs]}
\end{botherref}
\endbibitem

\bibitem[\protect\citeauthoryear{Biemann}{2005}]{Biemann2005}
\begin{barticle}
\bauthor{\bsnm{Biemann}, \binits{C.}}:
\batitle{{Ontology Learning from Text: A Survey of Methods}}.
\bjtitle{\textit{Journal for Language Technology and Computational Linguistics}}
\bvolume{20}(\bissue{2}),
\bfpage{75}--\blpage{93}
(\byear{2005})
\doiurl{10.21248/jlcl.20.2005.76}
\end{barticle}
\endbibitem

\bibitem[\protect\citeauthoryear{Antunes et~al.}{2022}]{Antunes2022}
\begin{botherref}
\oauthor{\bsnm{Antunes}, \binits{A.L.}},
\oauthor{\bsnm{Cardoso}, \binits{E.}},
\oauthor{\bsnm{Barateiro}, \binits{J.}}:
{Incorporation of Ontologies in Data Warehouse/Business Intelligence Systems - A Systematic Literature Review}.
\textit{International Journal of Information Management Data Insights}
\textbf{2}(2)
(2022)
\doiurl{10.1016/j.jjimei.2022.100131}
\end{botherref}
\endbibitem

\bibitem[\protect\citeauthoryear{Karabulut et~al.}{2024}]{Karabulut2024}
\begin{barticle}
\bauthor{\bsnm{Karabulut}, \binits{E.}},
\bauthor{\bsnm{Pileggi}, \binits{S.F.}},
\bauthor{\bsnm{Groth}, \binits{P.}},
\bauthor{\bsnm{Degeler}, \binits{V.}}:
\batitle{{Ontologies in digital twins: A systematic literature review}}.
\bjtitle{\textit{Future Generation Computer Systems}}
\bvolume{153}(\bissue{July 2023}),
\bfpage{442}--\blpage{456}
(\byear{2024})
\doiurl{10.1016/j.future.2023.12.013}
{\href{https://arxiv.org/abs/2308.15168}{{arXiv:2308.15168}}}
\end{barticle}
\endbibitem

\bibitem[\protect\citeauthoryear{W\k{a}tr\'{o}bski}{2020}]{Watrobski2020}
\begin{barticle}
\bauthor{\bsnm{W\k{a}tr\'{o}bski}, \binits{J.}}:
\batitle{Ontology learning methods from text - an extensive knowledge-based approach}.
\bjtitle{Procedia Computer Science}
\bvolume{176},
\bfpage{3356}--\blpage{3368}
(\byear{2020})
\doiurl{10.1016/j.procs.2020.09.061} .
\bcomment{Knowledge-Based and Intelligent Information \& Engineering Systems: Proceedings of the 24th International Conference KES2020}
\end{barticle}
\endbibitem

\bibitem[\protect\citeauthoryear{Maedche and Staab}{2001}]{Maedche2001}
\begin{barticle}
\bauthor{\bsnm{Maedche}, \binits{A.}},
\bauthor{\bsnm{Staab}, \binits{S.}}:
\batitle{{Ontology learning for the Semantic Web}}.
\bjtitle{\textit{IEEE Intelligent Systems}}
\bvolume{16}(\bissue{2}),
\bfpage{72}--\blpage{79}
(\byear{2001})
\doiurl{10.1109/5254.920602}
\end{barticle}
\endbibitem

\bibitem[\protect\citeauthoryear{Reyes-Pe{\~{n}}a and Tovar-Vidal}{2019}]{Reyes-Pena2019}
\begin{bchapter}
\bauthor{\bsnm{Reyes-Pe{\~{n}}a}, \binits{C.}},
\bauthor{\bsnm{Tovar-Vidal}, \binits{M.}}:
\bctitle{{Ontology: Components and Evaluation, a Review}}.
In: \bbtitle{\textit{Research in Computing Science}},
vol. \bseriesno{148},
pp. \bfpage{257}--\blpage{265}
(\byear{2019}).
\doiurl{10.13053/rcs-148-3-21}
\end{bchapter}
\endbibitem

\bibitem[\protect\citeauthoryear{Gruber}{1993}]{Gruber1993}
\begin{barticle}
\bauthor{\bsnm{Gruber}, \binits{T.R.}}:
\batitle{{A translation approach to portable ontology specifications}}.
\bjtitle{\textit{Knowledge Acquisition}}
\bvolume{5}(\bissue{2}),
\bfpage{199}--\blpage{220}
(\byear{1993})
\doiurl{10.1006/knac.1993.1008}
\end{barticle}
\endbibitem

\bibitem[\protect\citeauthoryear{Roussey et~al.}{2011}]{Roussey2011}
\begin{bchapter}
\bauthor{\bsnm{Roussey}, \binits{C.}},
\bauthor{\bsnm{Pinet}, \binits{F.}},
\bauthor{\bsnm{Kang}, \binits{M.A.}},
\bauthor{\bsnm{Corcho}, \binits{O.}}:
\bctitle{{An Introduction to Ontologies and Ontology Engineering}}.
In: \bbtitle{\textit{Ontologies in Urban Development Projects}},
pp. \bfpage{9}--\blpage{38}
(\byear{2011}).
\doiurl{10.1007/978-0-85729-724-2_2} .
\burl{http://link.springer.com/10.1007/978-0-85729-724-2{\_}2}
\end{bchapter}
\endbibitem

\bibitem[\protect\citeauthoryear{Stephan et~al.}{2009}]{Stephan}
\begin{bchapter}
\bauthor{\bsnm{Stephan}, \binits{G.}},
\bauthor{\bsnm{Pascal}, \binits{H.}},
\bauthor{\bsnm{Andreas}, \binits{A.}}:
\bctitle{{Knowledge Representation and Ontologies}}.
In: \bbtitle{\textit{Semantic Web Services}},
pp. \bfpage{51}--\blpage{105}.
\bpublisher{Springer},
\blocation{Berlin, Heidelberg}
(\byear{2009}).
\doiurl{10.1007/3-540-70894-4_3} .
\burl{http://link.springer.com/10.1007/3-540-70894-4{\_}3}
\end{bchapter}
\endbibitem

\bibitem[\protect\citeauthoryear{Xi et~al.}{2023}]{Xi2023}
\begin{botherref}
\oauthor{\bsnm{Xi}, \binits{Z.}},
\oauthor{\bsnm{Chen}, \binits{W.}},
\oauthor{\bsnm{Guo}, \binits{X.}},
\oauthor{\bsnm{He}, \binits{W.}},
\oauthor{\bsnm{Ding}, \binits{Y.}},
\oauthor{\bsnm{Hong}, \binits{B.}},
\oauthor{\bsnm{Zhang}, \binits{M.}},
\oauthor{\bsnm{Wang}, \binits{J.}},
\oauthor{\bsnm{Jin}, \binits{S.}},
\oauthor{\bsnm{Zhou}, \binits{T.} \bsuffix{...}}:
{The Rise and Potential of Large Language Model Based Agents: A Survey}.
\textit{arXiv: 2309.07864}
(2023)
{\href{https://arxiv.org/abs/2309.07864}{{arXiv:2309.07864}}}
\end{botherref}
\endbibitem

\bibitem[\protect\citeauthoryear{OpenAI et~al.}{2023}]{OpenAI2023}
\begin{botherref}
\oauthor{\bsnm{OpenAI}},
\oauthor{\bsnm{:}},
\oauthor{\bsnm{Achiam}, \binits{J.}},
\oauthor{\bsnm{Adler}, \binits{S.}},
\oauthor{\bsnm{Agarwal}, \binits{S.}},
\oauthor{\bsnm{Ahmad}, \binits{L.}},
\oauthor{\bsnm{Akkaya}, \binits{I.}},
\oauthor{\bsnm{Aleman}, \binits{F.L.}},
\oauthor{\bsnm{Almeida}, \binits{D.}},
\oauthor{\bsnm{Altenschmidt}, \binits{J.}},
\oauthor{\bsnm{Altman}, \binits{S.}},
\oauthor{\bsnm{Anadkat}, \binits{B.} \bsuffix{...}}:
{GPT-4 Technical Report}.
\textit{arXiv: 2303.08774}
(2023)
{\href{https://arxiv.org/abs/2303.08774}{{arXiv:2303.08774}}}
\end{botherref}
\endbibitem

\bibitem[\protect\citeauthoryear{{Gemini Team} et~al.}{2023}]{GeminiTeam2023}
\begin{botherref}
\oauthor{\bsnm{{Gemini Team}}},
\oauthor{\bsnm{Anil}, \binits{R.}},
\oauthor{\bsnm{Borgeaud}, \binits{S.}},
\oauthor{\bsnm{Wu}, \binits{Y.}},
\oauthor{\bsnm{Alayrac}, \binits{J.-B.}},
\oauthor{\bsnm{Yu}, \binits{J.}},
\oauthor{\bsnm{Soricut}, \binits{R.}},
\oauthor{\bsnm{Schalkwyk}, \binits{J.}},
\oauthor{\bsnm{Dai}, \binits{A.M.}},
\oauthor{\bsnm{Hauth}, \binits{O.} \bsuffix{...}}:
{Gemini: A Family of Highly Capable Multimodal Models}.
\textit{arXiv: 2312.11805}
(2023)
{\href{https://arxiv.org/abs/2312.11805}{{arXiv:2312.11805}}}
\end{botherref}
\endbibitem

\bibitem[\protect\citeauthoryear{Giglou et~al.}{2023}]{Giglou2023}
\begin{botherref}
\oauthor{\bsnm{Giglou}, \binits{H.B.}},
\oauthor{\bsnm{D'Souza}, \binits{J.}},
\oauthor{\bsnm{Auer}, \binits{S.}}:
{LLMs4OL: Large Language Models for Ontology Learning}.
\textit{arXiv: 2307.16648}
(Dcmi)
(2023)
{\href{https://arxiv.org/abs/2307.16648}{{arXiv:2307.16648}}}
\end{botherref}
\endbibitem

\bibitem[\protect\citeauthoryear{Konys}{2018}]{Konys2018}
\begin{barticle}
\bauthor{\bsnm{Konys}, \binits{A.}}:
\batitle{{Knowledge systematization for ontology learning methods}}.
\bjtitle{\textit{Procedia Computer Science}}
\bvolume{126},
\bfpage{2194}--\blpage{2207}
(\byear{2018})
\doiurl{10.1016/j.procs.2018.07.229}
\end{barticle}
\endbibitem

\bibitem[\protect\citeauthoryear{Ji et~al.}{2023}]{Ji2023}
\begin{bchapter}
\bauthor{\bsnm{Ji}, \binits{Z.}},
\bauthor{\bsnm{Yu}, \binits{T.}},
\bauthor{\bsnm{Xu}, \binits{Y.}},
\bauthor{\bsnm{Lee}, \binits{N.}},
\bauthor{\bsnm{Ishii}, \binits{E.}},
\bauthor{\bsnm{Fung}, \binits{P.}}:
\bctitle{{Towards Mitigating LLM Hallucination via Self Reflection}}.
In: \bbtitle{\textit{Findings of the Association for Computational Linguistics: EMNLP 2023}},
pp. \bfpage{1827}--\blpage{1843}.
\bpublisher{Association for Computational Linguistics},
\blocation{Stroudsburg, PA, USA}
(\byear{2023}).
\doiurl{10.18653/v1/2023.findings-emnlp.123} .
\burl{https://aclanthology.org/2023.findings-emnlp.123}
\end{bchapter}
\endbibitem

\bibitem[\protect\citeauthoryear{Martino et~al.}{2023}]{Martino2023}
\begin{botherref}
\oauthor{\bsnm{Martino}, \binits{A.}},
\oauthor{\bsnm{Iannelli}, \binits{M.}},
\oauthor{\bsnm{Truong}, \binits{C.}}:
{Knowledge Injection to Counter Large Language Model (LLM) Hallucination}.
\textit{The Semantic Web: ESWC 2023 Satellite Events},
182--185
(2023)
\doiurl{10.1007/978-3-031-43458-7_34}
\end{botherref}
\endbibitem

\bibitem[\protect\citeauthoryear{Rawte et~al.}{2023}]{Rawte2023}
\begin{botherref}
\oauthor{\bsnm{Rawte}, \binits{V.}},
\oauthor{\bsnm{Sheth}, \binits{A.}},
\oauthor{\bsnm{Das}, \binits{A.}}:
{A Survey of Hallucination in Large Foundation Models}.
\textit{arXiv:2309.05922}
(2023)
{\href{https://arxiv.org/abs/2309.05922}{{arXiv:2309.05922}}}
\end{botherref}
\endbibitem

\bibitem[\protect\citeauthoryear{Zhang et~al.}{2023}]{Zhang2023a}
\begin{botherref}
\oauthor{\bsnm{Zhang}, \binits{Y.}},
\oauthor{\bsnm{Li}, \binits{Y.}},
\oauthor{\bsnm{Cui}, \binits{L.}},
\oauthor{\bsnm{Cai}, \binits{D.}},
\oauthor{\bsnm{Liu}, \binits{L.}},
\oauthor{\bsnm{Fu}, \binits{T.}},
\oauthor{\bsnm{Huang}, \binits{X.}},
\oauthor{\bsnm{Zhao}, \binits{E.}},
\oauthor{\bsnm{Zhang}, \binits{Y.}},
\oauthor{\bsnm{Chen}, \binits{Y.}},
\oauthor{\bsnm{Wang}, \binits{L.}},
\oauthor{\bsnm{Luu}, \binits{A.T.}},
\oauthor{\bsnm{Bi}, \binits{W.}},
\oauthor{\bsnm{Shi}, \binits{F.}},
\oauthor{\bsnm{Shi}, \binits{S.}}:
{Siren's Song in the AI Ocean: A Survey on Hallucination in Large Language Models}.
\textit{arXiv: 2309.01219}
(2023)
{\href{https://arxiv.org/abs/2309.01219}{{arXiv:2309.01219}}}
\end{botherref}
\endbibitem

\bibitem[\protect\citeauthoryear{Huo et~al.}{2023}]{Huo2023}
\begin{botherref}
\oauthor{\bsnm{Huo}, \binits{S.}},
\oauthor{\bsnm{Arabzadeh}, \binits{N.}},
\oauthor{\bsnm{Clarke}, \binits{C.L.A.}}:
{Retrieving Supporting Evidence for LLMs Generated Answers}.
\textit{arXiv: 2306.13781}
(Figure 4)
(2023)
{\href{https://arxiv.org/abs/2306.13781}{{arXiv:2306.13781}}}
\end{botherref}
\endbibitem

\bibitem[\protect\citeauthoryear{Mosqueira-Rey et~al.}{2023}]{Mosqueira-Rey2023}
\begin{barticle}
\bauthor{\bsnm{Mosqueira-Rey}, \binits{E.}},
\bauthor{\bsnm{Hern{\'{a}}ndez-Pereira}, \binits{E.}},
\bauthor{\bsnm{Alonso-R{\'{i}}os}, \binits{D.}},
\bauthor{\bsnm{Bobes-Bascar{\'{a}}n}, \binits{J.}},
\bauthor{\bsnm{Fern{\'{a}}ndez-Leal}, \binits{{\'{A}}.}}:
\batitle{{Human-in-the-loop machine learning: a state of the art}}.
\bjtitle{\textit{Artificial Intelligence Review}}
\bvolume{56}(\bissue{4}),
\bfpage{3005}--\blpage{3054}
(\byear{2023})
\doiurl{10.1007/s10462-022-10246-w}
\end{barticle}
\endbibitem

\bibitem[\protect\citeauthoryear{Li et~al.}{2020}]{Li2020}
\begin{botherref}
\oauthor{\bsnm{Li}, \binits{N.}},
\oauthor{\bsnm{Adepu}, \binits{S.}},
\oauthor{\bsnm{Kang}, \binits{E.}},
\oauthor{\bsnm{Garlan}, \binits{D.}}:
{Explanations for human-on-the-loop: A probabilistic model checking approach}.
\textit{Proceedings - 2020 IEEE/ACM 15th International Symposium on Software Engineering for Adaptive and Self-Managing Systems, SEAMS 2020}
(April),
181--187
(2020)
\doiurl{10.1145/3387939.3391592}
\end{botherref}
\endbibitem

\bibitem[\protect\citeauthoryear{Nahavandi}{2017}]{Nahavandi2017}
\begin{barticle}
\bauthor{\bsnm{Nahavandi}, \binits{S.}}:
\batitle{{Trusted Autonomy Between Humans and Robots: Toward Human-on-the-Loop in Robotics and Autonomous Systems}}.
\bjtitle{\textit{IEEE Systems, Man, and Cybernetics Magazine}}
\bvolume{3}(\bissue{1}),
\bfpage{10}--\blpage{17}
(\byear{2017})
\doiurl{10.1109/MSMC.2016.2623867}
\end{barticle}
\endbibitem

\bibitem[\protect\citeauthoryear{Bastian et~al.}{2009}]{Bastian2009}
\begin{barticle}
\bauthor{\bsnm{Bastian}, \binits{M.}},
\bauthor{\bsnm{Heymann}, \binits{S.}},
\bauthor{\bsnm{Jacomy}, \binits{M.}}:
\batitle{{Gephi: An Open Source Software for Exploring and Manipulating Networks}}.
\bjtitle{\textit{Proceedings of the International AAAI Conference on Web and Social Media}}
\bvolume{3}(\bissue{1}),
\bfpage{361}--\blpage{362}
(\byear{2009})
\doiurl{10.1609/icwsm.v3i1.13937}
\end{barticle}
\endbibitem

\bibitem[\protect\citeauthoryear{Zhang and Peixoto}{2020}]{Zhang2020}
\begin{barticle}
\bauthor{\bsnm{Zhang}, \binits{L.}},
\bauthor{\bsnm{Peixoto}, \binits{T.P.}}:
\batitle{{Statistical inference of assortative community structures}}.
\bjtitle{\textit{Physical Review Research}}
\bvolume{2}(\bissue{4}),
\bfpage{043271}
(\byear{2020})
\doiurl{10.1103/PhysRevResearch.2.043271}
\end{barticle}
\endbibitem

\bibitem[\protect\citeauthoryear{Spinde et~al.}{2021}]{Spinde2021}
\begin{barticle}
\bauthor{\bsnm{Spinde}, \binits{T.}},
\bauthor{\bsnm{Rudnitckaia}, \binits{L.}},
\bauthor{\bsnm{Mitrovi{\'{c}}}, \binits{J.}},
\bauthor{\bsnm{Hamborg}, \binits{F.}},
\bauthor{\bsnm{Granitzer}, \binits{M.}},
\bauthor{\bsnm{Gipp}, \binits{B.}},
\bauthor{\bsnm{Donnay}, \binits{K.}}:
\batitle{{Automated identification of bias inducing words in news articles using linguistic and context-oriented features}}.
\bjtitle{\textit{Information Processing and Management}}
\bvolume{58}(\bissue{3}),
\bfpage{102505}
(\byear{2021})
\doiurl{10.1016/j.ipm.2021.102505}
\end{barticle}
\endbibitem

\bibitem[\protect\citeauthoryear{Hamborg et~al.}{2019}]{Hamborg2019a}
\begin{barticle}
\bauthor{\bsnm{Hamborg}, \binits{F.}},
\bauthor{\bsnm{Zhukova}, \binits{A.}},
\bauthor{\bsnm{Gipp}, \binits{B.}}:
\batitle{{Automated identification of media bias by word choice and labeling in news articles}}.
\bjtitle{\textit{Proceedings of the ACM/IEEE Joint Conference on Digital Libraries}}
\bvolume{2019-June}(\bissue{1}),
\bfpage{196}--\blpage{205}
(\byear{2019})
\doiurl{10.1109/JCDL.2019.00036}
\end{barticle}
\endbibitem

\bibitem[\protect\citeauthoryear{Bennett}{2021}]{Bennett2021}
\begin{bchapter}
\bauthor{\bsnm{Bennett}, \binits{J.T.}}:
\bctitle{{Left, Right, or Always Establishment? The Bias Issue}}.
In: \bbtitle{\textit{The History and Politics of Public Radio}},
pp. \bfpage{85}--\blpage{97}
(\byear{2021}).
\doiurl{10.1007/978-3-030-80019-2_6} .
\burl{https://link.springer.com/10.1007/978-3-030-80019-2{\_}6}
\end{bchapter}
\endbibitem

\bibitem[\protect\citeauthoryear{Newman et~al.}{2023}]{Newman2023}
\begin{botherref}
\oauthor{\bsnm{Newman}, \binits{N.}},
\oauthor{\bsnm{Fletcher}, \binits{R.}},
\oauthor{\bsnm{Eddy}, \binits{K.}},
\oauthor{\bsnm{Robertson}, \binits{C.T.}},
\oauthor{\bsnm{Nielsen}, \binits{R.K.}}:
{Digital News Report 2023}.
Technical report
(2023)
\end{botherref}
\endbibitem

\bibitem[\protect\citeauthoryear{Gan et~al.}{2024}]{Gan2024}
\begin{botherref}
\oauthor{\bsnm{Gan}, \binits{L.}},
\oauthor{\bsnm{Yang}, \binits{T.}},
\oauthor{\bsnm{Huang}, \binits{Y.}},
\oauthor{\bsnm{Yang}, \binits{B.}},
\oauthor{\bsnm{Luo}, \binits{Y.Y.}},
\oauthor{\bsnm{Richard}, \binits{L.W.C.}},
\oauthor{\bsnm{Guo}, \binits{D.}}:
{Experimental Comparison of Three Topic Modeling Methods with LDA, Top2Vec and BERTopic}.
\textit{Artificial Intelligence and Robotics},
376--391
(2024)
\doiurl{10.1007/978-981-99-9109-9_37}
\end{botherref}
\endbibitem

\bibitem[\protect\citeauthoryear{Hamborg et~al.}{2020}]{Hamborg2020}
\begin{barticle}
\bauthor{\bsnm{Hamborg}, \binits{F.}},
\bauthor{\bsnm{Meuschke}, \binits{N.}},
\bauthor{\bsnm{Gipp}, \binits{B.}}:
\batitle{{Bias-aware news analysis using matrix-based news aggregation}}.
\bjtitle{\textit{International Journal on Digital Libraries}}
\bvolume{21}(\bissue{2}),
\bfpage{129}--\blpage{147}
(\byear{2020})
\doiurl{10.1007/s00799-018-0239-9}
\end{barticle}
\endbibitem

\bibitem[\protect\citeauthoryear{Hamborg et~al.}{2017}]{Hamborg2017}
\begin{bchapter}
\bauthor{\bsnm{Hamborg}, \binits{F.}},
\bauthor{\bsnm{Meuschke}, \binits{N.}},
\bauthor{\bsnm{Gipp}, \binits{B.}}:
\bctitle{Matrix-based news aggregation: Exploring different news perspectives}.
In: \bbtitle{2017 ACM/IEEE Joint Conference on Digital Libraries (JCDL)},
pp. \bfpage{1}--\blpage{10}
(\byear{2017}).
\doiurl{10.1109/JCDL.2017.7991561}
\end{bchapter}
\endbibitem

\end{thebibliography}

\end{document}